%% file: main.tex
\documentclass[10pt,journal,compsoc]{IEEEtran}

\ifCLASSOPTIONcompsoc
\usepackage[nocompress]{cite}
\else
\usepackage{cite}
\fi

\ifCLASSINFOpdf
\usepackage[pdftex]{graphicx}
\graphicspath{{./images/}}
\DeclareGraphicsExtensions{.pdf,.png}
\else
\fi
\usepackage{amsmath,amssymb,amsthm, upgreek}
\usepackage{bm}

\usepackage{algorithm}
\usepackage{algpseudocode}
\newcommand{\keypoint}[1]{\vspace{0.1cm}\noindent\textbf{#1}}
\newcommand{\cut}[1]{}

\newcommand{\name}{H-SRDC}%

\def\eg{\emph{e.g.}}
\def\ie{\emph{i.e.}}
\def\cf{cf.}
\def\etal{\emph{et al.}}

\usepackage{array}

\ifCLASSOPTIONcompsoc
\usepackage[caption=false,font=normalsize,labelfont=sf,textfont=sf]{subfig}
\else
\usepackage[caption=false,font=footnotesize]{subfig}
\fi

\usepackage{dblfloatfix}

\usepackage{url}

\usepackage{epsfig}
\usepackage{color}
\usepackage{verbatim}
\usepackage{multirow}

\usepackage{hyperref}

\begin{document}

\title{Towards Uncovering the Intrinsic Data Structures for Unsupervised Domain Adaptation using Structurally Regularized Deep Clustering}

	%
	
\author{Hui~Tang,
		Xiatian~Zhu,
		Ke~Chen,
		Kui~Jia, 
		and 
		C. L. Philip Chen,~\IEEEmembership{Fellow,~IEEE}
		\IEEEcompsocitemizethanks{\IEEEcompsocthanksitem Hui Tang, Ke Chen, and Kui Jia are with the School of Electronic and Information Engineering, South China University of Technology, Guangzhou, China. 
			E-mail: eehuitang@mail.scut.edu.cn,
			chenk@scut.edu.cn,
			kuijia@scut.edu.cn. 
			Correspondence to Kui Jia. 
			\IEEEcompsocthanksitem Xiatian Zhu is with Centre for Vision, Speech and Signal Processing (CVSSP), University of Surrey, UK.
			E-mail: eddy.zhuxt@gmail.com.
			\IEEEcompsocthanksitem C. L. Philip Chen is with the School of Computer Science and Engineering, South China University of Technology, Guangzhou, China. 
			E-mail: philip.chen@ieee.org.
		}
}

	%



\IEEEtitleabstractindextext{
\begin{abstract}
Unsupervised domain adaptation (UDA) is to learn classification models that make predictions for unlabeled data on a target domain, given labeled data on a source domain whose distribution diverges from the target one. Mainstream UDA methods strive to learn domain-aligned features such that classifiers trained on the source features can be readily applied to the target ones. Although impressive results have been achieved, these methods have a potential risk of damaging the \emph{intrinsic} data structures of target discrimination, raising an issue of generalization particularly for UDA tasks in an inductive setting. To address this issue, we are motivated by a UDA assumption of \emph{structural similarity} across domains, and propose to directly uncover the intrinsic target discrimination via constrained clustering, where we constrain the clustering solutions using structural source regularization that hinges on the very same assumption. Technically, we propose a hybrid model of \emph{Structurally Regularized Deep Clustering}, which integrates the regularized discriminative clustering of target data with a generative one, and we thus term our method as H-SRDC. Our hybrid model is based on a deep clustering framework that minimizes the Kullback-Leibler divergence between the distribution of network prediction and an auxiliary one, where we impose structural regularization by learning domain-shared classifier and cluster centroids. By enriching the structural similarity assumption, we are able to extend H-SRDC for a pixel-level UDA task of semantic segmentation. We conduct extensive experiments on seven UDA benchmarks of image classification and semantic segmentation. With no explicit feature alignment, our proposed H-SRDC outperforms all the existing methods under both the inductive and transductive settings. We make our implementation codes publicly available at \url{https://github.com/huitangtang/H-SRDC}.
\end{abstract}

}

\maketitle

	\IEEEdisplaynontitleabstractindextext

	%
\IEEEpeerreviewmaketitle

\IEEEraisesectionheading{\section{Introduction}
\label{sec:introduction}}






\IEEEPARstart{I}{n} many practical applications of machine learning, the problem of interest is concerned with learning from data on a domain where, due to practical constraints and/or expenses, data annotations are difficult to acquire, and a standard supervised training cannot be readily applied; instead, labeled data on a \emph{different but related} domain can be obtained relatively easily.
This creates a learning scenario in which one is tempted to leverage the labeled data on the \emph{source} domain to help learn machine learning models for a transferrable use on the unlabeled \emph{target} domain, \ie, the problem of \emph{unsupervised domain adaptation (UDA)} \cite{transfer_learning_survey,da_theory2}. UDA typically assumes a shared label space between the source and target domains, and its technical challenge arises from the assumed existence of distribution divergence between the two domains.

A rich literature of UDA research has been developed in the past decades \cite{survey_deep_vis_da}. Among them, mainstream methods \cite{dan,dann,mcd,symnets,pfan,tpn,bnm} are motivated by the seminal theories  \cite{da_theory2,da_theory1,mansour09} that bound the expected errors of classification models on the target domain by quantities involving classifier-induced divergence between feature distributions of the two domains, \eg, those recent ones based on adversarial training of deep networks \cite{dann,mcd,symnets}. Consequently, these methods strive to minimize domain divergence by learning aligned features between the two domains, such that classifiers trained on the features of source domain can be readily applied to the target ones. Despite the impressive results achieved, these methods have a potential risk of damaging the \emph{intrinsic} data structures of target discrimination, as analyzed recently in \cite{da_theory3}. We note that more importantly, this shortcoming would become severer in the practically more useful setting of \emph{inductive} UDA (analogous to the setting of inductive transfer learning in \cite{transfer_learning_survey}), where the objective is to learn classification models as off-the-shelf ones such that they can be used for held-out data sampled from the same target domain; adapting classifiers to the damaged discrimination of target data by feature alignment of existing methods would be less effective for inductive UDA, since the held-out target data still follow the undamaged, intrinsic data discrimination.

To overcome such limitations in existing methods, we first revisit the general UDA assumptions made in existing research \cite{da_theory2,mansour09,it_cluster_uda2}, and summarize those in \cite{it_cluster_uda2} as the \emph{structural similarity} between the source and target domains
, which includes the notions of \emph{domain-wise discrimination} and \emph{class-wise closeness}. Simply put, the former notion assumes the existence of intrinsic structures of discriminative data clusters in individual domains, and the latter one assumes that clusters of the two domains corresponding to the same semantic class are geometrically close. These assumptions motivate us to consider a UDA approach that directly \emph{uncovers} the intrinsic discrimination of target data via constrained clustering, where we propose to constrain the clustering solutions using structural source regularization hinging on the very same assumptions.

Technically, we propose a hybrid model of deep clustering that integrates a regularized discriminative clustering with a generative one. Among various deep network based clustering algorithms \cite{InfoGAN,GMVAE,DeepClusterRelativeEMICCV17}, we choose a simple but flexible framework \cite{DeepClusterRelativeEMICCV17}, which performs clustering by minimizing the Kullback-Leibler (KL) divergence between the predictive label distribution of the network and an introduced auxiliary one. For discriminative part of our hybrid model, the structural source regularization is simply achieved by training the same network layers of classifier using labeled source data, \ie, a strategy of joint network training; for the generative part, we learn cluster centroids in the deep feature space to enable probabilistic data modeling, and the structural source regularization is achieved by making the centroids common to the source and target domains, where we borrow ideas from set transformer \cite{sab} and learn the cluster centroids in a feed-forward manner using self-attentive feature interactions of training instances. We empirically observe that generative clustering modulates the feature space learning, which potentially enhances the uncovering of intrinsic discrimination by providing benefits complementary to the discriminative ones. We term our method of Structurally Regularized Deep Clustering as H-SRDC, to emphasize both its hybrid nature and its extension to the method of SRDC proposed in the preliminary version \cite{srdc} of this work.

In the present paper, we also contribute in a second UDA task of semantic segmentation, which is to learn domain-adapted models to classify each pixel in an input image into one of multiple semantic classes. We note that by treating observations at each pixel of the image as a data instance, our proposed H-SRDC can be readily applied. To further improve the performance, we borrow ideas from existing methods \cite{Adapt_SegMap,advent} and propose to enrich our UDA assumption of structural similarity with a third notion of \emph{layout-wise consistency}, which states that the spatial layout of semantic segmentation maps is consistent between the source and target domains. Implementing this notion into the H-SRDC objective gives our method for domain-adapted semantic segmentation. We present extensive experiments on the benchmark datasets of Office-31 \cite{office31}, ImageCLEF-DA \cite{imageclefda}, Office-Home \cite{officehome}, VisDA-2017 \cite{visda2017}, and Digits \cite{svhn,mnist,usps} for image classification, and on the UDA tasks among GTA5 \cite{gta5}, SYNTHIA \cite{synthia}, and Cityscapes \cite{cityscapes} for semantic segmentation. Our proposed H-SRDC outperforms all the existing methods under both the setting of inductive UDA and the more conventional setting of transductive UDA, where domain-adapted models are directly evaluated on the target data that are involved in the training.

\subsection{Relations with Existing Works}
\label{SecLiterature}

In this section, we organize our brief review of existing methods into the following three categories. We also discuss their relations with our proposed one. 

\keypoint{Domain Adaptation for Image Classification.}
There exists a huge literature of UDA methods for image classification. We focus our review on those representative ones, particularly those by learning aligned deep features and those incorporating modern techniques of deep clustering.
One may refer to \cite{survey_deep_vis_da} for a comprehensive review of existing methods.

To achieve domain adaptation, the methods \cite{dan,rtn,BeyondSW} learn deep features to reduce the classical measure of maximum mean discrepancy (MMD) across domains. A strategy of adversarial training is subsequently used in \cite{dann,adda,iCAN} to further reduce the domain discrepancy. These methods are designed for domain-level alignment; however, reducing discrepancies of conditional distributions towards category-level alignments are more desired for better UDA. To this end, multiplicative interactions of feature and category predictions are used in \cite{cdan,mada} to achieve the goal. The methods \cite{pfan,cat,tpn} do so in an alternative manner by aligning feature means of individual categories between the source and target data. More recently, Chen \etal~\cite{bsp} show that discriminative structures of target data may be degraded by adversarial feature alignment; they apply spectral decompositions to the instance features, and penalize the singular values corresponding to the singular vectors that learn aligned features. We are motivated by the same degradation of intrinsic target discrimination; we instead propose H-SRDC as a regularized deep clustering solution to address the issue.

Unsupervised domain adaptation is by nature to cluster the unlabeled target data, given regularization from the labeled source data. As such, principles of unsupervised learning, \eg, the cluster assumption \cite{ClusterAssumption}, are typically applicable to UDA tasks. The cluster assumption states that the classification boundaries should not pass through high-density regions, but instead lie in low-density regions. To enforce the cluster assumption, conditional entropy minimization \cite{min_ent,em} 
is widely used in the UDA community \cite{rca,rtn,it_cluster_uda,dwt_mec,it_cluster_uda2,dirt_t,larger_norm,symnets}. 
Kang \etal~\cite{can} adopt the spherical $K$-means to assign target labels. Deng \etal~\cite{cat} employ a deep clustering loss based on a Fisher-like criterion \cite{sntg}. Most of these methods use clustering of target data to help improve feature alignment; Shi and Sha \cite{it_cluster_uda2} also explicitly force domain alignment. In contrast, with no explicit feature alignment, the present method of H-SRDC uses regularized deep clustering aiming to directly uncover the intrinsic discrimination of target data.

\keypoint{Domain Adaptation for Semantic Segmentation.}
Existing domain adaptation strategies for semantic segmentation \cite{siban,cycada,dise} are similar to those for image classification, with additional consideration such as constraints on the spatial layout of output segmentation maps \cite{Adapt_SegMap,advent,clan,curriculum_da}.
For example, Hoffman \etal~\cite{cycada} adopt adversarial domain adaptation at both pixel and feature levels, by applying cycle-consistency \cite{ccan} and semantic consistency losses. 
In \cite{siban}, information bottleneck is utilized to eliminate nuisance factors and maintain pure semantic information in features
. 
Zhang \etal~\cite{curriculum_da} first infer the properties on semantic layout for target domain images, and then regularize their output segmentation maps to follow the inferred properties. 
Vu \etal~\cite{advent} enforce structural consistency across domains by conditional entropy minimization and distribution matching in terms of weighted self-information maps. Our extension of H-SRDC for semantic segmentation incorporates these established domain knowledge into our regularized deep clustering framework. As far as we know, we are the first to use a direct clustering solution and achieve superior performance on the semantic segmentation task.

\keypoint{Transductive or Inductive Domain Adaptation.}
Existing research on UDA tasks does not consider the nuanced difference between the transductive and inductive settings. Comparisons are usually made on different benchmarks by following the setting conventions. For example, for image classification, existing methods report results in the transductive setting for the benchmarks of Office-31 \cite{office31}, ImageCLEF-DA \cite{imageclefda}, Office-Home \cite{officehome}, and VisDA-2017 \cite{visda2017}, and report results in the inductive setting for the benchmark of Digits. The experiments on the semantic segmentation benchmarks of GTA5 \cite{gta5}, SYNTHIA \cite{synthia}, and Cityscapes \cite{cityscapes} are also reported in the inductive setting. It is arguably more useful to study the inductive setting of UDA, since once trained in this setting, the adaptation models can be directly used for held-out test sets. In this work, we report comprehensive experiments in both the inductive and transductive settings. We expect that our results contribute to the community as the new benchmarks.

\subsection{Contributions}

A preliminary version of this work has been published as an oral presentation in \cite{srdc}, where we have proposed the basic strategy of Structurally Regularized Deep discriminative Clustering (SRDC) for UDA tasks in the transductive setting. We re-state its main technical contributions as follows.
\begin{itemize}
\item To address a potential issue of damaging the \emph{intrinsic} data discrimination by explicitly learning domain-aligned features, we propose a method of SRDC \cite{srdc} that makes use of source-regularized, deep discriminative clustering to directly uncover the intrinsic structures of target discrimination. The method is motivated by our assumption of \emph{structural similarity} between the two domains.

\item We technically achieve SRDC based on a simple but flexible framework of deep clustering, which minimizes the KL divergence between the distribution of network prediction and an auxiliary one; replacing the auxiliary distribution with that of ground-truth labels of source data implements the structural source regularization via a simple strategy of joint network training.
\end{itemize}
In the present paper, we improve the method in \cite{srdc} as a hybrid model of H-SRDC, and extend H-SRDC for a second UDA task of semantic segmentation. We organize our evaluations in both the inductive and transductive UDA settings. H-SRDC achieves the new state of the art across a range of UDA benchmarks in both the settings. Our new contributions are summarized as follows.
\begin{itemize}
\item We propose a hybrid model of H-SRDC that integrates regularized discriminative clustering with a generative one. Regularized generative clustering is achieved by learning in a feed-forward manner a set of cluster centroids common to the source and target domains. We empirically verify that our learning of cluster centroids modulates the feature space learning, rather than align the features across domains; it enhances the uncovering of intrinsic data structures by providing benefits complementary to the discriminative one.

\item Our proposed H-SRDC can be readily applied to a pixel-level UDA task of semantic segmentation. To further improve the performance, we enrich our UDA assumption of \emph{structural similarity} with a third notion of \emph{layout-wise consistency}. Implementing this notion into the learning objective of H-SRDC gives our method for domain-adapted semantic segmentation. Efficacy of this extension is verified empirically.

\item To the best of our knowledge, we organize, for the first time, a comprehensive evaluation of different UDA methods in both the inductive and transductive settings. Across a range of UDA benchmarks for image classification and semantic segmentation, our proposed H-SRDC is consistently superior to existing ones under both the settings. Careful ablation studies also reveal the internal mechanism of H-SRDC.
\end{itemize}

\section{Problem Statement}
\label{SecProbDefinition}

In unsupervised domain adaptation (UDA), we assume a {\em labeled} set of examples $\{(\bm{x}_i^s, y_i^s)\}_{i=1}^{n_s}$ from a source domain $\mathcal{S}$, and an {\em unlabeled} set $\{\bm{x}_i^t\}_{i=1}^{n_t}$ from a target domain $\mathcal{T}$. The two domains share a common label space $\mathcal{Y}$. Let $|\mathcal{Y}| = K$, and we have $y^s \in \{1, 2, \dots, K\}$ for any source instance $\bm{x}^s$. There exist two typical settings in the literature of UDA: \emph{transductive} UDA aims to learn prediction models that directly assign labels to the target instances $\{\bm{x}_i^t\}_{i=1}^{n_t}$, and \emph{inductive} UDA is to measure performance of the learned models on held-out sets of instances that are sampled from the same $\mathcal{T}$. By convention, the two settings are respectively considered in the UDA tasks for object classification \cite{dann,mcd,dan,can} and semantic segmentation \cite{cycada,dise,clan,advent}.


\begin{figure*}[t]
	\begin{center}
		\includegraphics[width=1.0\textwidth]{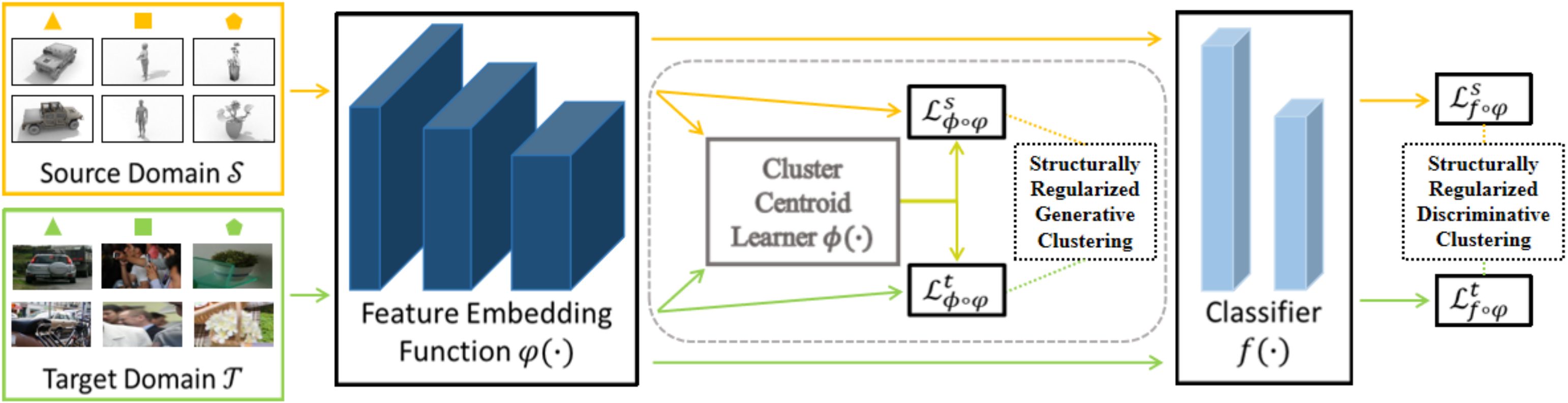}
	\end{center}
	\vskip -0.5cm
	\caption{
		A schematic illustration of our proposed hybrid model of {\em Structurally Regularized Deep Clustering} (\name{}).
		It is formulated as a constrained clustering framework with two key components:
		(a) structurally regularized discriminative clustering, which uncovers the intrinsic discrimination of unlabeled target data with structural regularization from the labeled source data (\cf~Section \ref{SecSRDisC}),
		and (b) structurally regularized generative clustering, which modulates and potentially enhances the learning in the feature space by generative learning of cluster centroids using self-attentive interactions of instance features (\cf~Section \ref{SecSRGenC}). Once trained, the classification model $f\circ \varphi$ is deployed for the UDA task. In this figure, orange and green arrows represent the data flows from the source and target domains, respectively.
	}
	\label{fig:pipeline}\vspace{-0.3cm}
\end{figure*}

\subsection{Motivations for Uncovering the Intrinsic Target Discrimination}
\label{SecMotivation}

Given the discrepancies between the source and target domains, domain adaptation is less feasible without certain assumptions on their similarities. In fact, domain closeness is generally assumed in UDA either theoretically \cite{da_theory2,mansour09} or intuitively \cite{it_cluster_uda2}. In this work, we summarize the assumptions in \cite{it_cluster_uda2} as the \emph{structural similarity} between the source and target domains, which includes the following notions of domain-wise discrimination and class-wise closeness
.
\begin{itemize}
\item \emph{Domain-wise discrimination} assumes that there exist \emph{intrinsic} structures of data discrimination in individual domains, \ie, data in either source or target domains are discriminatively clustered corresponding to the shared label space.
\item \emph{Class-wise closeness} assumes that clusters of the two domains corresponding to the same class label are geometrically close.
\end{itemize}

Based on these assumptions, many of existing works \cite{dann,cdan,mada,mcd,adda,mdd} take the strategy of learning \emph{aligned} feature representations between the two domains, such that classifiers trained on source features can be readily applied to the target ones. However, such a strategy has a potential risk of damaging the intrinsic data discrimination on the target domain, as discussed in some of recent works \cite{da_theory3,bsp,it_cluster_uda2}. 
We note that more importantly, classifiers adapting to the damaged discrimination of target data would be less effective for tasks of inductive UDA, since in inductive UDA, the learned classifiers are expected to be used as off-the-shelf models for held-out target data, which still follow the undamaged, intrinsic data discrimination. As such, it is ideal that domain-adapted classifiers should be \emph{consistent} when learning with different source domains, and they should not deviate too much from the oracle one obtained, \eg, by training on target samples with ground-truth labels, which is deemed to represent the intrinsic target discrimination.

Based on the above analysis, we are motivated to directly uncover the intrinsic target discrimination via \emph{discriminative} clustering of the target data. To leverage the labeled source data, we propose to constrain the clustering solutions using \emph{structural source regularization} that hinges on our assumed structural similarity across domains. We note that quite a few recent methods \cite{cat,can,dirt_t} consider clustering of target data as well; however, they still do \emph{explicit} feature alignment across domains, thus prone to the aforementioned risk of damaged intrinsic target discrimination. In contrast, we propose to modulate and potentially enhance the intrinsic target discrimination via \emph{generative} learning of cluster centroids that are common to the two domains. Generative clustering provides benefits complementary to the discriminative one, and we empirically observe that it only modulates, rather than aligns, the feature learning for instances on the two domains.

\subsection{Learning Setups and Notations}

We consider a feature embedding function $\varphi: \mathcal{X} \to \mathcal{Z}$, parameterized by $\bm{\theta}_{\varphi}$, which lifts any input instance $\mathbf{x} \in \mathcal{X}$ to the feature space $\mathcal{Z}$, and a classifier $f: \mathcal{Z} \to [0,1]^K$ parameterized by $\bm{\theta}_f$. For any input $\bm{x}$, we write its $d$-dimensional feature representation as $\bm{z} = \varphi(\bm{x}) \in \mathbb{R}^d$, and its classification probability vector as $\bm{p} = f(\bm{z}) = [p_1; \dots; p_K]$, subject to $\sum_{k=1}^K p_k = 1$. In this work, we implement the composite function $f \circ \varphi$ as deep networks, which are trained on the labeled source data $\{(\bm{x}_i^s, y_i^s)\}_{i=1}^{n_s}$ and unlabeled target data $\{\bm{x}_i^t\}_{i=1}^{n_t}$. We correspondingly write as $\{\bm{z}_i^s\}_{i=1}^{n_s}$ and $\{\bm{z}_i^t\}_{i=1}^{n_t}$ for their feature vectors computed from $\varphi(\cdot)$, and as $\{\bm{p}_i^s\}_{i=1}^{n_s}$ and $\{\bm{p}_i^t\}_{i=1}^{n_t}$ for their probability vectors of network output. The $k^{th}$ element of any $\bm{p}_i^s$ or $\bm{p}_i^t$ is respectively written as $p_{i,k}^s$ and $p_{i,k}^t$. For ease of presentation, we also write collectively as $\bm{X}^s = \{\bm{x}_i^s\}_{i=1}^{n_s}$, $\bm{X}^t = \{\bm{x}_i^t\}_{i=1}^{n_t}$, $\bm{P}^s = \{\bm{p}_i^s\}_{i=1}^{n_s}$, $\bm{P}^t = \{\bm{p}_i^t\}_{i=1}^{n_t}$, $\bm{Z}^s = \{\bm{z}_i^s\}_{i=1}^{n_s}$, and $\bm{Z}^t = \{\bm{z}_i^t\}_{i=1}^{n_t}$
. With a slight abuse of notations, we also let $\bm{Z}^s = [\bm{z}_1^s, \dots, \bm{z}_{n_s}^s] \in \mathbb{R}^{d\times n_s}$, $\bm{Z}^t = [\bm{z}_1^t, \dots, \bm{z}_{n_t}^t] \in \mathbb{R}^{d\times n_t}$, and $\bm{Z} = [ \bm{Z}^s, \bm{Z}^t ] \in \mathbb{R}^{d\times n}$ with $n = n_s + n_t$, when the contexts require.

\section{The Proposed Method}
\label{SecMethod}

\subsection{Deep Discriminative Target Clustering with Structural Source Regularization}
\label{SecSRDisC}

In order to uncover the intrinsic discrimination of the target domain, we opt for direct clustering of target instances with structural regularization from the source domain. Among various clustering methods \cite{InfoGAN,GMVAE,DeepClusterRelativeEMICCV17}, we choose a flexible framework of deep discriminative clustering \cite{DeepClusterRelativeEMICCV17}, which minimizes the Kullback-Leibler (KL) divergence between the predictive label distribution of the network and an introduced auxiliary one; by replacing the auxiliary distribution with that of ground-truth labels of source data, we easily implement the structural source regularization via a simple strategy of joint network training. We also enhance structural regularization with soft selection of less divergent source examples. Technical details are presented as follows.

\subsubsection{Deep Discriminative Target Clustering}
\label{sec:deep_discrimimnative_target_clustering}

Given the unlabeled $\{\bm{x}_i^t\}_{i=1}^{n_t}$, the network $f\circ \varphi$ outputs probability vectors of $\bm{P}^t = \{\bm{p}_i^t\}_{i=1}^{n_t}$. To implement deep discriminative clustering \cite{DeepClusterRelativeEMICCV17,DeepClusterLink}, we first introduce an auxiliary counterpart $\bm{Q}^t = \{\bm{q}_i^t\}_{i=1}^{n_t}$. The proposed method then alternates in (1) updating $\bm{Q}^t$, and (2) using the updated $\bm{Q}^t$ as labels to train the network to update parameters $\{\bm{\theta}_{\varphi},\bm{\theta}_f\}$, by optimizing the following objective of deep discriminative clustering
\begin{eqnarray}\label{EqnAuxiTarDistrConstrain}
	\begin{aligned}
	\min_{\bm{Q}^t, \{\bm{\theta}_{\varphi},\bm{\theta}_f\}} \mathcal{L}_{f\circ\varphi}^t = {\rm KL}(\bm{Q}^t||\bm{P}^t) + \sum_{k=1}^K {\varrho}_k^t \log {\varrho}_k^t ,
	\end{aligned}
\end{eqnarray}
where ${\rm KL}(\cdot || \cdot)$ defines KL divergence between the two discrete probability distributions $\bm{P}^t$ and $\bm{Q}^t$, which is spelled out as
	\begin{align}
	\notag {\rm KL}(\bm{Q}^t||\bm{P}^t) & = \frac{1}{n_t}\sum_{i=1}^{n_t}\sum_{k=1}^K q_{i,k}^t \log \frac{q_{i,k}^t}{p_{i,k}^t} ,
\end{align}
and ${\varrho}_k^t$ in the second term is computed as
\begin{align}
{\varrho}_k^t = \frac{1}{n_t} \sum_{i=1}^{n_t} q_{i,k}^t . \nonumber
\end{align}
The use of second term in Eq. \eqref{EqnAuxiTarDistrConstrain} is to encourage entropy maximization of the averaged $K$ probability predictions over the $n_t$ instances, such that the assignments of $\{\bm{p}_i^t\}_{i=1}^{n_t}$ (via $\{\bm{q}_i^t\}_{i=1}^{n_t}$) into the $K$ clusters are balanced; otherwise as suggested by \cite{PeronaMIDisCluster}, degenerate solutions would be obtained that merge clusters by removing cluster boundaries. Given the lack of prior knowledge about target label distribution, we simply rely on this term to account for a uniform one.

Optimizing Eq. \eqref{EqnAuxiTarDistrConstrain} takes the following alternating steps.
\begin{itemize}
\item \textbf{Auxiliary distribution update.} Fix network parameters $\{\bm{\theta}_{\varphi},\bm{\theta}_f\}$ (and $\{\bm{p}_i^t\}_{i=1}^{n_t}$ of target instances). Setting the (approximate) gradient of Eq. \eqref{EqnAuxiTarDistrConstrain} as zero
, 
the following closed-form solution is given by \cite{DeepClusterRelativeEMICCV17}
		\begin{eqnarray}\label{EqnAuxiTarDistrCloseSol_DisClust}
		\begin{aligned}
		q_{i,k}^t = \frac{p_{i,k}^t / (\sum_{i'=1}^{n_t}p_{i',k}^t)^{\frac{1}{2}}}{\sum_{k'=1}^K p_{i,k'}^t / (\sum_{i'=1}^{n_t}p_{i',k'}^t)^{\frac{1}{2}}}.
		\end{aligned}
		\end{eqnarray}
		
		\item \textbf{Network update.} By fixing $\bm{Q}^t$, this step is equivalent to training the network via a cross-entropy loss using $\bm{Q}^t$ as labels, giving rise to
		\begin{eqnarray}\label{EqnDeepDisTar}
		\begin{aligned}
		\min\limits_{\bm{\theta}_{\varphi},\bm{\theta}_f} - \frac{1}{n_t} \sum_{i=1}^{n_t} \sum_{k=1}^{K} q_{i,k}^t \log p_{i,k}^t .
		\end{aligned}
		\end{eqnarray}
	\end{itemize}

\noindent\textbf{Remarks.} Given unlabeled target data alone, the objective (\ref{EqnAuxiTarDistrConstrain}) itself is not guaranteed to have sensible solutions to uncover the intrinsic discrimination of target data, since the auxiliary distribution $\bm{Q}^t$ could be arbitrary whose optimization is subject to no proper constraints. To guarantee sensible solutions, deep clustering methods \cite{DeepClusterRelativeEMICCV17,UnsupervisedEmbeddingICML16} usually employ an additional reconstruction loss as a data-dependent regularizer. In this work, we introduce the following structural source regularization that serves a similar purpose as that of the reconstruction ones used in \cite{DeepClusterRelativeEMICCV17,UnsupervisedEmbeddingICML16}.

\subsubsection{Structural Source Regularization by Learning a Common Model of Classifier}
\label{sec:structural_source_regularization}
	
Based on the UDA assumption of structural similarity made in Section \ref{SecProbDefinition}, we propose to regularize the clustering of target data simply by training the same network $f\circ\varphi$ using labeled source data, \ie, a strategy of joint network training. Note that the $K$-way classifier $f$ defines hyperplanes that partition the feature space $\mathcal{Z}$ into regions whose number is bounded by $2^K$, and $K$ ones among them are uniquely responsible for the $K$ classes. Given that the two domains share the same label space, joint training would \emph{ideally} push instances of the two domains from any class into a same region in $\mathcal{Z}$, thus \emph{implicitly} achieving adaptation between the two domains. 
	
Technically, for the labeled source data $\{(\bm{x}_i^s, y_i^s)\}_{i=1}^{n_s}$, we simply replace the auxiliary distribution in Eq. \eqref{EqnAuxiTarDistrConstrain} with that formed by the ground-truth labels $\{y_i^s\}_{i=1}^{n_s}$, resulting in a supervised network training via cross-entropy minimization
\begin{eqnarray}\label{EqnDeepDisClustSrc}
\begin{aligned}
\min\limits_{\bm{\theta}_{\varphi},\bm{\theta}_f}  - \frac{1}{n_s} \sum_{i=1}^{n_s} \log p_{i, y_i^s}^s .
\end{aligned}
\end{eqnarray}
Supervised training enables us to learn a discriminative feature space defined by class labels, and we simply consider all source examples with the same label as a cluster, \ie, the clusters are semantically defined.

\keypoint{A Soft Selection of Source Examples.}
The \emph{class-wise closeness} assumed in the structural similarity across domains also suggests that, depending on the distances of different source instances to the target domain, their regularization effects may vary. This motivates us to weight different examples in $\{(\bm{x}_i^s, y_i^s)\}_{i=1}^{n_s}$ based on their respective distances to the corresponding target clusters, \ie, a strategy of soft sample selection. Similar ideas are adopted to address the issue of sample selection bias in \cite{kmm,density_estimate}. To this end, we first compute the $K$ cluster centroids $\{ {\bm{\mu}}_k^t \in \mathcal{Z} \}_{k=1}^K$ of target instances in the feature space $\mathcal{Z}$ via k-means clustering, where initial cluster assignments of target features $\{ \bm{z}_i^t \}_{i=1}^{n_t}$ are based on network predictions $\{ \bm{p}_i^t = f(\bm{z}_i^t) \}_{i=1}^{n_t}$ --- in other words, any $\bm{z}_i^t$ is assigned to the prediction of cluster $\hat{y}_i^t \in \{1, \dots, K\}$ when $p_{i, \hat{y}_i^t}^t$ is the largest element in $\bm{p}_i^t$. For any source example $(\bm{x}^s, y^s)$, we then compute its weight $w^s$ of soft selection based on the cosine similarity between the feature $\bm{z}^s$ and the target centroids $\bm{\mu}_{y^s}^t$ of cluster $y^s$, \ie,
\begin{eqnarray}\label{EqnCosSim}
\begin{aligned}
w^s = \frac{1}{2} \left(1 + \bm{\mu}_{y^s}^{t\top}\bm{z}^s / ( ||\bm{\mu}_{y^s}^t|| \, ||\bm{z}^s|| ) \right) \in [0, 1],
\end{aligned}
\end{eqnarray}
where we sum the cosine similarity with a constant $1$ in parenthesis to shift the weight in $[0,1]$.  Eq. (\ref{EqnCosSim}) gives the weights $\{w_i^s\}_{i=1}^{n_s}$ for source examples $\{(\bm{x}_i^s, y_i^s)\}_{i=1}^{n_s}$. We use $\{w_i^s\}_{i=1}^{n_s}$ to weight the loss (\ref{EqnDeepDisClustSrc}), giving rise to
\begin{eqnarray}\label{EqnDeepDisClustWeightedSrc}
\begin{aligned}
\min\limits_{\bm{\theta}_{\varphi},\bm{\theta}_f} {\cal{L}}_{f\circ\varphi}^s = - \frac{1}{n_s} \sum_{i=1}^{n_s} w_i^s \log p_{i, y_i^s}^s .
\end{aligned}
\end{eqnarray}
By doing so, we potentially improve the structural source regularization by better adapting to the intrinsic structures of target data. Note that we update the target cluster centroids $\{ {\bm{\mu}}_k^t \}_{k=1}^K$, and consequently the weights $\{ w_i^s \}_{i=1}^{n_s}$ computed by Eq. (\ref{EqnCosSim}), iteratively during network training (practically per training epoch), which makes the soft selection evolve with feature learning.

Combining the target clustering loss (\ref{EqnAuxiTarDistrConstrain}) with the source regularization (\ref{EqnDeepDisClustWeightedSrc}) gives our objective of Structurally Regularized deep Discriminative Clustering (SRDisC)
\begin{eqnarray}\label{EqnDeepDisClustAll}
\begin{aligned}
\min\limits_{\bm{Q}^t, \bm{\theta}_{\varphi}, \bm{\theta}_f} \mathcal{L}_{SRDisC} = \mathcal{L}_{f\circ\varphi}^t  + \lambda \mathcal{L}_{f\circ\varphi}^s,
\end{aligned}
\end{eqnarray}
where $\lambda$ is a penalty parameter.

\subsection{Modulating the Intrinsic Target Structures via Generative Learning with Self-Attentive Feature Interactions}
\label{SecSRGenC}

Features learned by the objective (\ref{EqnDeepDisClustAll}) of regularized discriminative clustering tend to be amenable to domain adaptation, since the objective ideally pushes source and target instances of the same classes/clusters into the respective same regions in the feature space $\mathcal{Z}$, where generative modeling of data distributions has not been taken in account yet. In this section, we aim to further modulate the feature space learning from a \emph{generative} perspective, by learning from the embedding $\varphi: \mathcal{X} \rightarrow \mathcal{Z}$ a set of $K$ cluster centroids common to the two domains \footnote{We emphasize that the generative learning of cluster centroids in Section \ref{SecSRGenC} is different from that in (\ref{EqnCosSim}) for computing the weights used for soft selection of source examples, where $\{ \bm{\mu}_k^t \}_{k=1}^K$ are the centroids obtained by k-means clustering of the target instances in the feature space, and the initial cluster assignments in k-means clustering are based on the pseudo-labels of target instances. Instead, the cluster centroids $\{ \bm{c}_k \}_{k=1}^K$ learned in the present section are common to the source and target domains, whose introduction is to enable a regularized generative clustering, and whose learning is conducted in a feed-forward manner. Refer to the main text for the details. }. Learning cluster centroids enables probabilistic assignments of data instances to clusters, which potentially enhances the uncovering of intrinsic data structures in a manner complementary to the discriminative one in Section \ref{SecSRDisC}. To enable feed-forward learning of cluster centroids, we technically rely on self-attentive feature interactions of data instances. Details are presented as follows.


\subsubsection{Deep Generative Target Clustering}
\label{SecSRGenCTarget}

Assume the availability of a parametric mapping $\phi: \mathcal{Z}^n \rightarrow \mathbb{R}^{d\times K}$, which learns from $n$ $d$-dimensional instance features in $\mathcal{Z}$ to produce a set of $K$ cluster centroids, denoted as $\{ \bm{c}_k \in \mathbb{R}^d \}_{k=1}^K$. We implement $\phi$ as a trainable feed-forward subnetwork parameterized by $\bm{\theta}_{\phi}$, as illustrated in Fig. \ref{fig:att_blocks}, whose details are presented shortly in Section \ref{SecSelfAttCentroidLearning}. To enable generative modeling of target data, we compute the following probability that softly assigns any $\bm{z}^t = \varphi(\bm{x}^t)$ of target instance to the cluster $k$
\begin{eqnarray}\label{EqnDeepGenAssignProb}
\begin{aligned}
\widetilde{p}_k^t = \frac{\exp((1 + ||\bm{z}^t - \bm{c}_k||^2)^{-1})}{\sum_{k'=1}^K \exp((1 + ||\bm{z}^t - \bm{c}_{k'}||^2)^{-1})} .
\end{aligned}
\end{eqnarray}
Eq. (\ref{EqnDeepGenAssignProb}) is a variant of Student t-distribution \cite{UnsupervisedEmbeddingICML16,t_sne} and we have $\widetilde{p}_k^t \rightarrow  e/(e + K -1) < 1$, which scales down with the increase of the cluster number $K > 1$ --- values of $\widetilde{p}_k^t$ are in fact very small for typical domain adaptation tasks that have more than a few dozens of classes. For ease of presentation, we compactly write the soft assignment probabilities of $\bm{z}^t$ to all the $K$ clusters as $\widetilde{\bm{p}}^t \in [0, 1]^K$, and write $\widetilde{\bm{P}}^t = \{ \widetilde{\bm{p}}_i^t \}_{i=1}^{n_t} $ for all the target instances.

Similar to the objective (\ref{EqnAuxiTarDistrConstrain}) of deep discriminative clustering, we introduce an auxiliary distribution $\widetilde{\bm{Q}}^t = \{ \widetilde{\bm{q}}_i^t \}_{i=1}^{n_t}$ to match $\widetilde{\bm{P}}^t$, giving rise to
\begin{eqnarray}\label{EqnAuxiTarDistrConstrainGenModeling}
\begin{aligned}
\min_{\widetilde{\bm{Q}}^t, \{\bm{\theta}_{\varphi},\bm{\theta}_{\phi}\}} \mathcal{L}_{\phi\circ\varphi}^t = {\rm KL}(\widetilde{\bm{Q}}^t||\widetilde{\bm{P}}^t) + \sum_{k=1}^K {\widetilde{\varrho}}_k^t \log {\widetilde{\varrho}}_k^t ,
\end{aligned}
\end{eqnarray}
where $\widetilde{\varrho}_k^t$ is defined similarly as ${\varrho}_k^t$ in (\ref{EqnAuxiTarDistrConstrain}). Optimization of (\ref{EqnAuxiTarDistrConstrainGenModeling}) is again conducted by alternating in updating $\widetilde{\bm{Q}}^t$, and using the updated $\widetilde{\bm{Q}}^t$ as labels to train the network $\phi\circ\varphi$ to update the parameters $\{\bm{\theta}_{\varphi},\bm{\theta}_{\phi}\}$, similar to the self-training strategies popularly used in recent methods \cite{dsbn,iCAN,dwt_mec}. It is obvious that the choice of auxiliary $\widetilde{\bm{Q}}^t$ is crucial for the learning success of (\ref{EqnAuxiTarDistrConstrain}); given that the cluster centroids $\{ \bm{c}_k \}_{k=1}^K$ are also involved in the learning, without any constraints, learning could end with less sensible solutions that would not model intrinsic structures of target data generatively. To remedy, we again rely on labeled source data to impose structural regularization, as presented shortly.

\subsubsection{Structural Source Regularization by Learning a Common Set of Cluster Centroids}
\label{SecSRGenCSrc}

Based on the \emph{class-wise closeness} assumed in Section \ref{SecProbDefinition}, we propose to regularize the generative target clustering (\ref{EqnAuxiTarDistrConstrainGenModeling}) by making the learning of cluster centroids $\{ \bm{c}_k \}_{k=1}^K$ common to the labeled source data $\{(\bm{z}_i^s, y_i^s)\}_{i=1}^{n_s}$. Similar to (\ref{EqnDeepGenAssignProb}), the probability of softly assigning any $\bm{z}^s$ to its ground-truth centroid $\bm{c}_{y^s}$ is written as
\begin{eqnarray}\label{EqnDeepGenAssignProbSrc}
\begin{aligned}
\widetilde{p}_{y^s}^s = \frac{\exp((1 + ||\bm{z}^s - \bm{c}_{y^s}||^2)^{-1})}{\sum_{k=1}^K \exp((1 + ||\bm{z}^s - \bm{c}_{k}||^2)^{-1})} .
\end{aligned}
\end{eqnarray}
Given the definition in (\ref{EqnDeepGenAssignProbSrc}), a cross-entropy loss can be imposed on $\{(\bm{z}_i^s, y_i^s)\}_{i=1}^{n_s}$, which is equivalent to maximizing the probability of assigning any $\bm{z}^s$ to its true cluster $y^s$. We also use the weights $\{w_i^s\}_{i=1}^{n_s}$ computed by (\ref{EqnCosSim}) to re-weight the loss contributions from individual source instances, resulting in
\begin{eqnarray}\label{EqnDeepGenClustWeightedSrc}
\begin{aligned}
\min\limits_{\bm{\theta}_{\varphi},\bm{\theta}_{\phi}} {\cal{L}}_{\phi\circ\varphi}^s = - \frac{1}{n_s} \sum_{i=1}^{n_s} w_i^s \log \widetilde{p}_{i, y_i^s}^s .
\end{aligned}
\end{eqnarray}
Regularizing the generative learning of $\{ \bm{c}_k \}_{k=1}^K$ via (\ref{EqnDeepGenClustWeightedSrc}) makes the optimization well conditioned, since cluster assignments of $\{ \bm{z}_i^s \}_{i=1}^{n_s}$ have been determined by their ground-truth labels $\{y_i^s\}_{i=1}^{n_s}$.

Combining (\ref{EqnAuxiTarDistrConstrainGenModeling}) with the source regularization (\ref{EqnDeepGenClustWeightedSrc}) gives our objective of Structurally Regularized deep Generative Clustering (SRGenC)
\begin{eqnarray}\label{EqnDeepGenClustAll}
\begin{aligned}
\min\limits_{\widetilde{\bm{Q}}^t, \bm{\theta}_{\varphi}, \bm{\theta}_{\phi}} \mathcal{L}_{SRGenC} = \mathcal{L}_{\phi\circ\varphi}^t  + \lambda \mathcal{L}_{\phi\circ\varphi}^s ,
\end{aligned}
\end{eqnarray}
where $\lambda$ is the penalty parameter that we take the same value as for (\ref{EqnDeepDisClustAll}).

\keypoint{Remarks.} Equations (\ref{EqnDeepGenAssignProb}) and (\ref{EqnDeepGenAssignProbSrc}) are based on a variant of Student t-distribution. The original Student t-distribution used in \cite{UnsupervisedEmbeddingICML16,t_sne} has the effect of preserving the distances between instances and cluster centroids when they are moderately dissimilar; our variant by converting it as an exponential function tends to magnify the effect. We empirically observe that for $\bm{z}^t$ with $\arg\max_{k'} \widetilde{p}_{k'}^t = k$ or $\bm{z}^s$ with its true label $y^s = k$, they are learned to drift together with its cluster centroid $\bm{c}_k$ in the feature space, rather than to collapse to $\bm{c}_k$. In some cases, the class-wise distances between the source and target domains are even getting larger by applying (\ref{EqnDeepGenAssignProb}) and (\ref{EqnDeepGenAssignProbSrc}), suggesting that the SRGenC objective (\ref{EqnDeepGenClustAll}) is indeed modulating the feature space learning via generative clustering, which is in contrast to existing methods \cite{dann,mcd,dan,can,pfan,cat} that explicitly align the features across the two domains. Results of these empirical studies are presented in Fig. \ref{fig:l2_dist}. 

\subsubsection{Learning Cluster Centroids via Self-Attentive Feature Interactions}
\label{SecSelfAttCentroidLearning}

\begin{figure}[!t]
		\centering
		\subfloat[Multihead Attention Block (MAB)]{
			\label{fig:mab}
			\includegraphics[height=0.18\linewidth]{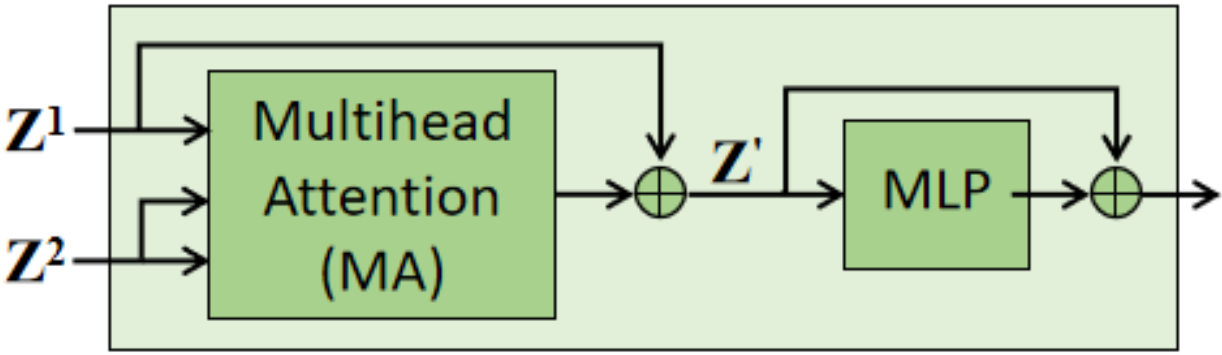}
		}\vspace{-0.3cm}
		\hspace{0.2in}
		\centering
		\subfloat[Cluster Centroid Learner]{
			\label{fig:ccl}
			\includegraphics[height=0.18\linewidth]{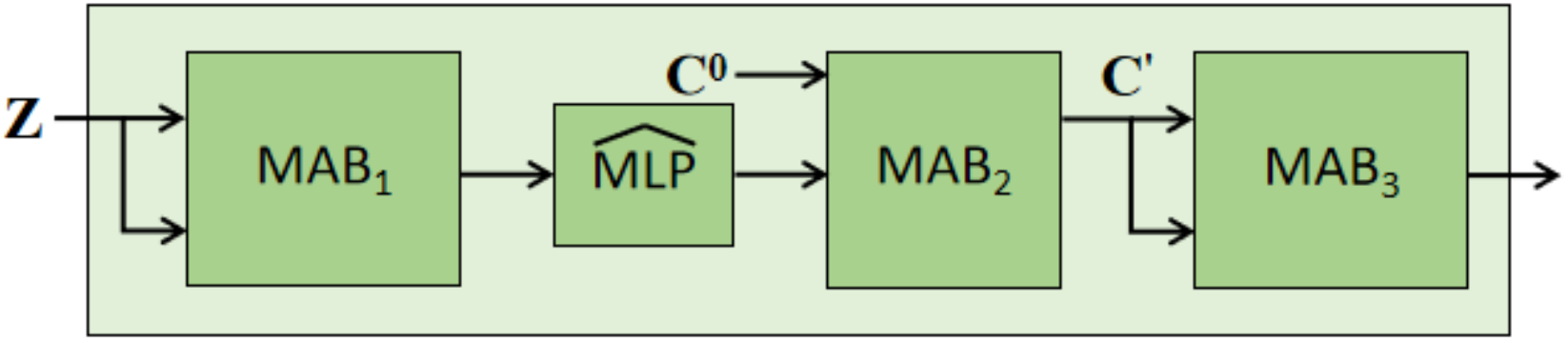}
		}\vspace{-0.3cm}
		\caption{An illustration of our used subnetwork for self-attentive learning of cluster centroids. One may refer to the main text for definitions of the used terms and math notations. 
		}
		\label{fig:att_blocks}\vspace{-0.3cm}
\end{figure}

We have so far assumed that the parametric mapping $\phi: \mathcal{Z}^n \rightarrow \mathbb{R}^{d\times K}$ has been given, which learns the $K$ cluster centroids $\{ \bm{c}_k \in \mathbb{R}^d \}_{k=1}^K$ in a feed-forward manner. Let $\bm{C} = [\bm{c}_1, \dots, \bm{c}_K] \in \mathbb{R}^{d\times K}$. Considering that the input to the learning of $\phi$ contains the set of features $\bm{Z} = [\bm{Z}^s, \bm{Z}^t] \in \mathbb{R}^{d\times n}$, with $\bm{Z}^s = [\bm{z}_1^s, \dots, \bm{z}_{n_s}^s] \in \mathbb{R}^{d\times n_s}$, $\bm{Z}^t = [\bm{z}_1^t, \dots, \bm{z}_{n_t}^t] \in \mathbb{R}^{d\times n_t}$, and $n = n_s + n_t$, we choose to adapt a framework of set transformer \cite{multi_head_attention,sab} for this purpose (\cf~\cite{sab} for its time complexity analysis during training). It uses a self-attention mechanism to learn from instance features in $\bm{Z}$ to produce $\bm{C}$. By encoding pairwise and even higher-order interactions among element features in $\bm{Z}$, the framework is beneficial to iterative clustering in a feed-forward deep network. Fig. \ref{fig:att_blocks} gives the illustration.

Technically, the framework relies on the following parametric mapping of Multihead Attention (MA) \cite{multi_head_attention}
\begin{eqnarray}\label{EqnMultiHeadAtt}
\begin{aligned}
\phi^{\rm MA} (\bm{Z}^1, \bm{Z}^2, \bm{Z}^3) = \bm{W}^{4\top} [\bm{Z}_1^{\downarrow}; \cdots; \bm{Z}_m^{\downarrow}] \qquad
\end{aligned}
\\
{\rm s.t.} \ \bm{Z}_i^{\downarrow} = \bm{W}_i^{3\top}\bm{Z}^3 \ \sigma\left( (\bm{W}_i^{1\top}\bm{Z}^1)^{\top} (\bm{W}_i^{2\top}\bm{Z}^2) / \sqrt{d}  \right) , \nonumber \\ i = 1, \dots, m  . \nonumber
\end{eqnarray}
Let $\bm{Z}^1 = \bm{Z}^2 = \bm{Z}^3 = \bm{Z}$. $\{ \bm{W}_i^1, \bm{W}_i^2, \bm{W}_i^3 \in \mathbb{R}^{d\times \frac{d}{m}} \}_{i=1}^m$ and $\bm{W}^4 \in \mathbb{R}^{d\times d}$ are the trainable parameters, and we collectively write $\bm{\theta}_{\phi^{\rm MA}} = \{ \{ \bm{W}_i^1, \bm{W}_i^2, \bm{W}_i^3 \}_{i=1}^m, \bm{W}^4\}$. We have the intermediate $\bm{Z}_i^{\downarrow} \in \mathbb{R}^{\frac{d}{m}\times n}$. Eq. (\ref{EqnMultiHeadAtt}) computes, after projecting the $d$-dimensional element features in $\bm{Z}$ into $d/m$ ones, the self similarities among the elements in $\bm{Z}$, and then uses the obtained similarities to re-mix the elements, where $\sigma(\cdot)$ is the softmax function that applies to the matrix-formed argument in a row-wise manner. The following parametric mapping of Multihead Attention Block (MAB) can be built upon the mapping (\ref{EqnMultiHeadAtt})
\begin{eqnarray}\label{EqnMultiHeadAttBlock}
\begin{aligned}
\phi^{\rm MAB} (\bm{Z}^1, \bm{Z}^2) = \bm{Z}' + \phi^{\rm MLP} (\bm{Z}')
\end{aligned}
\\
{\rm s.t.} \ \bm{Z}' = \bm{Z}^1 + \phi^{\rm MA} (\bm{Z}^1, \bm{Z}^2, \bm{Z}^2) , \nonumber
\end{eqnarray}
where $\phi^{\rm MLP}(\cdot)$ is a parametric mapping function implemented as a Multi-layer Perceptron (MLP), whose parameters are denoted as $\bm{\theta}_{\phi^{\rm MLP}}$. We write parameters of the MAB collectively as $\bm{\theta}_{\phi^{\rm MAB}} = \{\bm{\theta}_{\phi^{\rm MA}}, \bm{\theta}_{\phi^{\rm MLP}}\}$. Let $\bm{Z}^1 = \bm{Z}^2 = \bm{Z}$. Eq. (\ref{EqnMultiHeadAttBlock}) outputs a feature matrix of equal size, which contains the information of pairwise interactions among the elements in $\bm{Z}$. By stacking multiple such blocks, information about higher-order interactions can also be encoded.

We finally have the feed-forward function $\phi: \mathcal{Z}^n \rightarrow \mathbb{R}^{d\times K}$ for learning cluster centroids, defined as
\begin{eqnarray}\label{EqnACRLwSAB}
\begin{aligned}
\phi(\bm{Z}, \bm{C}^0) := \phi^{\rm MAB_3} (\bm{C}', \bm{C}') \qquad
\end{aligned}
\\
{\rm s.t.} \ \bm{C}' = \phi^{\rm MAB_2} \left( \bm{C}^0, \phi^{\widehat{\rm MLP}}( \phi^{\rm MAB_1} (\bm{Z}, \bm{Z}) ) \right) , \nonumber
\end{eqnarray}
where $\phi^{\widehat{\rm MLP}}(\cdot)$ denotes another MLP parameterized by $\bm{\theta}_{\phi^{\widehat{\rm MLP}}}$, and $\bm{C}^0 \in \mathbb{R}^{d\times K}$ contains $K$ initial learning seeds, which we initialize by sampling their $d$ entries from a normal distribution. 
Intuitively, individual seeds aggregate instance features by self-attentive interactions, and cluster centroids would be finally obtained with the learning. The intermediate $\bm{C}' \in \mathbb{R}^{d\times K}$ in (\ref{EqnACRLwSAB}) is already in the form of cluster centroids, and the final $\bm{C} = \phi(\bm{Z}, \bm{C}^0)$ by an additional MAB mapping may further improve the clustering purity, as discussed in \cite{sab}.

Overall, our subnetwork $\phi$ of learning cluster centroids via self-attentive feature interactions is parameterized by
$\bm{\theta}_{\phi} = \{ \bm{\theta}_{\phi^{\rm MAB_1}}, \bm{\theta}_{\phi^{\rm MAB_2}}, \bm{\theta}_{\phi^{\rm MAB_3}}, \bm{\theta}_{\phi^{\widehat{\rm MLP}}} \}$.
Note that this auxiliary subnetwork is only used during training, which helps modulate the learning of feature space $\mathcal{Z}$ via the SRGenC objective (\ref{EqnDeepGenClustAll}); during testing, it is discarded and only the network $f\circ\varphi$ is used for inference.


\begin{algorithm}[!t]
	\caption{H-SRDC 
	}
	\label{alg:new_method}
	\begin{scriptsize}
	\begin{algorithmic}[1]
		\Require Labeled source data $(\bm{X}^s, \bm{Y}^s) = \{(\bm{x}_i^s, y_i^s)\}_{i=1}^{n_s}$, and unlabeled target data $\bm{X}^t = \{\bm{x}_i^t\}_{i=1}^{n_t}$
		\Ensure Domain-adapted model $f\circ\varphi$
		\State Initialize the weights $\{w_i^s=1\}_{i=1}^{n_s}$; initialize $\{\{q_{i,k}^t=\widetilde{q}_{i,k}^t={\rm I}[k=\hat{y}_i^t]\}_{k=1}^K\}_{i=1}^{n_t}$
		, where ${\rm I}[\cdot]$ is an indicator function and $\hat{y}_i^t$ is the cluster assignment by k-means clustering in the feature space $\mathcal{Z}$; $E=1$
		\While {$MAX\_EPOCH$ is not reached}
		\While {$MAX\_ITERATION$ is not reached}
		\State Sample batch data of $(\bm{X}_{batch}^s, \bm{Y}_{batch}^s)$ and $\bm{X}_{batch}^t$
		\State Extract features $\bm{Z}_{batch}^s$ and $\bm{Z}_{batch}^t$ using current $\varphi(\cdot)$
		\State Compute probability predictions $\bm{P}_{batch}^s \!$ and $\bm{P}_{batch}^t \!$ using current $f(\cdot)$
		\State Generate centroids $\bm{C} = \{\bm{c}_k\}_{k=1}^K$ using current $\phi(\cdot)$
		\State Compute probabilities $\widetilde{\bm{P}}_{batch}^s$ and $\widetilde{\bm{P}}_{batch}^t$ respectively by \eqref{EqnDeepGenAssignProb} and \eqref{EqnDeepGenAssignProbSrc}
		\If{$E>1$}
		\State Update the auxiliary distributions $\bm{Q}_{batch}^t$ and $\widetilde{\bm{Q}}_{batch}^t$ by \eqref{EqnAuxiTarDistrCloseSol_DisClust}
		\EndIf
		\State Evaluate the objective \eqref{EqnDeepJointClust} and compute gradients to update the network parameters $\bm{\theta}$
		\EndWhile
		\State Update the weights $\{w_i^s\}_{i=1}^{n_s}$ by \eqref{EqnCosSim}
		\State $E=E+1$
		\EndWhile
	\end{algorithmic}
	\end{scriptsize}
\end{algorithm}

\begin{figure*}[!t]
	\begin{center}
		\includegraphics[width=0.9\textwidth]{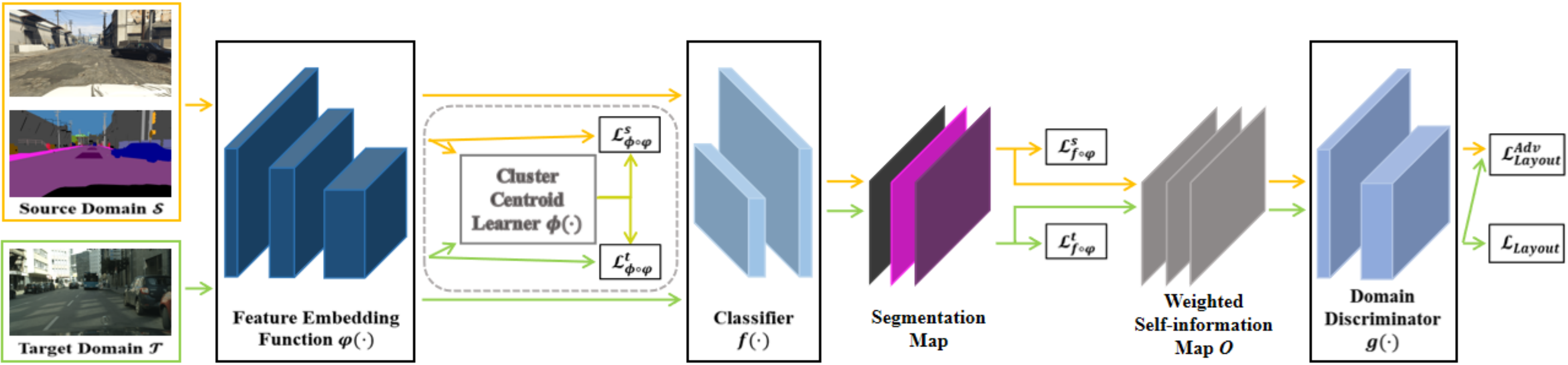}
	\end{center}
	\vskip -0.5cm
	\caption{A schematic illustration of extending H-SRDC for image semantic segmentation in an unsupervised domain adaptation setting. A task-related notion of \emph{layout-wise consistency} is introduced, which inspires the design of additional loss terms for adaptation of segmentation maps in an adversarial training manner. Once trained, the segmentation model $f\circ \varphi$ is deployed for the UDA task. In this figure, orange and green arrows represent the data flows from the source and target domains, respectively.
	}
	\label{fig:seg_pipeline}\vspace{-0.4cm}
\end{figure*}

\subsection{Overall Training and Inference}

The overall training objective combines (\ref{EqnDeepDisClustAll}) for structurally regularized discriminative clustering and (\ref{EqnDeepGenClustAll}) for the generative counterpart, resulting in a hybrid model of Structurally Regularized Deep Clustering. We term the method as H-SRDC to emphasize both its hybrid nature of regularized discriminative and generative clusterings, and the extension to the method of SRDC proposed in our preliminary version \cite{srdc}. By writing the network parameters collectively as $\bm{\theta}=\{\bm{\theta}_{\varphi}, \bm{\theta}_f, \bm{\theta}_{\phi}\}$, we have
\begin{eqnarray}\label{EqnDeepJointClust}
\begin{aligned}
\min\limits_{\bm{Q}^t, \widetilde{\bm{Q}}^t, \bm{\theta}} \mathcal{L}_{H-SRDC} = \mathcal{L}_{SRDisC} + \mathcal{L}_{SRGenC}.
\end{aligned}
\end{eqnarray}
Note that the objective (\ref{EqnDeepJointClust}) is applied to the whole network (\cf~Fig. \ref{fig:pipeline}) in an end-to-end fashion, which learns all the parameters in $\bm{\theta}$ simultaneously. We summarize the training process of H-SRDC using stochastic gradient descent in Algorithm \ref{alg:new_method} (\cf~Appendix B for more details). During inference, the trained network is used to classify any testing instance on the target domain via a simple forward pass of $f\circ\varphi$.



\section{An Extension for Domain-Adapted Semantic Segmentation}

In this section, we present an extension of our proposed H-SRDC for the task of semantic segmentation of images in a UDA setting. Fig. \ref{fig:seg_pipeline} illustrates the modified network architecture. The task is by nature to classify each pixel in an input image into one of multiple semantic classes. To facilitate the discussion, we inherit most of the math notations used in the previous sections, and override some of them when the context requires. Assume the input image is of the size $h\times w$, we now denote the labeled source data as $\{ \{ \bm{x}_{i,j}^s \}_{j=1}^{hw}, \{ y_{i,j}^s \}_{j=1}^{hw} \}_{i=1}^{n_s}$, and the unlabeled target ones as $\{ \{ \bm{x}_{i,j}^t \}_{j=1}^{hw} \}_{i=1}^{n_t}$; in other words, each instance $\bm{x}$ now represents the observations at a pixel, and a set $\{ \bm{x}_j \}_{j=1}^{hw}$ represents an image. As illustrated in Fig. \ref{fig:seg_pipeline}, our feature extractor $\varphi$ reduces the input resolution of $h\times w$ via multiple layers of convolution and pooling, resulting in $d$ feature maps of the size $h/a\times w/a$ for each input image, where $a$ is the ratio of subsampling, and we denote $h^a = h/a$ and $w^a = w/a$; for simplicity, we assume that they have no rounding issue. The feature maps are passed through the classifier $f$ and upsampled via bilinear interpolation to have the network output of the size $h\times w\times K$, \ie, one segmentation map per class of the total $K$ classes. We correspondingly write $\{ \{ \bm{p}_{i,j}^s = f\circ\varphi(\bm{x}_{i,j}^s) \in [0, 1]^K \}_{j=1}^{hw}  \}_{i=1}^{n_s}$ for the source data, and the same applies to the target ones; we also write $p_{i,j,k}$ for the $k^{th}$ element of $\bm{p}_{i, j}$.

\begin{figure}[!t]
	\begin{center}
		\includegraphics[width=0.5\linewidth]{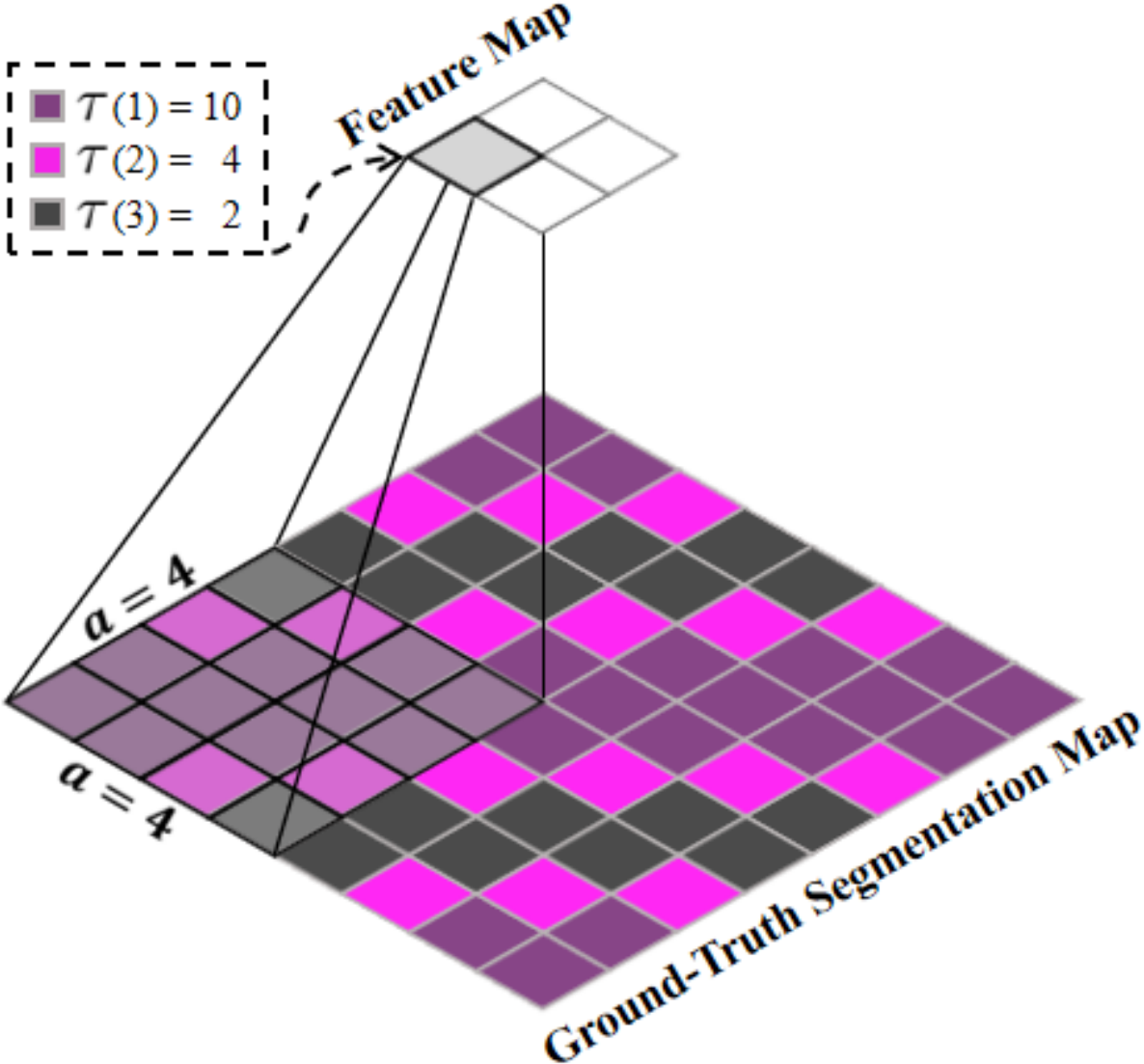}
	\end{center}
	\vskip -0.6cm
	\caption{An illustration of our used weighted combination scheme in the loss term (\ref{EqnDeepGenClustWeightedSrc4Seg}) for learning cluster centroids in the feature maps of reduced resolution, which receive supervision signals from the labeled source segmentation maps of original resolution.
	}
	\label{fig:seg_srgenc_srcpart}\vspace{-0.4cm}
\end{figure}

Given these definitions, the component of SRDisC objective (\ref{EqnDeepDisClustAll}) in H-SRDC can be readily applied. To apply the component of SRGenC (\ref{EqnDeepGenClustAll}) that operates in the feature space $\mathcal{Z}$, a subtle issue is that $\mathcal{Z}$ is now defined on the feature maps of reduced resolution, which is incompatible with the original resolution of segmentation maps on which the ground-truth labels of source data are defined. To address this issue, we note that during back-propagation of network training, each pixel located at the feature maps receives supervision signals from a field of the size $a\times a$ in the output segmentation maps; in the extreme case, all the $K$ classes may appear in such an $a\times a$ local field. We thus propose a weighted combination scheme that enables receiving supervision from any class that appears in such a local receptive field, by overriding the structural source regularization term ${\cal{L}}_{\phi\circ\varphi}^s$ (\ref{EqnDeepGenClustWeightedSrc}) in the SRGenC objective (\ref{EqnDeepGenClustAll}) as follows
\begin{eqnarray}\label{EqnDeepGenClustWeightedSrc4Seg}
\begin{aligned}
\min\limits_{\bm{\theta}_{\varphi},\bm{\theta}_{\phi}} {\cal{L}}_{\phi\circ\varphi}^s = - \frac{1}{n_sh^aw^a} \sum_{i=1}^{n_s} \sum_{j=1}^{h^aw^a} w_{i,j}^s \sum_{k=1}^K \frac{\tau_{i,j}(k)}{a^2} \log \widetilde{p}_{i, j, k}^s ,
\end{aligned}
\end{eqnarray}
where $\tau_{i,j}(k)$ counts the pixel number of the $k^{th}$ class in a local $a\times a$ field in the ground-truth segmentation map, which corresponds to the $j^{th}$ pixel location in feature maps of the $i^{th}$ training source image. Fig. \ref{fig:seg_srgenc_srcpart} illustrates the scheme. 

Our proposed H-SRDC is motivated from the assumption of \emph{structural similarity} between the source and target domains, which includes the notions of \emph{domain-wise discrimination} and \emph{class-wise closeness} specified in Section \ref{SecMotivation}. 
For the task of adapting semantic segmentation across domains, the output is spatially structured. Inspired by this, a basic strategy exists that accounts for the spatial segmentation layout of the structured output maps \cite{Adapt_SegMap,advent,curriculum_da} (\cf~Appendix C for more insights). 
To uncover the intrinsic data structures at both spatial and pixel levels, we should also account for the domain similarity in terms of the structural dependency between local semantics (\ie, semantic layout), since the spatial and pixel-level distributions interact with each other. 
We thus add a third notion into our structural similarity assumption, as follows.
\begin{itemize}
\item \emph{Layout-wise consistency} assumes that the spatial layout of semantic segmentation maps is consistent between the source and target domains.
\end{itemize}
Layout-wise consistency acts as a spatial regularization that constrains the pixel-level distribution and thus effectively reduces the search space, which is not covered by the main UDA assumption of domain-wise discrimination and class-wise closeness. 
To implement this notion into our learning objective, we first define a weighted self-information map $\bm{O} = \{\bm{o}_j \in \mathbb{R}^K \}_{j=1}^{hw}$ for any input image $\{\bm{x}_j\}_{j=1}^{hw}$, which follows \cite{advent}; the $k^{th}$ element of any $\bm{o}_j$ in an $i^{th}$ image is computed as 
\begin{equation}\label{EqnSelfInfoComp} 
o_{i, j, k} = - p_{i, j, k}\log p_{i, j, k} .
\end{equation}
Thus $\bm{o}_{i, j}$ represents the entropy of assigning the pixel $\bm{x}_{i, j}$ to the $K$ classes. We implement the notion of layout-wise consistency by statistically enforcing the consistency between the distributions of $\{ \bm{O}_i^s \}_{i=1}^{n_s}$ and $\{ \bm{O}_i^t \}_{i=1}^{n_t}$. Technically, there exist many quantities that measure the first- or high-order statistics between the distributions, such as maximum mean discrepancy (MMD) \cite{mmd} and central moment discrepancy (CMD) \cite{cmd}; minimizing such quantities would promote the distribution consistency. In this work, we follow the adversarial training strategy used in the state-of-the-art methods \cite{Adapt_SegMap,advent}. Define $g: \mathbb{R}^{K\times hw} \rightarrow [0, 1]$, parameterized by $\bm{\theta}_g$, as the domain discriminator that classifies the source $\{ \bm{O}_i^s \}_{i=1}^{n_s}$ as positive and $\{ \bm{O}_i^t \}_{i=1}^{n_t}$ otherwise. We implement $g$ as a convolutional subnetwork that applies to any map $\bm{O}$, which is followed by a sigmoid layer for binary classification. 
Since the spatial layout of the source $\{ \bm{O}_i^s \}_{i=1}^{n_s}$ is determined by the ground-truth segmentation maps, we use the following objective to enforce consistency of the target layout to the source one
\begin{eqnarray}\label{EqnLayoutConsistencyLoss}
\begin{aligned}
\min\limits_{\bm{\theta}_{\varphi}, \bm{\theta}_f} {\cal{L}}_{Layout} = - \frac{1}{n_t} \sum_{i=1}^{n_t} \log g(\bm{O}_{i}^t),
\end{aligned}
\end{eqnarray}
where elements of any $\bm{O}_{i}^t$ are computed by (\ref{EqnSelfInfoComp}), which involves network predictions from the function $f\circ \varphi$. We use the following objective to account for an adversarial training
\begin{eqnarray}\label{EqnLayoutConsistencyLossAdv}
\begin{aligned}
\min\limits_{\bm{\theta}_g} {\cal{L}}_{Layout}^{Adv} \! = \! - \frac{1}{n_s} \sum_{i=1}^{n_s} \log g(\bm{O}_{i}^s) \! - \frac{1}{n_t} \sum_{i=1}^{n_t} \log (1 \! - \! g(\bm{O}_{i}^t)).
\end{aligned}
\end{eqnarray}

Combining the objective (\ref{EqnDeepJointClust}) of H-SRDC with the above (\ref{EqnLayoutConsistencyLoss}) and (\ref{EqnLayoutConsistencyLossAdv}) gives our overall objective when applying the H-SRDC method to semantic segmentation
\begin{eqnarray}\label{EqnSRDCPP4Seg}
\begin{aligned}
\min\limits_{\bm{Q}^t, \widetilde{\bm{Q}}^t, \bm{\theta}} \max\limits_{\bm{\theta}_g} \mathcal{L}_{H-SRDC}^{Seg} = \mathcal{L}_{SRDisC} + \mathcal{L}_{SRGenC} \qquad \\+ \beta {\cal{L}}_{Layout} - \beta {\cal{L}}_{Layout}^{Adv} ,
\end{aligned}
\end{eqnarray}
where $\beta$ is a penalty parameter, and note that the regularization term ${\cal{L}}_{\phi\circ\varphi}^s$ in $\mathcal{L}_{SRGenC}$ has been overrode by (\ref{EqnDeepGenClustWeightedSrc4Seg}).


\section{Experiments of Domain Adaptation in an Inductive Setting}
\label{SecIndExp}

In this work, we verify the efficacy of our proposed H-SRDC on two UDA tasks of image classification and semantic segmentation. In the UDA literature, the former task is mainly conducted in a transductive setting, where classification of the unlabeled target data is achieved together with the model learning, and the latter one is mainly conducted in an inductive setting, where the learned model is to be applied to a held-out test set of target instances sampled from the same target domain. In this section, we first present experiments in the inductive setting for the tasks of image classification and semantic segmentation; to the best of our knowledge, we are the first to report comprehensive experiments in the inductive setting on benchmark UDA datasets of image classification. Our experiments of image classification in the transductive setting are presented in Section \ref{SecTransExps}.
 
\keypoint{Datasets.} 
We use the following five UDA benchmarks for our experiments of image classification. 
\textbf{\em Office-31} \cite{office31} contains $4,110$ images of $31$ classes shared by three distinct domains, namely, Amazon (\textbf{A}), Webcam (\textbf{W}), and DSLR (\textbf{D}); they define six adaptation tasks by pair-wise domain combination.
\textbf{\em ImageCLEF-DA} \cite{imageclefda} is a benchmark containing $600$ images of $12$ classes per data domain of the total three domains, namely, Caltech-256 (\textbf{C}), ImageNet ILSVRC 2012 (\textbf{I}), and Pascal VOC 2012 (\textbf{P}); they again define six adaptation tasks by pair-wise domain combination. 
\textbf{\em Office-Home} \cite{officehome} is a more challenging benchmark, containing around $15,500$ images of $65$ classes shared by four distinct domains, namely, Art (\textbf{Ar}), Clipart (\textbf{Cl}), Product (\textbf{Pr}), and Real-World (\textbf{Rw}); they define 12 adaptation tasks by pair-wise domain combination. 
\textbf{\em VisDA-2017} \cite{visda2017} is a dataset for the difficult task of synthetic-to-real transfer (\textbf{Synthetic}$\to$\textbf{Real}); it has images of $12$ classes, including $152,397$ synthetic ones and $55,388$ natural ones.
The benchmark of \textbf{\em Digits} includes three $10$-class digit datasets of \textbf{\em SVHN (S)} \cite{svhn}, \textbf{\em MNIST (M)} \cite{mnist}, and \textbf{\em USPS (U)} \cite{usps};  
SVHN has colored, extremely blurred images of real-world digits, MNIST has grayscale images of digits on clean background, and USPS has grayscale images with digits written in unconstrained styles; they define three conventional tasks of \textbf{S}$\rightarrow$\textbf{M}, \textbf{M}$\rightarrow$\textbf{U}, and \textbf{U}$\rightarrow$\textbf{M}.

We use the following three datasets for our experiments of semantic segmentation. 
\textbf{\em GTA5} \cite{gta5} includes $24,966$ synthetic images from the computer game of Grand Theft Auto V. \textbf{\em SYNTHIA} \cite{synthia} produces $9,400$ synthetic images by rendering a virtual city using Unity engine. \textbf{\em Cityscapes} \cite{cityscapes} captures real street scenes, including images of $2,975$ training ones, $500$ validation ones, and $1,525$ test ones.
We follow the literature and evaluate two common adaptation tasks of \textbf{GTA5} $\rightarrow$\textbf{Cityscapes} and \textbf{SYNTHIA} $\rightarrow$\textbf{Cityscapes}.

\begin{figure*}[!t]
	\centering
	\subfloat[\footnotesize Instance-to-Centroid (\textbf{A}$\rightarrow$\textbf{D}) ]{
		\begin{minipage}[t]{0.33\textwidth}
			\centering
			\includegraphics[height=1.5in]{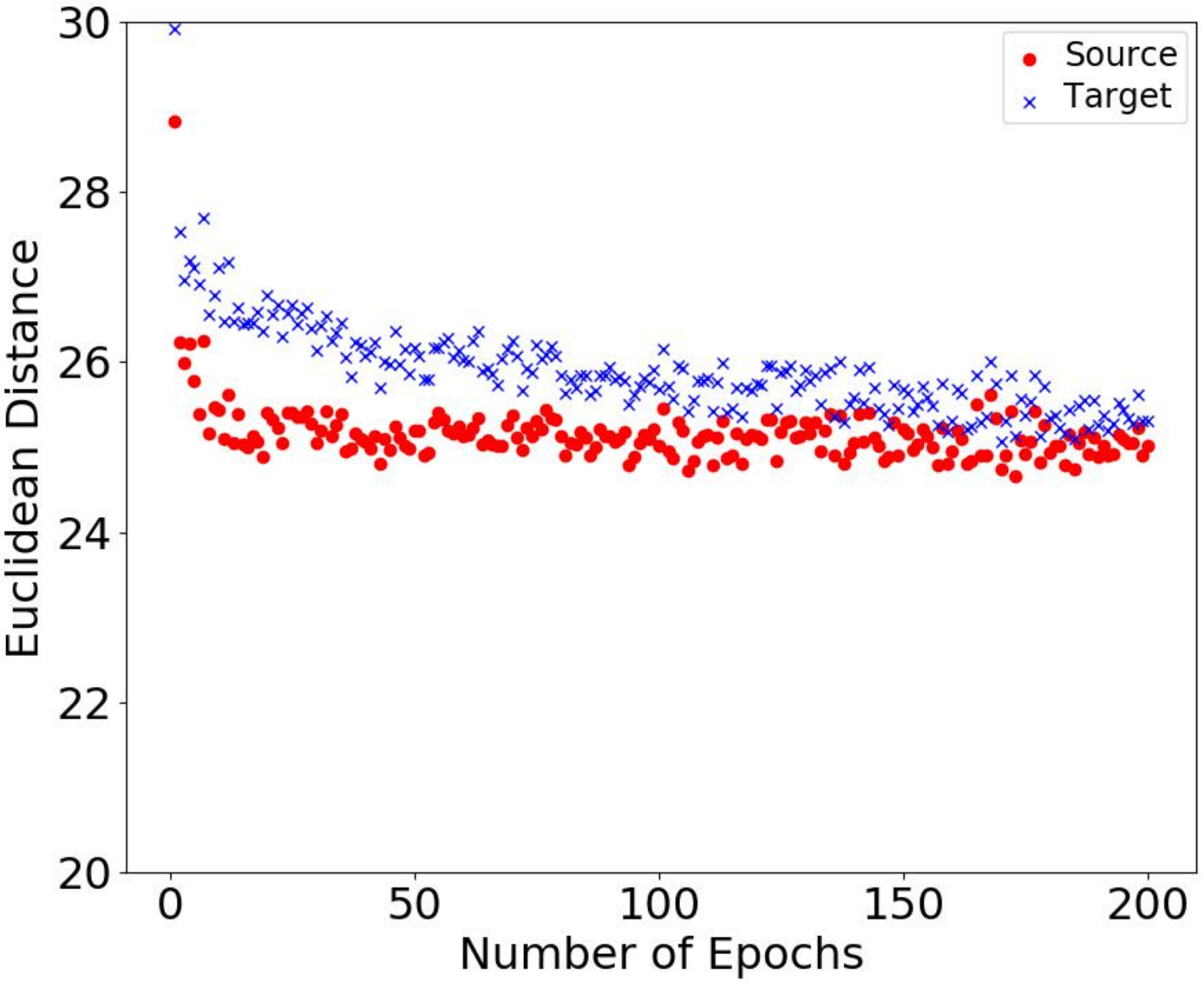}
			\label{fig:l2_dist:subfig1}
		\end{minipage}
	}%
	\subfloat[\footnotesize InsMean-to-Centroid (\textbf{A}$\rightarrow$\textbf{D})]{
		\begin{minipage}[t]{0.33\textwidth}
			\centering
			\includegraphics[height=1.5in]{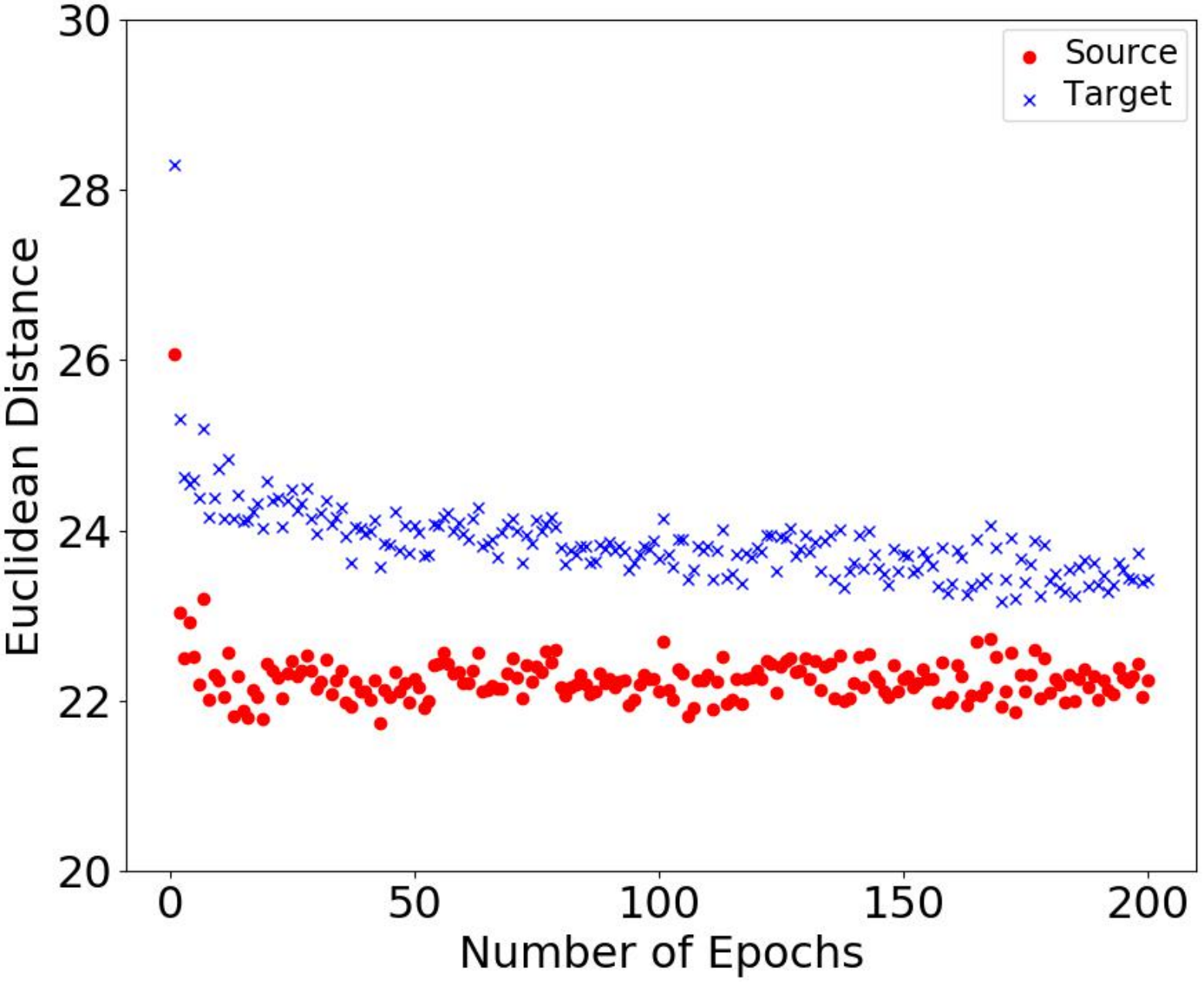}
			\label{fig:l2_dist:subfig2}
		\end{minipage}
	}%
	\subfloat[\footnotesize SrcInsMean-to-TgtInsMean (\textbf{A}$\rightarrow$\textbf{D})]{
		\begin{minipage}[t]{0.33\textwidth}
			\centering
			\includegraphics[height=1.5in]{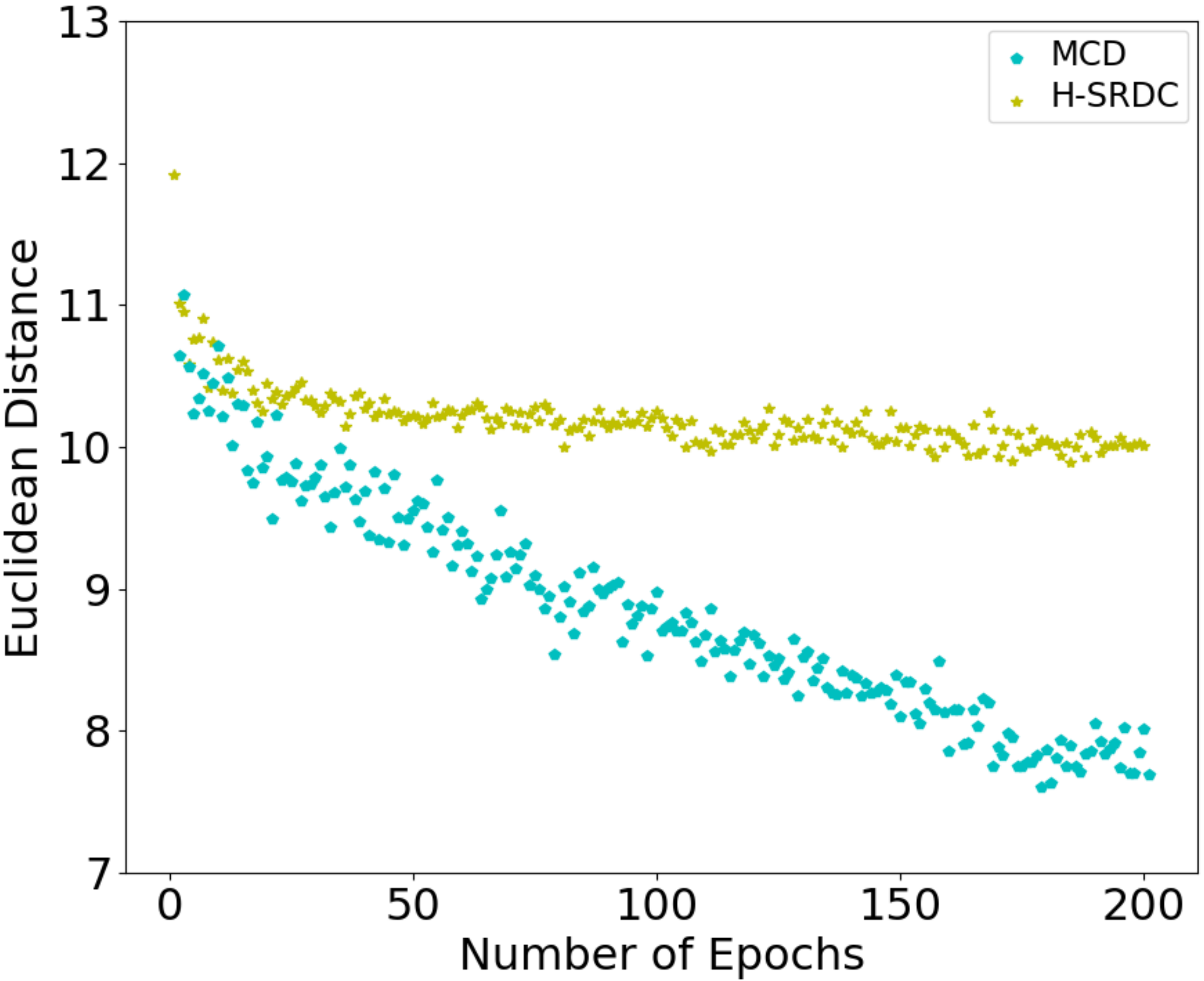}
			\label{fig:l2_dist:subfig3}
		\end{minipage}
	}%
	\\
	\subfloat[\footnotesize Instance-to-Centroid (\textbf{D}$\rightarrow$\textbf{A})]{
		\begin{minipage}[t]{0.33\textwidth}
			\centering
			\includegraphics[height=1.5in]{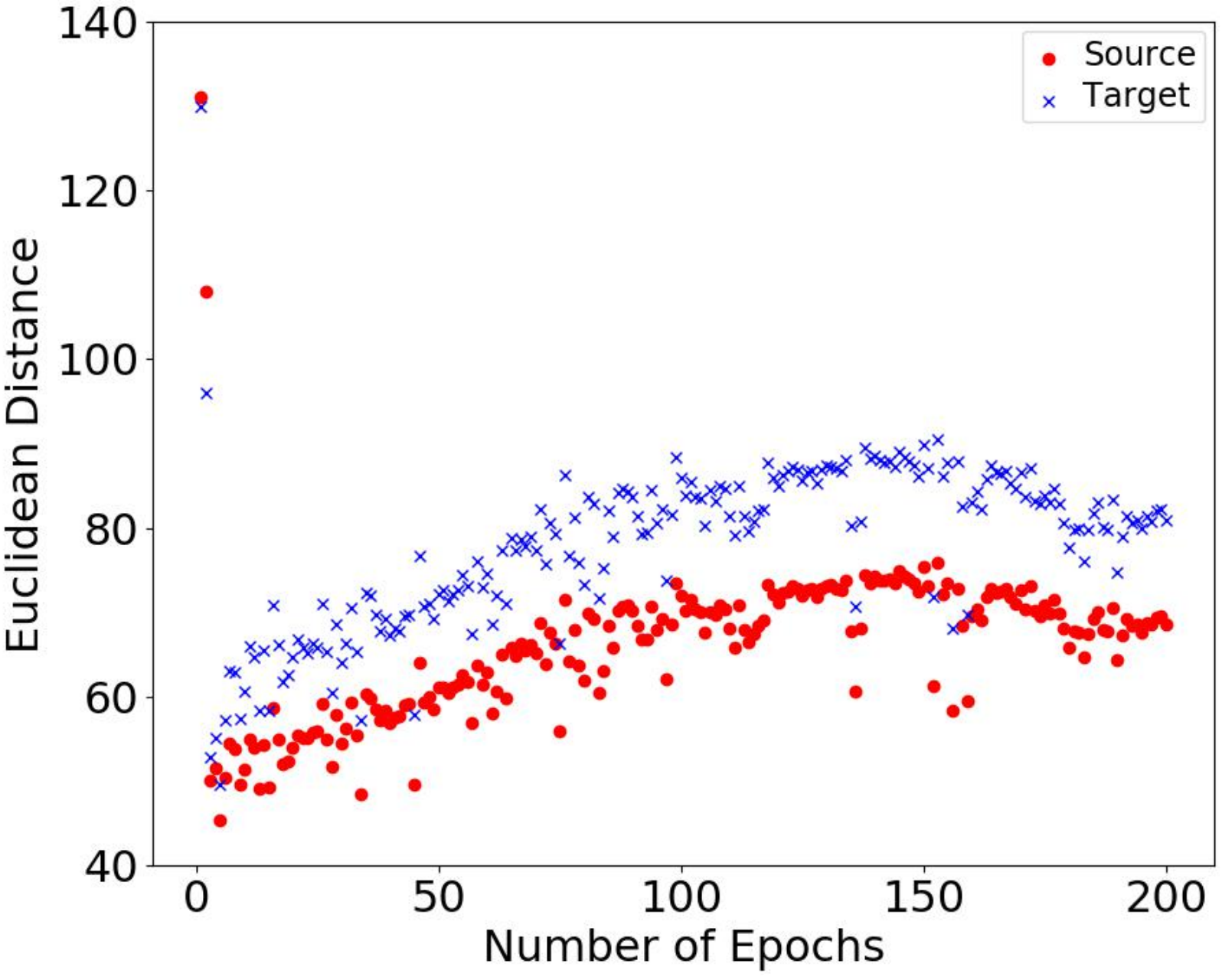}
			\label{fig:l2_dist:subfig7}
		\end{minipage}
	}%
	\subfloat[\footnotesize InsMean-to-Centroid (\textbf{D}$\rightarrow$\textbf{A})]{
		\begin{minipage}[t]{0.33\textwidth}
			\centering
			\includegraphics[height=1.5in]{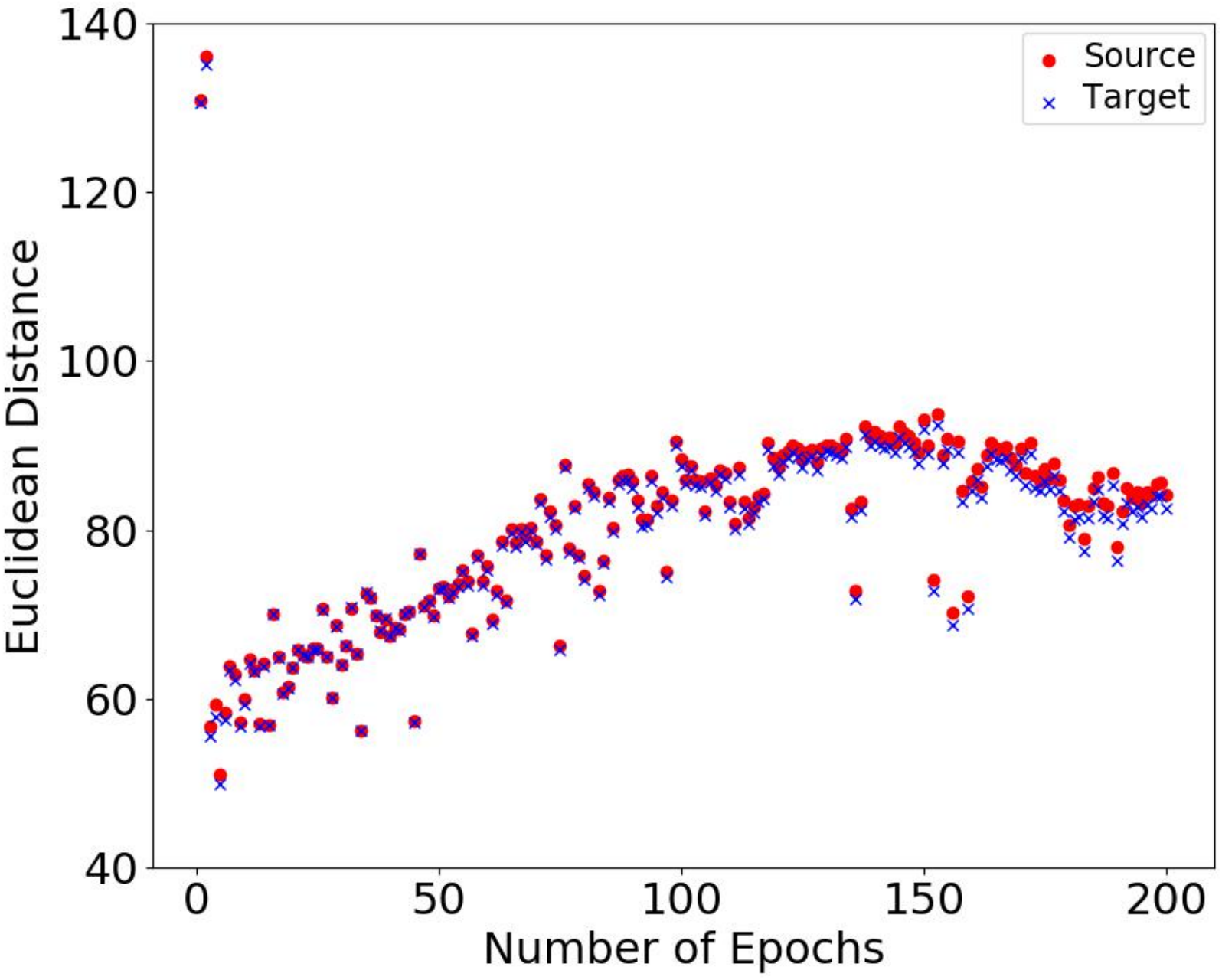}
			\label{fig:l2_dist:subfig8}
		\end{minipage}
	}%
	\subfloat[\footnotesize SrcInsMean-to-TgtInsMean (\textbf{D}$\rightarrow$\textbf{A})]{
		\begin{minipage}[t]{0.33\textwidth}
			\centering
			\includegraphics[height=1.5in]{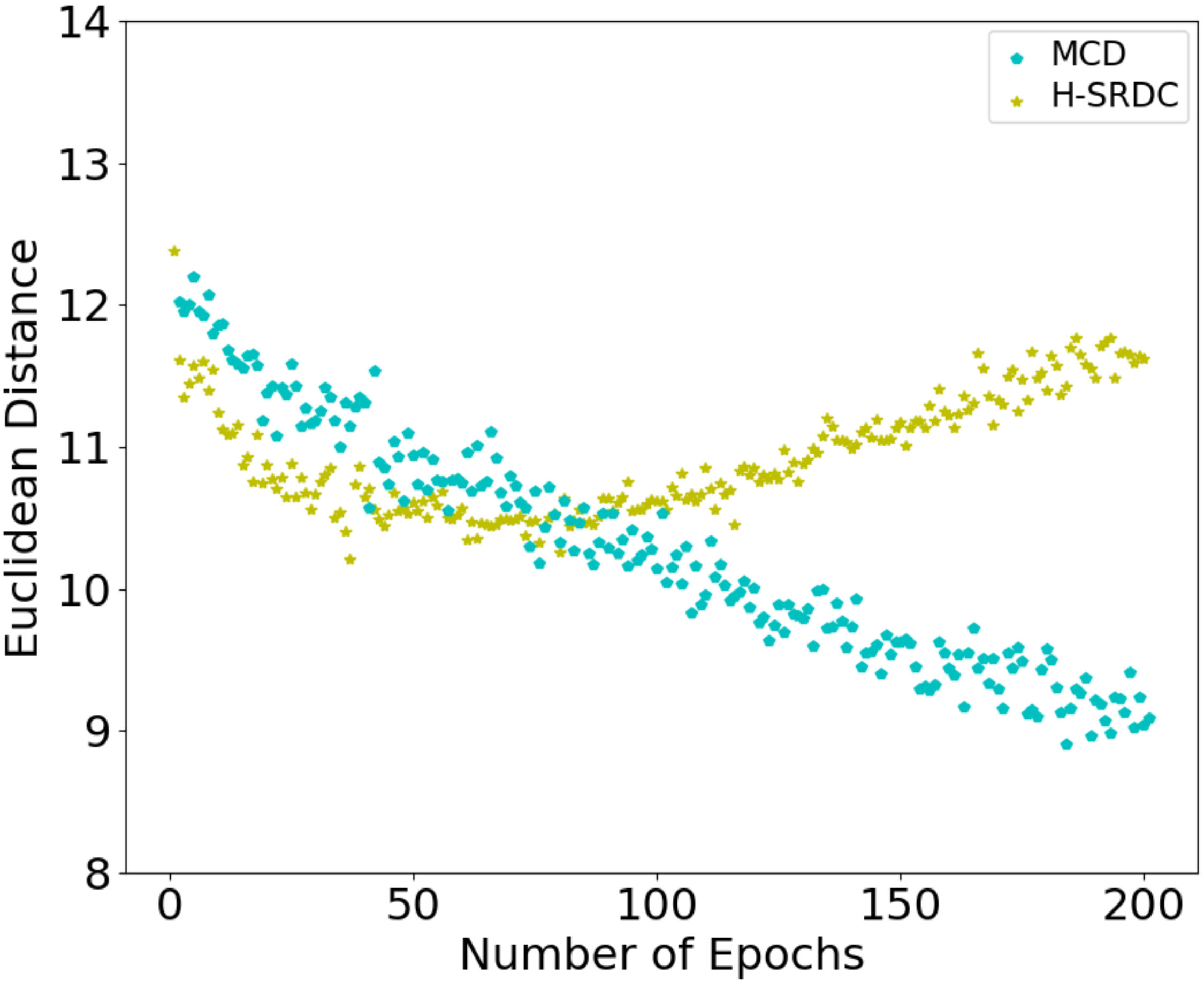}
			\label{fig:l2_dist:subfig9}
		\end{minipage}
	}%
	\vskip -0.2cm
	\caption{Learning diagnosis on the effect of the SRGenC objective (\ref{EqnDeepGenClustAll}) used in H-SRDC. Three types of distances for the source and target data are plotted against the training epochs. Comparisons between our H-SRDC and MCD \cite{mcd} are made in the two figures of last column. The experiments are conducted on the adaptation tasks of \textbf{A}$\to$\textbf{D} and \textbf{D}$\to$\textbf{A} on the Office-31 benchmark \cite{office31}. Refer to the main text for how these distances are defined and computed. 
	}
	\label{fig:l2_dist}\vspace{-0.3cm}
\end{figure*}

\keypoint{Settings and Implementation Details.} We first present our settings and implementation details for experiments of image classification. For all adaptation tasks in each benchmark, we use all the data on the source domain as the training ones, and make a random, half-half splitting of training and test data for samples of each class on the target domain; the data settings are fixed once prepared.
Note that for UDA in the inductive setting, each model is trained on the labeled source data and the training set of unlabeled target data, and is then evaluated on the target test set. 
To implement our H-SRDC for image classification, we use ImageNet \cite{imagenet} pre-trained ResNet-50 \cite{resnet}, with two newly added fully-connected (FC) layers respectively of $512$ and $K$ (\ie, the number of classes) neurons; the lower layers of the network are used as the feature extractor $\varphi(\cdot)$, and the two upper FC layers are used as the classifier $f(\cdot)$. The cluster centroid learner $\phi(\cdot)$ has three MABs (\cf~Fig. \ref{fig:att_blocks}), where we set the head number $m=4$ and each MLP has three FC layers with ReLU non-linearity. Setting of the penalty $\lambda$ in (\ref{EqnDeepDisClustAll}) and (\ref{EqnDeepGenClustAll}) follows the rule $\lambda_i = 2(1+\exp(-\gamma i))^{-1} - 1$, where $i$ is the epoch index normalized to $[0,1]$ and $\gamma=10$; the rule scales the values of $\lambda$ up from $0$ to $1$ during the training and gradually switches the learning from the labeled source data to the unlabeled target ones, \ie, $\lambda$ is empirically multiplied to losses on target data to suppress noises.
We perform regularized deep clustering in a 2048-dimensional feature space (\ie, $d=2048$).
Initializations of the weights $\{w_i^s\}_{i=1}^{n_s}$ for soft selection of source samples and those of the auxiliary distributions $\bm{q}^t$ and $\widetilde{\bm{q}}^t$ are given in Algorithm \ref{alg:new_method}.
When learning cluster centroids via the feed-forward function $\phi$, we apply Batch Whitening \cite{dwt_mec} to the respective feature batches of the two domains, which projects them into the same spherical distribution.
We use SGD to train the feature extractor $\varphi$ and classifier $f$, where the pre-trained layers are fine-tuned and the newly added FC layers are trained from scratch; learning rates for the FC layers follow $\eta_i = \eta_0(1+\alpha i)^{-\gamma}$, where $i$ is the epoch index normalized to $[0,1]$ and $\eta_0=0.01, \alpha=10, \gamma=0.75$, and those for the pre-trained layers are set as one tenth of the above schedule; we set the momentum, weight decay, batch size, and number of training epochs respectively as $0.9$, $0.0001$, $64$, and $200$. The cluster centroid learner $\phi$ is trained by Adam optimizer \cite{adam} with default parameters.
We use data augmentations of random region crop and horizontal flip and perform three random trials during training; during testing, we report the averaged classification result of center region crops on all test images.
For VisDA-2017, we follow the settings used in \cite{mcd}: we set $\lambda = 1$, the initial learning rate $\eta_0 = 0.001$, and train for 20 epochs; we report per-class, mean accuracy on $12$ classes of the dataset.
For Digits, we follow the settings used in \cite{mcd}: we adopt its same base network of a LeNet \cite{lenet}, and use Adam optimizer \cite{adam} with learning rate $0.0002$ for all network parameters of $\varphi$, $f$, and $\phi$, where we set the batch size as 128 and train for 200 epochs; we use the standard train/test split of each dataset, and do not use data augmentations; we report the averaged classification result over five random trials on test data.

For experiments of semantic segmentation, we follow the standard protocol where the validation set of Cityscapes is used for evaluation \cite{Adapt_SegMap,advent,clan}.
We report per-class IoU and mIoU on $19$ and $16$ classes respectively shared by GTA5 and SYNTHIA with Cityscapes.
We use ResNet-101 based DeepLab-v2 \cite{deeplab_v2} as the base network, where we set the subsampling ratio $a = 8$. We use the same domain discriminator $g(\cdot)$ as in \cite{advent}.
Following \cite{advent}, we set $\lambda = \beta = 0.001$. For this task of semantic segmentation, we do not do soft selection of source samples, \ie, $\{ w_i^s = 1 \}_{i=1}^{n_s}$.
We use SGD to optimize the segmentation network $f\circ\varphi$, where we set the learning rate, momentum, weight decay, batch size, and number of training iterations respectively as $2.5 \times 10^{-4}$, $0.9$, $5 \times 10^{-4}$, $1$, and $250,000$. For $\phi$ and $g$, we use Adam \cite{adam} optimizer with learning rates $2.5 \times 10^{-4}$ and $10^{-4}$ respectively. The learning rate is scheduled by a polynomial decay of power $0.9$. Other implementation details are the same as those for image classification.

\subsection{Ablation Studies and Learning Analyses}
\label{SecIndAbaltion}

In this section, we present fine-grained ablation studies and analyses of learning on our proposed H-SRDC. 

\keypoint{Ablation Studies.}
We first examine the effects of four key components in \name, namely, Structural Regularization (SR), deep Discriminative Clustering (DisC), deep Generative Clustering (GenC), and the scheme for Soft Selection of Source Samples (S$^4$). Experiments are conducted on the benchmarks of Office-31 \cite{office31}, Office-Home \cite{officehome}, and VisDA-2017 \cite{visda2017} in an inductive setting. Results of \emph{mean} over all the used adaptation tasks of individual benchmarks are reported in Table \ref{table:induct_ablation}. Please refer to Appendix A.1 for the results on individual adaptation tasks. Note that for DisC, GenC, and DisC+GenC (without structural source regularization), we fine-tune a source pre-trained model using (\ref{EqnAuxiTarDistrConstrain}), (\ref{EqnAuxiTarDistrConstrainGenModeling}), and (\ref{EqnAuxiTarDistrConstrain})+(\ref{EqnAuxiTarDistrConstrainGenModeling}) respectively; for GenC and SRGenC, we evaluate the performance based on cluster assignments computed by (\ref{EqnDeepGenAssignProb}). We have the following observations: 
{\bf (1)} each of the three components (\ie, SRDisC, SRGenC, and S$^4$) brings performance gains, verifying their complementary effects in our proposed method; 
{\bf (2)} with or without structural source regularization, a combination of DisC and GenC improves over the individual components, verifying the efficacy of our method design; 
{\bf (3)} SRDisC+SRGenC outperforms DisC+GenC by a large margin, indicating that structural source regularization is effective for improving target clustering; 
{\bf (4)} DisC exceeds GenC and SRDisC exceeds SRGenC, showing the superiority and necessity of classifier-based deep discriminative clustering. 

\begin{table}[!t]
	\begin{center}
		\caption{
			Fine-grained ablation studies on the four key components of our proposed H-SRDC. They are Structural Regularization (SR), deep Discriminative Clustering (DisC), deep Generative Clustering (GenC), and the scheme for Soft Selection of Source Samples (S$^4$). Experiments are conducted on the benchmarks of Office-31 \cite{office31}, Office-Home \cite{officehome}, and VisDA-2017 \cite{visda2017} in an inductive setting. Please refer to Appendix A.1 for the results on individual adaptation tasks.
		}
		\label{table:induct_ablation}
		\vskip -0.4cm		
		\begin{tabular}{|l|c|c|c|}
			\hline
			Component
			& Office-31 & Office-Home & VisDA-2017 \\
			\hline
			\hline
			Source Only                             
			& 72.4 & 60.0 & 40.3 \\
			\hline
			
			DisC
			& 79.6 & 64.8 & 78.1 \\
			
			GenC
			& 78.6 & 63.5 & 66.7 \\
			
			DisC+GenC
			& 82.0 & 65.8 & 79.4 \\
			
			SRDisC
			& 81.8 & 66.5 & 81.2 \\
			
			SRGenC
			& 80.7 & 64.4 & 69.4 \\
			
			SRDisC+SRGenC
			& 84.0 & 67.3 & 82.4 \\
			
			SRDisC+SRGenC+S$^4$
			& \textbf{85.1} & \textbf{68.0} & \textbf{83.4} \\
			\hline
		\end{tabular}
	\end{center}\vspace{-0.6cm}
\end{table}

\begin{figure}[!t]
	\centering
	\subfloat[\textbf{A}$\to$\textbf{D}]{
		\begin{minipage}[t]{0.49\linewidth}
			\centering
			\includegraphics[height=1.6in]{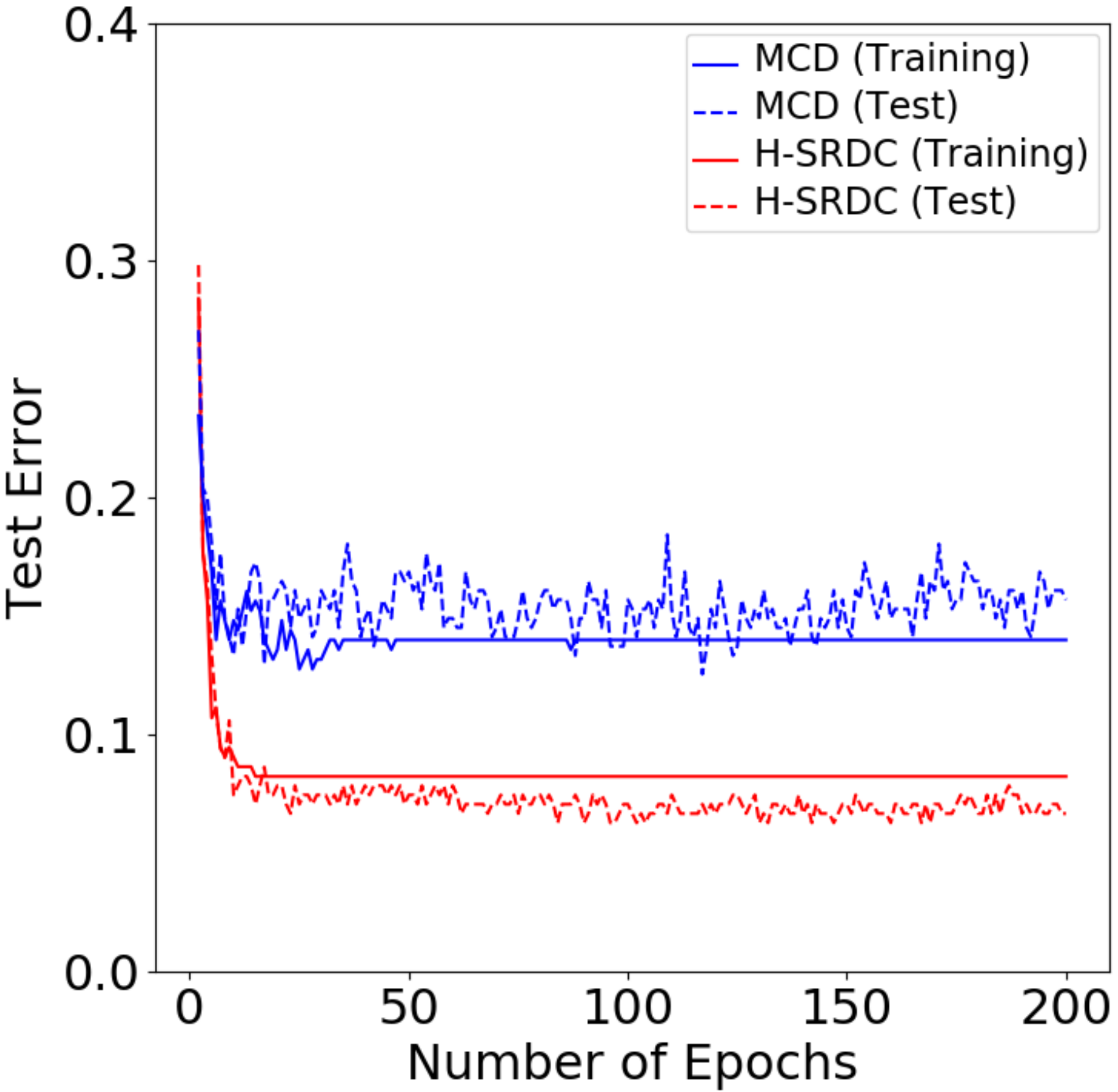}
			\label{fig:convergence:subfig1}
		\end{minipage}
	}%
	\subfloat[\textbf{D}$\to$\textbf{A}]{
		\begin{minipage}[t]{0.49\linewidth}
			\centering
			\includegraphics[height=1.6in]{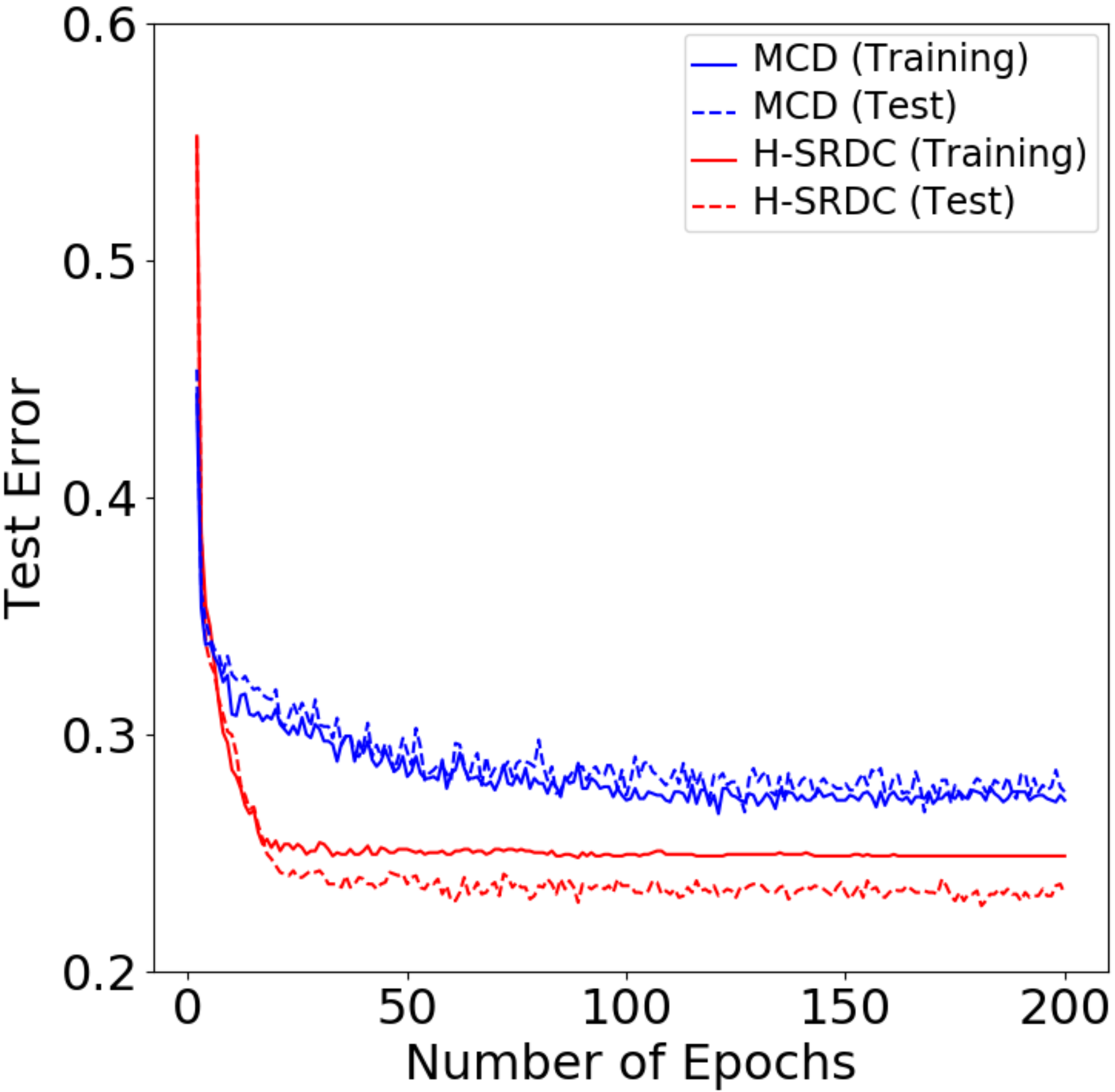}
			\label{fig:convergence:subfig2}
		\end{minipage}
	}%
	\\ \vskip -0.2cm
	\subfloat[\textbf{Ar}$\to$\textbf{Rw}]{
		\begin{minipage}[t]{0.49\linewidth}
			\centering
			\includegraphics[height=1.6in]{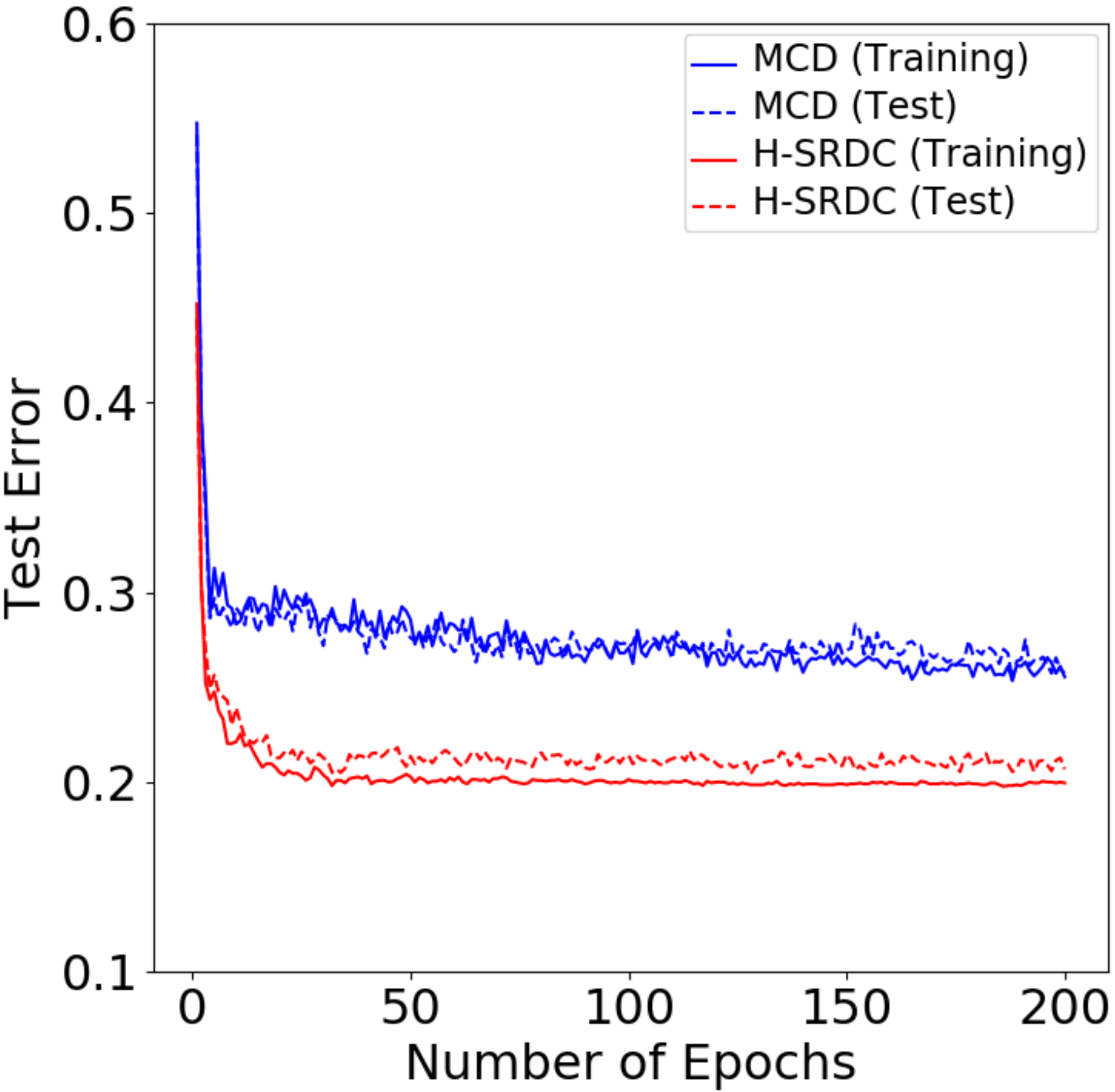}
			\label{fig:convergence:subfig3}
		\end{minipage}
	}%
	\subfloat[\textbf{Rw}$\to$\textbf{Ar}]{
		\begin{minipage}[t]{0.49\linewidth}
			\centering
			\includegraphics[height=1.6in]{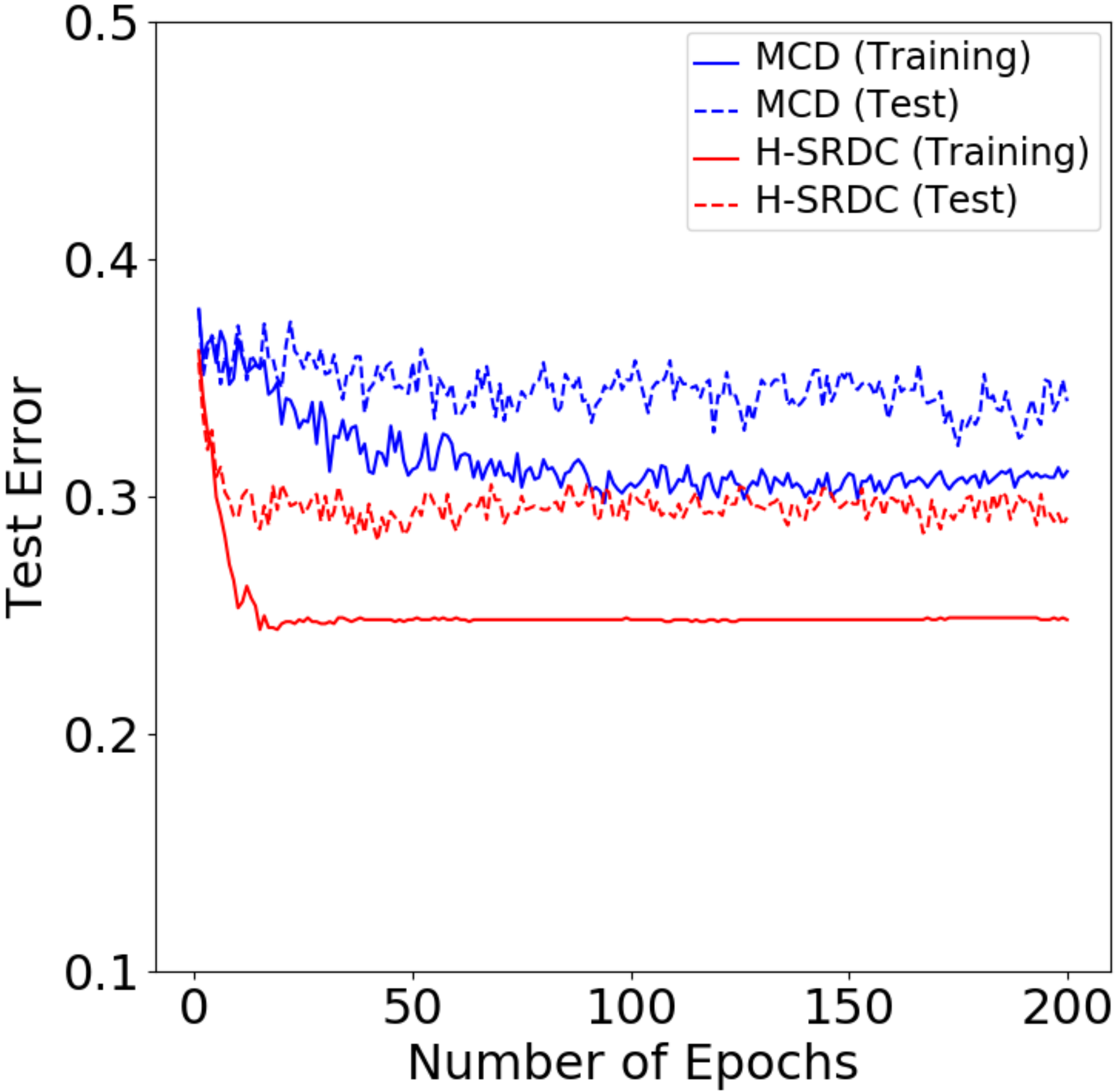}
			\label{fig:convergence:subfig4}
		\end{minipage}
	}%
	\vskip -0.3cm
	\caption{Analysis of convergence and generalization. ``Training'' and ``Test'' refer to results on training and held-out test sets on the target domain, respectively. Experiments in the inductive UDA setting are conducted on adaptation tasks of \textbf{A}$\to$\textbf{D} and \textbf{D}$\to$\textbf{A} on the Office-31 benchmark \cite{office31}, and \textbf{Ar}$\to$\textbf{Rw} and \textbf{Rw}$\to$\textbf{Ar} on the Office-Home benchmark \cite{officehome}. 
	}
	\label{fig:convergence}\vspace{-0.4cm}
\end{figure}

\keypoint{Diagnoses of Learning.}
As discussed in Section \ref{SecSRGenCSrc}, our SRGenC objective (\ref{EqnDeepGenClustAll}) is to modulate the feature space learning via generative clustering, rather than to explicitly align the features across domains, as in many of existing methods \cite{dann,mcd,dan,can,pfan,cat}. To verify this empirically, we conduct experiments on the adaptation tasks of \textbf{A}$\to$\textbf{D} and \textbf{D}$\to$\textbf{A} on the Office-31 benchmark, and examine how the following three types of distances in the feature space evolve during training; they are the Euclidean distance between each instance and its learned cluster centroid, averaged over all the instances (dubbed as Instance-to-Centroid), the distance between the class mean of instances and their centroid, averaged over all the classes (dubbed as InsMean-to-Centroid), and the distance between the source and target instance means of the corresponding class, averaged over all the classes (dubbed as SrcInsMean-to-TgtInsMean), where class assignments of target instances are based on their pseudo labels; the former two types of distances are computed separately for source and target data. In Fig. \ref{fig:l2_dist}, we plot the evolving curves of these distances during training, by comparing with the representative domain-aligning method MCD \cite{mcd}. We have the following observations from Fig. \ref{fig:l2_dist}: 
{\bf (1)} for both the source and target domains, the distances of Instance-to-Centroid and InsMean-to-Centroid decrease or increase first, and then stabilize at certain levels with the training, suggesting that our method does not enforce either source or target instances to collapse to the learned cluster centroids;
{\bf (2)} the SrcInsMean-to-TgtInsMean distance of MCD decreases linearly with the training, while that of our H-SRDC does not, 
indicating that our method is indeed modulating the feature space learning towards uncovering the intrinsic discrimination of target data, rather than aligning the features across domains. 
Observations from experiments on other UDA tasks and benchmarks are of similar quality (\cf~Appendix A.2). These observations corroborate our discussions in Section \ref{SecSRGenCSrc}.

\begin{table}[!t]
	\begin{center}
		\caption{Comparative results (\%) in the {\bf inductive} setting on the Office-31 benchmark \cite{office31}. All methods are based on the base model of ResNet-50.  
		}
		\label{table:inductive_office31}
		\vskip -0.4cm
		\begin{tabular}{|l|c|c|c|c|c|}%
			\hline
			Method & A $\rightarrow$W & A $\rightarrow$D & D $\rightarrow$A & W $\rightarrow$A  & \em mean \\
			\hline
			\hline
			Source Only                               & 79.3 & 81.6 & 63.1 & 65.7 & 72.4 \\
			\hline
			\hline
			DANN \cite{dann}                            & 80.8 & 82.4 & 66.0 & 64.6 & 73.5 \\
			
			MCD \cite{mcd}                              & 86.5 & 86.7 & 72.4 & 70.9 & 79.1 \\
			
			SRDC \cite{srdc}                            & 91.9 & 91.6 & 75.6 & 75.7 & 83.7 \\
			
			\bf \name{} 
			& \textbf{92.9} & \textbf{93.7} & \textbf{77.0} & \textbf{76.7} & \textbf{85.1} \\
			\hline
			\hline
			Oracle                                & 98.8 & 97.6 & 87.8 & 87.8 & 93.0 \\
			\hline
		\end{tabular}
	\end{center}\vspace{-0.6cm}
\end{table}

\begin{figure}[!t]
	\centering
	\subfloat[MCD \cite{mcd}]{
		\begin{minipage}[t]{0.49\linewidth}
			\centering
			\includegraphics[width=1.4in]{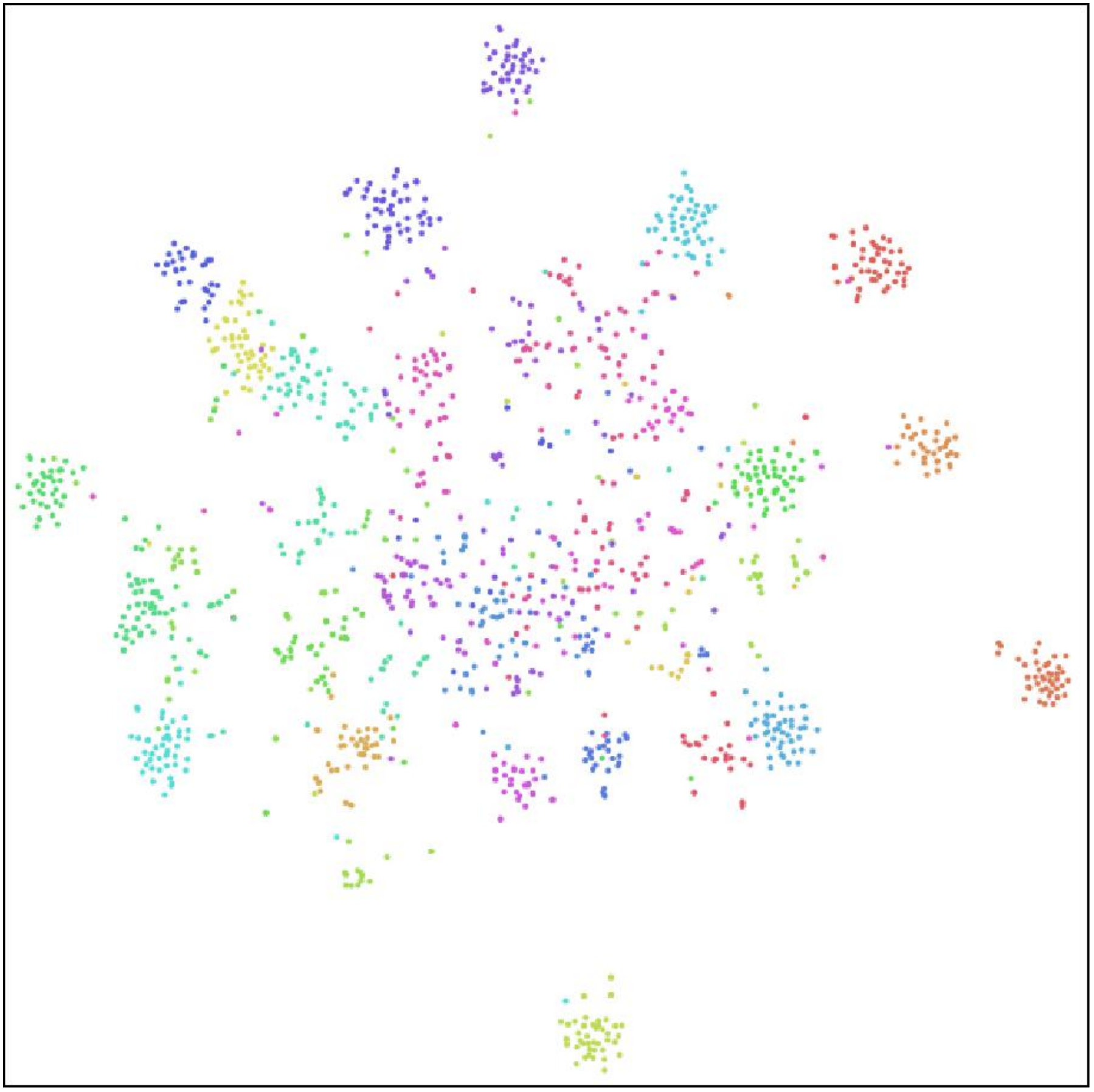}
		\end{minipage}%
	}%
	\subfloat[\name{}]{
		\begin{minipage}[t]{0.49\linewidth}
			\centering
			\includegraphics[width=1.4in]{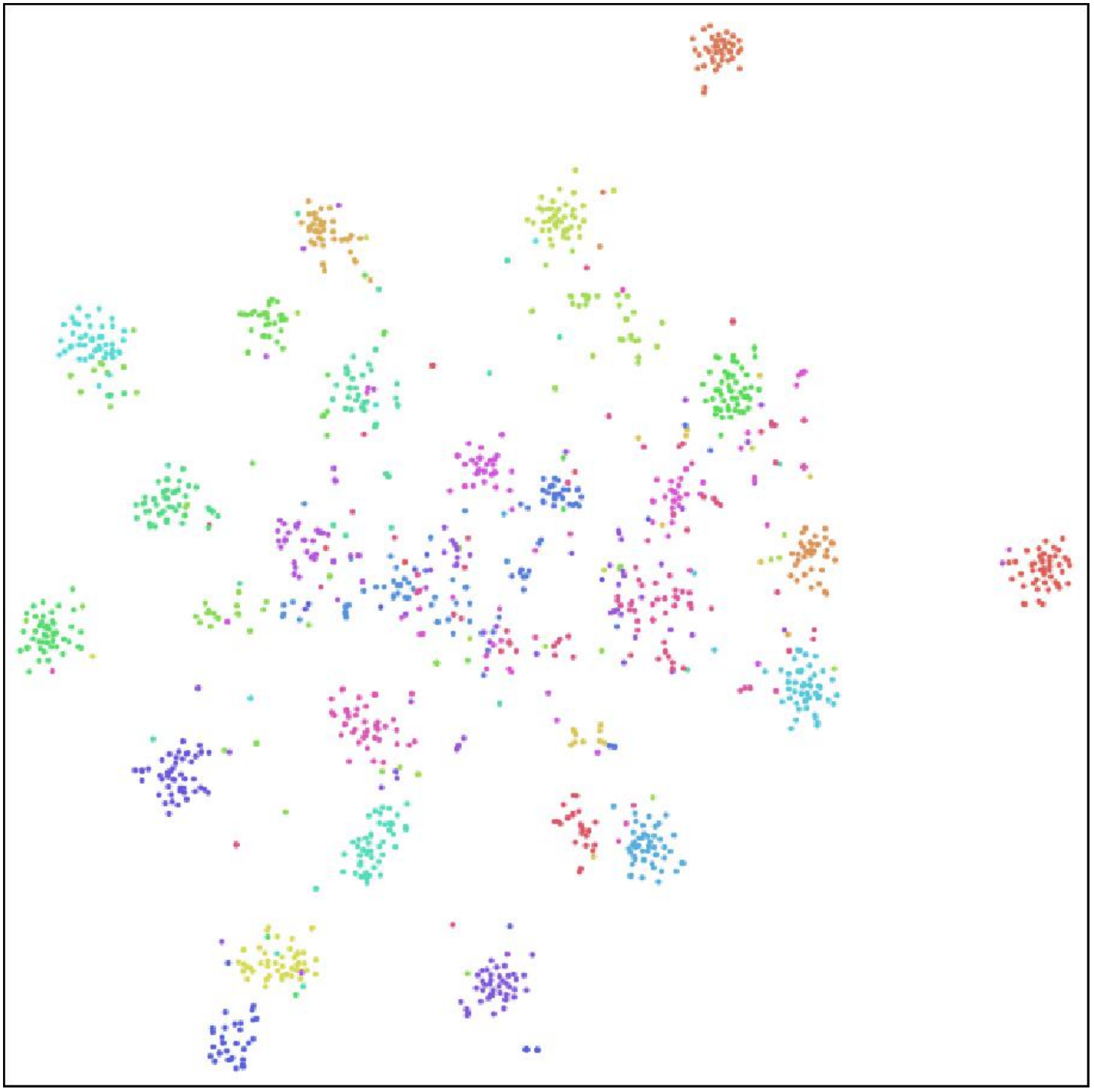}
		\end{minipage}
	}%
	\vskip -0.3cm
	\caption{The t-SNE visualization of feature distributions from MCD \cite{mcd} and \name{}, using the test data on the target domain (Amazon in Office-31). Classes are color coded.
	}
	\label{fig:inductive_tsne}\vspace{-0.4cm}
\end{figure}

\begin{table}[!t]
	\begin{center}
		\caption{Comparative results (\%) in the {\bf inductive} setting on the ImageCLEF-DA benchmark \cite{imageclefda}. All methods are based on the base model of ResNet-50.  
		}
		\label{table:inductive_imageclefda}
		\vskip -0.4cm
		\resizebox{1.0\linewidth}{!}{
			\begin{tabular}{|l|c|c|c|c|c|c|c|}
				\hline
				Method                 & I $\rightarrow$ P & P $\rightarrow$ I & I $\rightarrow$ C & C $\rightarrow$ I & C $\rightarrow$ P & P $\rightarrow$ C & \em mean \\
				\hline
				\hline
				Source Only        & 76.6 & 90.7 & 92.3 & 85.9 & 72.1 & 92.2 & 85.0 \\ 
				\hline
				\hline
				
				DANN \cite{dann}    & 77.2 & 88.4 & 92.9 & 87.6 & 72.9 & 92.2 & 85.2 \\
				
				MCD \cite{mcd}      & 78.4 & 88.9 & 95.2 & 88.3 & 73.8 & 92.1 & 86.1 \\
				
				SRDC \cite{srdc}    & \textbf{79.0} & 91.0 & 96.9 & 90.2 & 75.3 & 93.9 & 87.7 \\
				
				\bf \name{} 
				& \textbf{79.0} & \textbf{92.3} & \textbf{97.0} & \textbf{92.6} & \textbf{77.0} & \textbf{94.7} & \textbf{88.8} \\
				\hline
				\hline
				Oracle & 79.3 & 94.7 & 97.2 & 94.7 & 79.3 & 97.2 & 90.4 \\
				\hline
			\end{tabular}
		}
	\end{center}\vspace{-0.5cm}
\end{table}

\begin{table*}[!t]
	\begin{center}
		\caption{Comparative results (\%) in the {\bf inductive} setting on the Office-Home benchmark \cite{officehome}. All methods are based on the base model of ResNet-50.  
		}
		\label{table:inductive_officehome}
		\vskip -0.3cm
		\resizebox{0.92\textwidth}{!}{
			\begin{tabular}{|l|c|c|c|c|c|c|c|c|c|c|c|c|c|}
				\hline
				Method
				& Ar$\rightarrow$Cl & Ar$\rightarrow$Pr & Ar$\rightarrow$Rw & Cl$\rightarrow$Ar & Cl$\rightarrow$Pr & Cl$\rightarrow$Rw & Pr$\rightarrow$Ar & Pr$\rightarrow$Cl & Pr$\rightarrow$Rw & Rw$\rightarrow$Ar & Rw$\rightarrow$Cl & Rw$\rightarrow$Pr    &\em mean  \\
				\hline
				\hline
				Source Only     & 43.1      & 66.8     & 74.5      & 53.3      & 61.4      & 65.9     & 54.0     & 39.8     & 73.3     & 66.1     & 44.3     & 77.3 & 60.0 \\ 
				\hline
				\hline
				
				DANN \cite{dann}     & 44.3     & 64.8     & 74.7      & 55.0      & 61.0      & 66.4     & 55.3     & 41.1     & 73.9     & 66.2     & 48.1      & 77.7 & 60.7 \\
				
				MCD \cite{mcd}     & 47.4     & 68.2     & 74.6      & 54.7      & 64.8      & 67.2     & 56.5     & 46.7     & 74.2     & 66.4     & 52.2      & 77.7 & 62.6 \\
				
				SRDC \cite{srdc}     & 48.1 & 73.4 & 79.0 & 62.7 & 71.0 & 73.3 & 60.9 & 48.0 & 79.4 & 70.1 & \textbf{53.8} & 82.3 & 66.8 \\
				
				\bf \name{} 
				& \textbf{50.0} & \textbf{75.3} & \textbf{79.9} & \textbf{63.7} & \textbf{71.9} & \textbf{74.4} & \textbf{62.6} & \textbf{49.6} & \textbf{80.1} & \textbf{71.3} & 53.6 & \textbf{83.1} & \textbf{68.0} \\
				
				\hline
				\hline
				Oracle & 73.6 & 91.3 & 84.7 & 75.4 & 91.3 & 84.7 & 75.4 & 73.6 & 84.7 & 75.4 & 73.6 & 91.3 & 81.3 \\
				\hline
			\end{tabular}
		}
	\end{center}\vspace{-0.5cm}
\end{table*}

\begin{table*}[!t]
	\begin{center}
		\caption{Comparative results (\%) in the {\bf inductive} setting on the VisDA-2017 benchmark \cite{visda2017}. All methods are based on the base model of ResNet-50.  
		}
		\label{table:inductive_visda2017}
		\vskip -0.3cm
		\resizebox{0.88\textwidth}{!}{
			\begin{tabular}{|l|c|c|c|c|c|c|c|c|c|c|c|c|c|}
				\hline
				Method                & plane & bcycl & bus & car & horse & knife & mcycl & person & plant & sktbrd & train & truck &\em mean \\
				\hline
				\hline
				
				Source Only  & 58.7 & 12.1 & 59.1 & 63.3 & 34.6 & 2.4 & 84.5 & 4.9 & 58.2 & 19.0 & 82.6 & 4.6 & 40.3 \\ 
				\hline
				\hline
				
				DANN \cite{dann}  & 78.2 & 40.0 & 64.5 & 53.9 & 74.6 & 24.1 & 85.2 & 47.8 & 78.5 & 36.0 & 87.8 & 25.8 & 58.0 \\
				
				MCD \cite{mcd} & 90.2 & \textbf{78.8} & \textbf{84.7} & 72.7 & 91.1 & 67.1 & 85.3 & 75.2 & 91.6 & 75.1 & 83.6 & 32.2 & 77.3 \\
				
				SRDC \cite{srdc}     & 93.2 & 73.5 & 80.4 & 85.7 & 95.2 & 85.5 & 92.9 & 57.7 & 96.3 & \textbf{88.8} & \textbf{88.2} & \textbf{45.9} & 81.9 \\
				
				\bf \name{} 
				& \textbf{95.8} & 71.5 & 84.3 & \textbf{86.2} & \textbf{95.5} & \textbf{86.6} & \textbf{94.4} & \textbf{75.8} & \textbf{96.6} & 88.0 & 87.3 & 39.0 & \textbf{83.4} \\
				
				\hline
				\hline
				Oracle & 97.5 & 89.3 & 84.4 & 85.2 & 96.5 & 95.3 & 92.3 & 86.8 & 96.3 & 91.0 & 93.2 & 72.3 & 90.0 \\
				\hline
			\end{tabular}
		}
	\end{center}\vspace{-0.5cm}
\end{table*}

\begin{figure*}[!t]
	\begin{center}
		\includegraphics[width=0.83\linewidth]{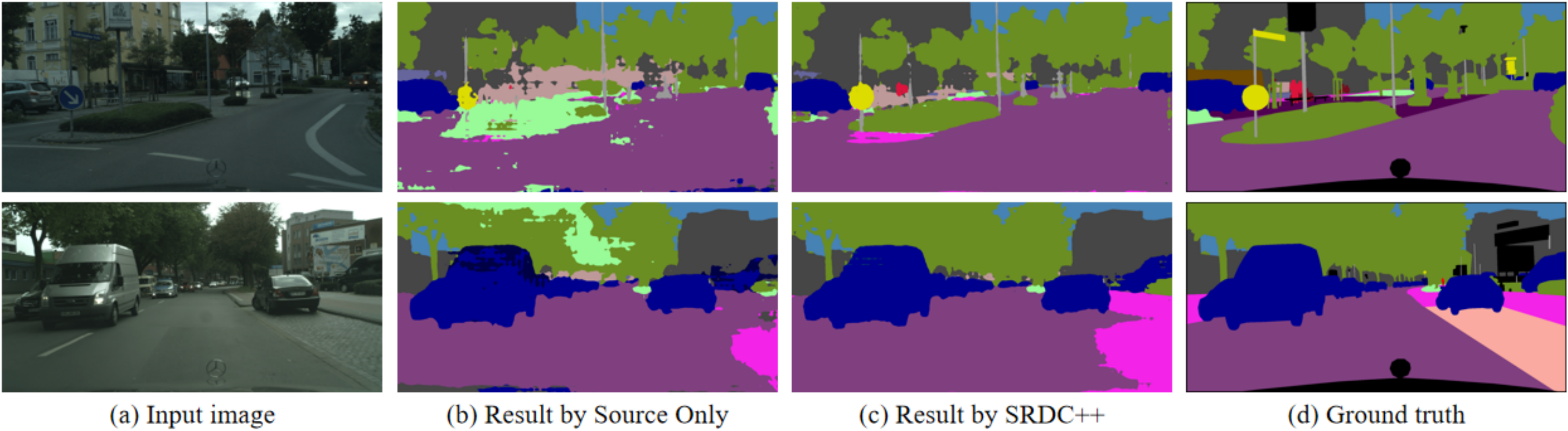}
	\end{center}
	\vskip -0.5cm
	\caption{Qualitative results on the inductive UDA task of \textbf{GTA5}$\rightarrow$\textbf{Cityscapes}. 
	}
	\label{fig:seg_examples}\vspace{-0.5cm}
\end{figure*}

\begin{table}[!t]
	\begin{center}
		\caption{Comparative results (\%) in the {\bf inductive} setting on the Digits benchmark. All methods are based on the base model of LeNet \cite{lenet}.  
		}
		\label{table:inductive_digits}
		\vskip -0.4cm
		\begin{tabular}{|l|c|c|c|c|}%
			\hline
			Method                                     & S $\rightarrow$M & M $\rightarrow$U & U $\rightarrow$M & \em mean \\
			\hline
			\hline
			Source Only                    & 67.1 & 76.7 & 63.4 & 69.1 \\
			\hline
			\hline            
			DANN \cite{dann}               & 71.1 & 77.1 & 73.0 & 73.7 \\
			
			
			DRCN \cite{drcn}               & 82.0 & 91.8 & 73.7 & 82.5 \\
			
			
			
			RAAN \cite{raan}               & 89.2 & 89.0 & 92.1 & 90.1 \\
			
			
			TPN \cite{tpn}                 & 93.0 & 92.1 & 94.1 & 93.1 \\
			
			CyCADA \cite{cycada}           & 90.4 & 95.6 & 96.5 & 94.2 \\
			
			
			
			
			MCD \cite{mcd}                 & 96.2 & 94.2 & 94.1 & 94.8 \\
			
			SRDC \cite{srdc}               & 98.1 & 96.2 & 96.5 & 96.9 \\
			
			GPDA \cite{gpda}               & 98.2 & 96.4 & 96.4 & 97.0 \\
			
			\bf \name{}                    & \textbf{99.0} & \textbf{96.9} & \textbf{97.1} & \textbf{97.7} \\
			\hline
			\hline
			Oracle      & 99.5 & 98.5 & 97.5 & 98.5 \\
			\hline
		\end{tabular}
	\end{center}\vspace{-0.6cm}
\end{table}

\keypoint{Analysis of Convergence and Generalization.}
We use the adaptation tasks of \textbf{A}$\to$\textbf{D} and \textbf{D}$\to$\textbf{A} on the Office-31 benchmark, and \textbf{Ar}$\to$\textbf{Rw} and \textbf{Rw}$\to$\textbf{Ar} on the Office-Home benchmark to report the training convergence of H-SRDC. Results are plotted in Fig. \ref{fig:convergence}, where we also compare with the MCD method \cite{mcd}. Fig. \ref{fig:convergence} shows that training of H-SRDC converges stably and fast. In the reported inductive setting of UDA tasks, when applying the learned models to held-out test instances on the target domain, H-SRDC has much lower test errors than MCD, showing the better generalization of our method. The analysis of convergence and generalization on the VisDA-2017 benchmark is of similar quality (\cf~Appendix A.3).

\subsection{Image Classification}
\label{SecIndImageClsExp}

In this section, we report comparative results of image classification in the inductive setting on the benchmarks of Office-31 \cite{office31}, ImageCLEF-DA \cite{imageclefda}, Office-Home \cite{officehome}, VisDA-2017 \cite{visda2017}, and Digits. As most of existing methods do not report results in the inductive setting (except on the Digits benchmark), we choose the most representative methods of DANN \cite{dann} and MCD \cite{mcd} to compare with; we also compare with our previous method of SRDC \cite{srdc}. We apply these methods in the inductive setting to report results. We use the methods of Source Only (\ie, training the same classification model $f\circ \varphi$ with labeled source data alone) and Oracle (\ie, training the model with labeled target data) as references of the lower and upper performance bounds, respectively. These results are reported in Tables \ref{table:inductive_office31}, \ref{table:inductive_imageclefda}, \ref{table:inductive_officehome}, \ref{table:inductive_visda2017}, and \ref{table:inductive_digits}.

On all the five benchmarks, H-SRDC outperforms DANN and MCD, often with a large margin; it outperforms SRDC as well. On Digits, H-SRDC also outperforms existing methods of TPN, CyCADA, and GPDA. These results confirm the superiority of our proposed method for UDA in the inductive setting. There still exist large performance gaps between results of H-SRDC and those of the Oracle method, suggesting that inductive UDA is indeed a challenging setting; further innovations are expected to close the gaps. For qualitative analysis, we visualize the distributions of the learned features from H-SRDC and MCD on the W$\rightarrow$A task of Office-31. Fig. \ref{fig:inductive_tsne} shows that H-SRDC learns a more discriminative feature space.


\begin{table*}[!t]
	\begin{center}
		\caption{ Comparative results (\%) in the {\bf inductive} setting on the benchmark of \textbf{GTA5}$\rightarrow$\textbf{Cityscapes}. All methods use ResNet-101 based DeepLab-v2 \cite{deeplab_v2} as the base model. Refer to the main text for the methods of SourceCE+Layout, H-SRDC w/o Layout, and H-SRDC w/o SRGenC. The mark $^\star$ indicates that results are obtained by running the officially released code from the marked method.
		}
		\label{table:inductive_gta52city}
		\vskip -0.6cm
		\resizebox{1.0\textwidth}{!}{
			\begin{tabular}{|c|c|c|c|c|c|c|c|c|c|c|c|c|c|c|c|c|c|c|c|c|}
				\hline
				Method  &
				\rotatebox{90}{road} & \rotatebox{90}{side.} & \rotatebox{90}{buil.} & \rotatebox{90}{wall} & \rotatebox{90}{fence} & \rotatebox{90}{pole} & \rotatebox{90}{light} & \rotatebox{90}{sign} & \rotatebox{90}{vege.} & \rotatebox{90}{terr.} & \rotatebox{90}{sky} & \rotatebox{90}{pers.} & \rotatebox{90}{rider} & \rotatebox{90}{car} & \rotatebox{90}{truck} & \rotatebox{90}{bus} & \rotatebox{90}{train} & \rotatebox{90}{motor} & \rotatebox{90}{bike} &  \rotatebox{90}{{\em mIoU}} \\
				\hline
				\hline
				
				Source Only  & 80.8 & 12.3 & 74.3 & 18.0 & 15.0 & 23.7 & 31.4 & 21.3 & 74.1 & 10.4 & 71.2 & 55.9 & \textbf{28.2} & 77.7 & 22.9 & 34.7 & 0.8 & 22.6 & 35.6 & 37.4 \\ 
                \hline
                \hline
				
				DANN \cite{dann}  & 83.9 & 27.8 & 78.4 & 28.4 & 19.0 & 26.6 & 31.0 & 24.4 & 82.0 & 24.9 & 76.3 & 56.5 & 26.9 & 75.1 & 22.7 & 32.8 & 0.4 & 25.4 & 38.5 & 41.1 \\
				
				MCD \cite{mcd} & 86.4 & 25.5 & 80.3 & 25.0 & 14.5 & 27.1 & 31.2 & \textbf{29.0} & 82.7 & 25.2 & 76.6 & 59.0 & 22.5 & 79.4 & 24.1 & 31.6 & 0.8 & 25.8 & 21.0 & 40.4 \\
				
				CLAN \cite{clan} & 87.0 & 27.1 & 79.6 & 27.3 & 23.3 & 28.3 & \textbf{35.5} & 24.2 & 83.6 & 27.4 & 74.2 & 58.6 & 28.0 & 76.2 & 33.1 & 36.7 & 6.7 & \textbf{31.9} & 31.4 & 43.2 \\
				
				AdaptSegNet \cite{Adapt_SegMap} & 86.5 & 36.0 & 79.9 & 23.4 & 23.3 & 23.9 & 35.2 & 14.8 & 83.4 & 33.3 & 75.6 & 58.5 & 27.6 & 73.7 & 32.5 & 35.4 & 3.9 & 30.1 & 28.1 & 42.4 \\
				
				AdvEnt \cite{advent} & \textbf{89.9} & \textbf{36.5} & 81.6 & \textbf{29.2} & \textbf{25.2} & \textbf{28.5} & 32.3 & 22.4 & \textbf{83.9} & \textbf{34.0} & 77.1 & 57.4 & 27.9 & 83.7 & 29.4 & 39.1 & 1.5 & 28.4 & 23.3 & 43.8 \\
				\hline
				SourceCE+Layout (AdvEnt$^\star$) & 87.2 & 35.1 & 80.9 & 24.1 & 22.5 & 26.5 & 30.8 & 20.4 & 83.6 & 30.5 & 76.8 & 56.4 & 26.9 & 81.7 & 28.8 & 37.2 & 1.9 & 29.9 & 27.7 & 42.6 \\
				
				 H-SRDC w/o Layout & 86.8 & 30.9 & 80.1 & 22.7 & 21.8 & 23.5 & 29.2 & 17.8 & 82.3 & 31.7 & 77.0 & 57.3 & 28.0 & 82.8 & \textbf{38.8} & \textbf{46.1} & \textbf{9.6} & 25.8 & 27.8 & 43.2 \\
				 
				 H-SRDC w/o SRGenC
				 & 89.7 & 31.5 & 80.3 & 27.8 & 25.0 & 22.2 & 32.3 & 16.5 & 83.0 & 33.7 & 77.0 & 56.7 & 27.2 & 84.1 & 33.4 & 38.1 & 3.0 & 28.3 & 24.2 & 42.8 \\
				
				\bf \name{} & &  &  &  &   &  &  &   &  &  &   &  &  &  &   &  &   &  &  &  \\
                (SRDisC+SRGenC+Layout) & 86.4 & 23.1 & \textbf{81.8} & 28.5 & 23.1 & 27.1 & 32.2 & 25.5 & 83.1 & 26.2 & \textbf{79.3} & \textbf{59.3} & 28.1 & \textbf{84.3} & 38.5 & 44.5 & 0.7 & 28.6 & \textbf{40.3} & \textbf{44.2} \\
				
				\hline
                \hline
				Oracle & 96.7 & 76.5 & 88.2 & 45.2 & 42.7 & 42.7 & 46.8 & 60.5 & 88.5 & 55.9 & 88.4 & 69.3 & 51.2 & 91.5 & 73.3 & 70.6 & 45.4 & 52.2 & 65.1 & 65.8 \\
				\hline
			\end{tabular}
		}
	\end{center}\vspace{-0.5cm}
\end{table*}

\begin{table*}[!t]
	\begin{center}
		\caption{Comparative results (\%) in the {\bf inductive} setting on the benchmark of \textbf{SYNTHIA}$\rightarrow$\textbf{Cityscapes}. All methods use ResNet-101 based DeepLab-v2 \cite{deeplab_v2} as the base model. Refer to the main text for the methods of SourceCE+Layout, H-SRDC w/o Layout, and H-SRDC w/o SRGenC. The mark $^\star$ indicates that results are obtained by running the officially released code from the marked method.
		}
		\label{table:inductive_syn2city}
		\vskip -0.6cm
		\resizebox{1.0\textwidth}{!}{
			\begin{tabular}{|c|c|c|c|c|c|c|c|c|c|c|c|c|c|c|c|c|c|}
				\hline
				Method  &
				\rotatebox{90}{road} & \rotatebox{90}{side.} & \rotatebox{90}{buil.} & \rotatebox{90}{wall} & \rotatebox{90}{fence} & \rotatebox{90}{pole} & \rotatebox{90}{light} & \rotatebox{90}{sign} & \rotatebox{90}{vege.} & \rotatebox{90}{sky} & \rotatebox{90}{pers.} & \rotatebox{90}{rider} & \rotatebox{90}{car} & \rotatebox{90}{bus} & \rotatebox{90}{motor} & \rotatebox{90}{bike} &  \rotatebox{90}{{\em mIoU}} \\
				\hline
				\hline
				
				Source Only
				& 41.1 & 20.1 & 71.6 & 3.8 & 0.0 & 26.6 & \textbf{11.9} & 12.1 & 75.3 & 79.4 & 53.7 & 17.8 & 39.1 & 21.4 & 12.3 & 26.2 & 32.0 \\ 
                \hline
                \hline
				
				DANN \cite{dann}  & 76.6 & 28.3 & 79.1 & 8.7 & \textbf{0.9} & 24.0 & 5.8 & 9.6 & 78.9 & \textbf{84.3} & 52.8 & 20.6 & 60.3 & 24.1 & 13.4 & 30.2 & 37.4 \\
				
				MCD \cite{mcd}    & 85.0 & 36.9 & 78.2 & 4.4 & 0.1 & \textbf{29.1} & 8.1 & 9.5 & 78.8 & 81.0 & 55.4 & 19.9 & 81.3 & 24.4 & 7.9 & 29.0 & 39.3 \\
				
				CLAN \cite{clan} & 78.0 & 34.1 & 78.1 & 7.5 & 0.2 & 27.3 & 8.8 & \textbf{13.4} & 78.1 & 81.5 & 55.3 & 21.1 & 66.4 & 22.3 & 12.4 & 31.5 & 38.5 \\
				
				AdaptSegNet \cite{Adapt_SegMap} & 81.7 & 39.1 & 78.4 & \textbf{11.1} & 0.3 & 25.8 & 6.8 & 9.0 & 79.1 & 80.8 & 54.8 & 21.0 & 66.8 & \textbf{34.7} & 13.8 & 29.9 & 39.6 \\
				
				AdvEnt \cite{advent} & 87.0 & \textbf{44.1} & \textbf{79.7} & 9.6 & 0.6 & 24.3 & 4.8 & 7.2 & \textbf{80.1} & 83.6 & 56.4 & \textbf{23.7} & 72.7 & 32.6 & 12.8 & \textbf{33.7} & 40.8 \\
                \hline
				
				SourceCE+Layout (AdvEnt$^\star$) & 86.7 & 39.2 & 78.0 & 7.9 & 0.3 & 22.0 & 5.0 & 7.0 & 78.0 & 82.3 & 54.3 & 18.7 & 73.5 & 30.7 & 17.5 & 29.3 & 39.4 \\
				
				H-SRDC w/o Layout & 83.2 & 39.0 & 78.8 & 2.1 & 0.1 & 23.7 & 6.7 & 8.4 & \textbf{80.6} & 83.6 & \textbf{57.9} & 15.5 & 76.7 & 33.4 & 5.8 & 27.5 & 38.9 \\
				
				H-SRDC w/o SRGenC
				& 83.7 & 38.4 & 77.5 & 7.9 & 0.6 & 24.8 & 8.6 & 8.1 & 78.1 & 83.1 & 54.2 & 19.5 & 67.4 & 35.4 & 18.2 & 28.3 & 39.6 \\
				
				\bf \name{}  &  & &  &  &  &  &  &  &  &  &  &  & &  &  &  &  \\
                (SRDisC+SRGenC+Layout) & \textbf{87.2} & 41.8 & 79.2 & 7.4 & 0.3 & 25.6 & 7.0 & 8.3 & 79.2 & 83.9 & \textbf{57.6} & 20.2 & \textbf{79.3} & 29.6 & \textbf{20.7} & 33.5 & \textbf{41.3}  \\
				
				\hline
                \hline
				Oracle & 96.9 & 77.7 & 88.3 & 45.7 & 43.0 & 43.7 & 47.5 & 62.7 & 89.4 & 90.9 & 69.1 & 50.9 & 91.9 & 70.4 & 51.4 & 65.3 & 67.8 \\
				\hline
			\end{tabular}
		}
	\end{center}\vspace{-0.5cm}
\end{table*}

\subsection{Semantic Segmentation}
\label{SecIndSemanticSegExp}

We use the objective (\ref{EqnSRDCPP4Seg}) when applying H-SRDC to adaptation tasks of semantic segmentation, which combines the terms of SRDisC and SRGenC, and also the two adversarial terms for promoting consistency of spatial layouts across domains (the two terms are dubbed as Layout). To investigate how these terms play roles in the learning, we conduct ablation studies by (1) removing the layout terms, which gives a method dubbed as H-SRDC w/o Layout, (2) replacing the terms of SRDisC and SRGenC with a standard cross-entropy loss applied to the labeled source data, which we dub as SourceCE+Layout, and is exactly the method of AdvEnt \cite{advent}, and (3) removing the SRGenC term, which gives a method dubbed as H-SRDC w/o SRGenC. We conduct experiments on the benchmark tasks of \textbf{GTA5} $\rightarrow$\textbf{Cityscapes} and \textbf{SYNTHIA} $\rightarrow$\textbf{Cityscapes}. Results in Tables \ref{table:inductive_gta52city} and \ref{table:inductive_syn2city} show that compared with the H-SRDC objective (\ref{EqnSRDCPP4Seg}), the three ablations cause clear degradation of performance. The experiments confirm both the efficacy of our proposed hybrid model of structurally regularized deep clustering (\ie, SRDisC + SRGenC) for semantic segmentation, and its compatibility with the established practice of enforcing the layout-wise consistency.

In Tables \ref{table:inductive_gta52city} and \ref{table:inductive_syn2city}, we also compare with the existing results reported in the literature, which are all achieved in the inductive UDA setting. On both of the two tasks, our proposed H-SRDC outperforms all the existing methods, confirming its efficacy for tasks other than image classification. However, the results are still largely behind those from the Oracle method, suggesting that semantic segmentation in an inductive UDA setting is very challenging; better approaches are to be developed to close the gaps. Example segmentation maps on the task of \textbf{GTA5} $\rightarrow$\textbf{Cityscapes} are given in Fig. \ref{fig:seg_examples}.

%

\begin{table*}[!t]
	\begin{center}
		\caption{Comparative results (\%) in the {\bf transductive} setting on the benchmarks of Office-31 \cite{office31}, ImageCLEF-DA \cite{imageclefda}, Office-Home \cite{officehome}, and VisDA-2017 \cite{visda2017}. All methods are based on the base model of ResNet-50 except for those on VisDA-2017 that use ResNet-101 as the base model. Please refer to Appendix D for the results on individual adaptation tasks. 
		}
		\label{table:transductive_results}
		\vskip -0.3cm
		\begin{tabular}{|l|c|c|c|c|}
			\hline
			Method                & Office-31 & ImageCLEF-DA & Office-Home & VisDA-2017 \\
			
			\hline
			\hline
			Source Only        & 81.1 & 80.7 & 46.1 & 52.4 \\
			
			
			DAN \cite{dan}          & 82.3 & 82.5 & 56.3 & 61.1 \\
			
			DANN \cite{dann}       & 82.6 & 85.0 & 57.6 & 57.4 \\
			
			
			
			
			
			
			ETD \cite{etd}          & 86.2 & 89.7 & 67.3 & - \\
			
			
			
			MCD \cite{mcd}                  & 86.5 & - & - & 71.9 \\
			
			SAFN \cite{larger_norm}   & 87.1 & 88.9 & 67.3 & - \\
			
			
			
			rRevGrad+CAT \cite{cat}   & 87.6 & 87.3 & - & - \\ 
			
			CDAN+E \cite{cdan}  & 87.7 & 87.7 & 65.8 & - \\
			
			
			
			
			SymNets \cite{symnets}          & 88.4 & 89.9 & 67.6 & - \\ 
			
			BSP+CDAN \cite{bsp}             & 88.5 & - & 66.3 & 75.9 \\
			
			
			CDAN+BNM \cite{bnm}             & 88.6 & - & 69.4 & - \\
			
			
			CADA-P \cite{cada}              & 89.5 & 88.3 & 70.2 & - \\ 
			
			CAN \cite{can}                  & 90.6 & 88.0 & 68.3 & 87.2 \\ 
			
			\hline
			SRDC \cite{srdc}        & 90.8 & 90.9 & 71.3 & 86.3 \\
			
			\bf \name{} 			& \textbf{90.9} & \textbf{91.2} & \textbf{72.6} & \textbf{87.4} \\			
			\hline
		\end{tabular}
	\end{center}\vspace{-0.5cm}
\end{table*}

\section{Domain Adaptation in a Transductive Setting}
\label{SecTransExps}

In this section, we present experiments of domain-adapted image classification in the transductive setting to verify the efficacy of our proposed H-SRDC. 

\keypoint{Settings.}
We use the same four benchmarks of Office-31, ImageCLEF-DA, Office-Home, and VisDA-2017 for our experiments. Different from the inductive setting, we follow the literature of transductive UDA for image classification and use all the labeled source data and unlabeled target data when training the UDA models; results are directly compared on the learned label assignments of target data. Following the standard protocol in the literature, we use ResNet-50 as the base network for all the benchmarks except VisDA-2017, for which we use ResNet-101 as the base network. Other settings and implementation details are the same as those used in the inductive setting (\cf~Section \ref{SecIndExp}).

\subsection{Comparative Results}

We compare \name{} with the state-of-the-art methods in Table \ref{table:transductive_results}, including our previous SRDC \cite{srdc}, in terms of \emph{mean} over all the used adaptation tasks of individual benchmarks. Please refer to Appendix D for the results on individual adaptation tasks. While the performance of existing methods may vary on the four benchmarks, H-SRDC performs consistently well and outperforms all existing methods on the four benchmarks. By comparing the results in this table with those in Tables \ref{table:inductive_office31}, \ref{table:inductive_imageclefda}, \ref{table:inductive_officehome}, and \ref{table:inductive_visda2017}, we observe that UDA tasks in the transductive setting are easier than the inductive counterparts; nevertheless, the efficacy of H-SRDC is still verified on these easier tasks. Note that most of the state-of-the-art methods achieve domain adaptation by explicitly aligning features between the source and target domains; instead, H-SRDC achieves the goal by directly uncovering the intrinsic structures of target discrimination, regularized by the labeled source discrimination. Our empirical results have verified that such a structurally regularized uncovering approach is a promising direction; more future research along the line is expected.

\section{Conclusion}

In this work, we have proposed a hybrid model of structurally regularized deep clustering, termed H-SRDC, for unsupervised domain adaptation. The objective of H-SRDC includes two key components respectively for discriminative and generative clusterings of unlabeled target data, which are constrained by structural regularization from the labeled source data. The discriminative component is based on a deep clustering framework that minimizes the KL divergence between predictive and auxiliary distributions of network outputs. To enable generative clustering, H-SRDC learns domain-shared cluster centroids in the feature space via self-attentive interactions of instance features. In this work, we also extend H-SRDC for the adaptation task of semantic segmentation, by incorporating into its objective additional terms for promoting consistency of spatial layouts between the segmentation maps on the source and target domains. We report comparative results on UDA benchmarks of image classification and semantic segmentation. On all the benchmarks, we achieve the new state of the art without explicitly learning and aligning features across the source and target domains. This shows that given regularization from the labeled source data, directly uncovering the intrinsic discrimination of target data is a promising approach; we expect that more future research along the line would be pursued. For adaptation tasks of image classification, we report comprehensive experiments in both the inductive and transductive settings; we expect that these results contribute as new benchmarks to the research community.

	
	%

	

	\section*{Acknowledgments}
	This work is supported in part by the National Natural Science Foundation of China (Grant No.: 61771201), the Program for Guangdong Introducing Innovative and Enterpreneurial Teams (Grant No.: 2017ZT07X183), the Guangdong R\&D key project of China (Grant No.: 2019B010155001), and Microsoft Research Asia.

	\ifCLASSOPTIONcaptionsoff
	\newpage
	\fi

	
	

	\bibliographystyle{IEEEtran}
	\bibliography{IEEEref}

	%
	
	
	
	%
	
	\begin{IEEEbiography}[{\includegraphics[width=1in,height=1.25in,clip,keepaspectratio]{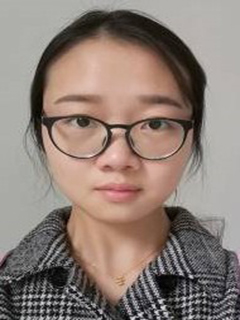}}]{Hui Tang}
		received the B.E. degree in the School of Electronic and Information Engineering, South China University of Technology, in 2018. She is currently pursuing the Ph.D. degree in the School of Electronic and Information Engineering, South China University of Technology. Her research interests are in computer vision and machine learning.
	\end{IEEEbiography}
	
	\begin{IEEEbiography}[{\includegraphics[width=1in,height=1.25in,clip,keepaspectratio]{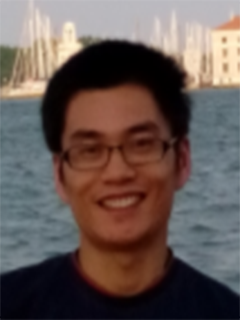}}]{Xiatian Zhu}
		received the B.E. and M.E. degrees from University of Electronic Science and Technology of China, and the Ph.D. degree from Queen Mary University of London, in 2015. He won The Sullivan Doctoral Thesis Prize in 2016, an annual award representing the best doctoral thesis submitted to a UK University in the field of computer or natural vision. His research interests include computer vision, pattern recognition, and machine learning.
		
	\end{IEEEbiography}
	
	\begin{IEEEbiography}[{\includegraphics[width=1in,height=1.25in,clip,keepaspectratio]{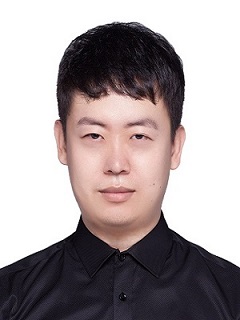}}]{Ke Chen}
		received the B.E. degree in automation and the M.E. degree in software engineering from Sun Yat-sen University, in 2007 and 2009, respectively
		, and the Ph.D. degree in computer vision from the School of Electronic Engineering and Computer Science, Queen Mary University of London, in 2013
		. He has been a Postdoctoral Research Fellow with the Department of Signal Processing, Tampere University of Technology, since 2013. He is currently an Associate Professor with the School of Electronic and Information Engineering, South China University of Technology (SCUT). 
		His research interests include computer vision, pattern recognition, neural dynamic modeling, and robotic inverse kinematics.
		
	\end{IEEEbiography}
	
	\begin{IEEEbiography}[{\includegraphics[width=1in,height=1.25in,clip,keepaspectratio]{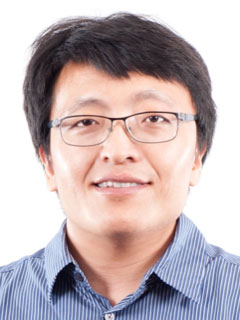}}]{Kui Jia}
		received the B.E. degree from Northwestern Polytechnic University, Xi’an, China, in 2001, the M.E. degree from the National University of Singapore, Singapore, in 2004, and the Ph.D. degree in computer science from the Queen Mary University of London, London, U.K., in 2007.
		He was with the Shenzhen Institute of Advanced Technology of the Chinese Academy of Sciences, Shenzhen, China, Chinese University of Hong Kong, Hong Kong, the Institute of Advanced Studies, University of Illinois at Urbana-Champaign, Champaign, IL, USA, and the University of Macau, Macau, China. He is currently a Professor with the School of Electronic and Information Engineering, South China University of Technology, Guangzhou, China. His recent research focuses on theoretical deep learning and its applications in vision and robotic problems, including deep learning of 3D data and deep transfer learning.
	\end{IEEEbiography}

	\begin{IEEEbiography}[{\includegraphics[width=1in,height=1.25in,clip,keepaspectratio]{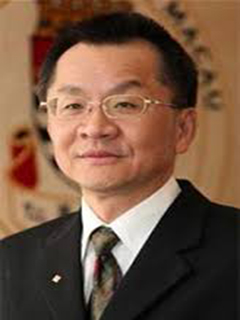}}]{C. L. Philip Chen}
		(S88-M88-SM94-F07) received the M.S. degree from the University of Michigan, Ann Arbor, MI, USA, in 1985, and the Ph.D. degree from Purdue University, West Lafayette, IN, USA, 1988, both in electrical engineering. He is currently the Chair Professor and Dean of the College of Computer Science and Engineering, South China University of Technology, Guangzhou, China. He received IEEE Norbert Wiener Award in 2018 for his contribution in systems and cybernetics, and machine learnings. His current research interests include systems, cybernetics, and computational intelligence. He is currently the Editor-in-Chief of the IEEE Transactions on Cybernetics. He is a Fellow of IEEE, AAAS, IAPR, CAA, and HKIE; a member of Academia Europaea (AE), European Academy of Sciences and Arts (EASA), and International Academy of Systems and Cybernetics Science (IASCYS).
	\end{IEEEbiography}

\include{appendix}
	
\end{document}

%% file: appendix.tex
\appendices

\section{More Ablation Studies and Learning Analyses}

\subsection{Ablation Studies}

\begin{table*}[!t]
	\begin{center}
		\caption{
			Fine-grained ablation studies on the four key components of our proposed H-SRDC. They are Structural Regularization (SR), deep Discriminative Clustering (DisC), deep Generative Clustering (GenC), and the scheme for Soft Selection of Source Samples (S$^4$). Experiments are conducted on the Office-31 benchmark \cite{office31} in an inductive setting.
		}
		\label{table:induct_ablation_office31}			
		\begin{tabular}{|l|c|c|c|c|c|}
			\hline
			Component
			& A $\rightarrow$W & A $\rightarrow$D & D $\rightarrow$A & W $\rightarrow$A  & \em mean \\
			\hline
			\hline
			Source Only                             
			& 79.3$\pm$1.1 & 81.6$\pm$0.6 & 63.1$\pm$0.7 & 65.7$\pm$0.8 & 72.4 \\
			\hline
			
			DisC
			& 86.3$\pm$0.2 & 84.4$\pm$0.9 & 72.7$\pm$0.4 & 75.1$\pm$0.2 & 79.6 \\
			
			GenC
			& 88.6$\pm$0.3 & 86.0$\pm$0.4 & 69.0$\pm$0.2 & 70.7$\pm$0.3 & 78.6 \\
			
			DisC+GenC
			& 89.4$\pm$0.4 & 88.1$\pm$0.2 & 74.9$\pm$0.3 & 75.5$\pm$0.3 & 82.0 \\
			
			SRDisC
			& 90.2$\pm$0.1 & 89.2$\pm$0.2 & 73.5$\pm$0.4 & 74.1$\pm$0.3 & 81.8 \\
			
			SRGenC
			& 88.9$\pm$0.5 & 88.2$\pm$0.4 & 72.1$\pm$0.2 & 73.5$\pm$0.1 & 80.7 \\
			
			SRDisC+SRGenC
			& 91.8$\pm$0.3 & 92.3$\pm$0.2 & 76.0$\pm$0.5 & 76.0$\pm$0.3 & 84.0 \\
			
			SRDisC+SRGenC+S$^4$
			& \textbf{92.9}$\pm$0.1 & \textbf{93.7}$\pm$0.6 & \textbf{77.0}$\pm$0.2 & \textbf{76.7}$\pm$0.1 & \textbf{85.1} \\
			\hline
		\end{tabular}
	\end{center}
\end{table*}

\begin{table*}[!t]
	\begin{center}
		\caption{Fine-grained ablation studies on the four key components of our proposed H-SRDC. They are Structural Regularization (SR), deep Discriminative Clustering (DisC), deep Generative Clustering (GenC), and the scheme for Soft Selection of Source Samples (S$^4$). Experiments are conducted on the Office-Home benchmark \cite{officehome} in an inductive setting.
		}
		\label{table:induct_ablation_officehome}			
		\resizebox{0.95\textwidth}{!}{
			\begin{tabular}{|l|c|c|c|c|c|c|c|c|c|c|c|c|c|}
				\hline
				Component
				& Ar$\rightarrow$Cl & Ar$\rightarrow$Pr & Ar$\rightarrow$Rw & Cl$\rightarrow$Ar & Cl$\rightarrow$Pr & Cl$\rightarrow$Rw & Pr$\rightarrow$Ar & Pr$\rightarrow$Cl & Pr$\rightarrow$Rw & Rw$\rightarrow$Ar & Rw$\rightarrow$Cl & Rw$\rightarrow$Pr    &\em mean  \\
				\hline
				\hline
				Source Only     
				& 43.1      & 66.8     & 74.5      & 53.3      & 61.4      & 65.9     & 54.0     & 39.8     & 73.3     & 66.1     & 44.3     & 77.3 & 60.0 \\ 
				\hline
				
				DisC
				& 44.8 & 70.5 & 76.3 & 61.2 & 68.5 & 72.4 & 60.0 & 46.3 & 78.5 & 69.2 & 49.8 & 80.4 & 64.8 \\
				
				GenC
				& 44.0 & 68.4 & 77.6 & 60.8 & 68.1 & 72.2 & 59.5 & 40.5 & 78.0 & 66.6 & 47.6 & 78.9 & 63.5 \\
				
				DisC+GenC
				& 46.5 & 71.7 & 77.8 & 62.6 & 69.5 & 72.8 & 61.5 & 47.1 & 79.4 & 69.4 & 50.2 & 81.6 & 65.8 \\
				
				SRDisC
				& 49.0 & 73.8 & 78.8 & 62.0 & 70.4 & 73.2 & 61.3 & 48.5 & 79.1 & 69.3 & 51.0 & 82.0 & 66.5 \\
				
				SRGenC
				& 46.9 & 71.9 & 78.7 & 59.5 & 67.9 & 72.1 & 58.9 & 42.1 & 78.3 & 68.5 & 48.3 & 79.9 & 64.4 \\
				
				SRDisC+SRGenC
				& 49.5 & 74.5 & 79.5 & 62.8 & 71.0 & 73.7 & 62.2 & 49.5 & 80.1 & 70.1 & 52.3 & 82.6 & 67.3 \\
				
				SRDisC+SRGenC+S$^4$
				& \textbf{50.0} & \textbf{75.3} & \textbf{79.9} & \textbf{63.7} & \textbf{71.9} & \textbf{74.4} & \textbf{62.6} & \textbf{49.6} & \textbf{80.1} & \textbf{71.3} & 53.6 & \textbf{83.1} & \textbf{68.0} \\					
				\hline
			\end{tabular}
		}
	\end{center}
\end{table*}

\begin{table*}[!t]
	\begin{center}
		\caption{Fine-grained ablation studies on the four key components of our proposed H-SRDC. They are Structural Regularization (SR), deep Discriminative Clustering (DisC), deep Generative Clustering (GenC), and the scheme for Soft Selection of Source Samples (S$^4$). Experiments are conducted on the VisDA-2017 benchmark \cite{visda2017} in an inductive setting.
		}
		\label{table:induct_ablation_visda2017}
		\begin{tabular}{|l|c|c|c|c|c|c|c|c|c|c|c|c|c|}
			\hline
			Component                & plane & bcycl & bus & car & horse & knife & mcycl & person & plant & sktbrd & train & truck &\em mean \\
			\hline
			\hline
			
			Source Only  & 58.7 & 12.1 & 59.1 & 63.3 & 34.6 & 2.4 & 84.5 & 4.9 & 58.2 & 19.0 & 82.6 & 4.6 & 40.3 \\ 
			\hline
			
			DisC
			& 94.7 & 69.9 & 88.1 & 71.3 & 91.7 & 83.4 & 92.3 & 77.4 & 87.8 & 79.0 & 84.9 & 17.1 & 78.1 \\
			
			GenC
			& 95.5 & 76.8 & 86.7 & 59.1 & 89.5 & 38.2 & 66.7 & 71.8 & 76.3 & 59.8 & 72.1 & 7.5 & 66.7 \\
			
			DisC+GenC
			& 94.8 & 72.0 & 88.8 & 70.6 & 91.8 & 91.1 & 91.4 & 79.2 & 89.1 & 82.7 & 85.3 & 16.0 & 79.4 \\
			
			SRDisC
			& 93.3 & 75.0 & 76.4 & 87.2 & 93.1 & 82.8 & 92.6 & 68.1 & 96.2 & 84.1 & 85.6 & 39.4 & 81.2 \\
			
			SRGenC
			& 75.2 & 62.5 & 67.9 & 50.6 & 89.2 & 95.9 & 71.5 & 47.8 & 74.3 & 62.2 & 78.0 & 57.3 & 69.4 \\
			
			SRDisC+SRGenC
			& 95.0 & 78.8 & 80.0 & 86.3 & 93.6 & 82.4 & 92.3 & 61.7 & 97.0 & 91.2 & 87.1 & 42.8 & 82.4 \\
			
			SRDisC+SRGenC+S$^4$
			& \textbf{95.8} & 71.5 & 84.3 & \textbf{86.2} & \textbf{95.5} & \textbf{86.6} & \textbf{94.4} & \textbf{75.8} & \textbf{96.6} & 88.0 & 87.3 & 39.0 & \textbf{83.4} \\
			
			\hline
		\end{tabular}
	\end{center}
\end{table*}

Fine-grained ablation studies on individual adaptation tasks of Office-31 \cite{office31}, Office-Home \cite{officehome}, and VisDA-2017 \cite{visda2017} are shown in Tables \ref{table:induct_ablation_office31}, \ref{table:induct_ablation_officehome}, and \ref{table:induct_ablation_visda2017} respectively. Experiments are conducted in an inductive setting.

\subsection{Diagnoses of Learning}

\begin{figure*}[!t]
	\centering
	\subfloat[\footnotesize Instance-to-Center (\textbf{Ar}$\rightarrow$\textbf{Rw})]{
		\begin{minipage}[t]{0.33\textwidth}
			\centering
			\includegraphics[height=1.6in]{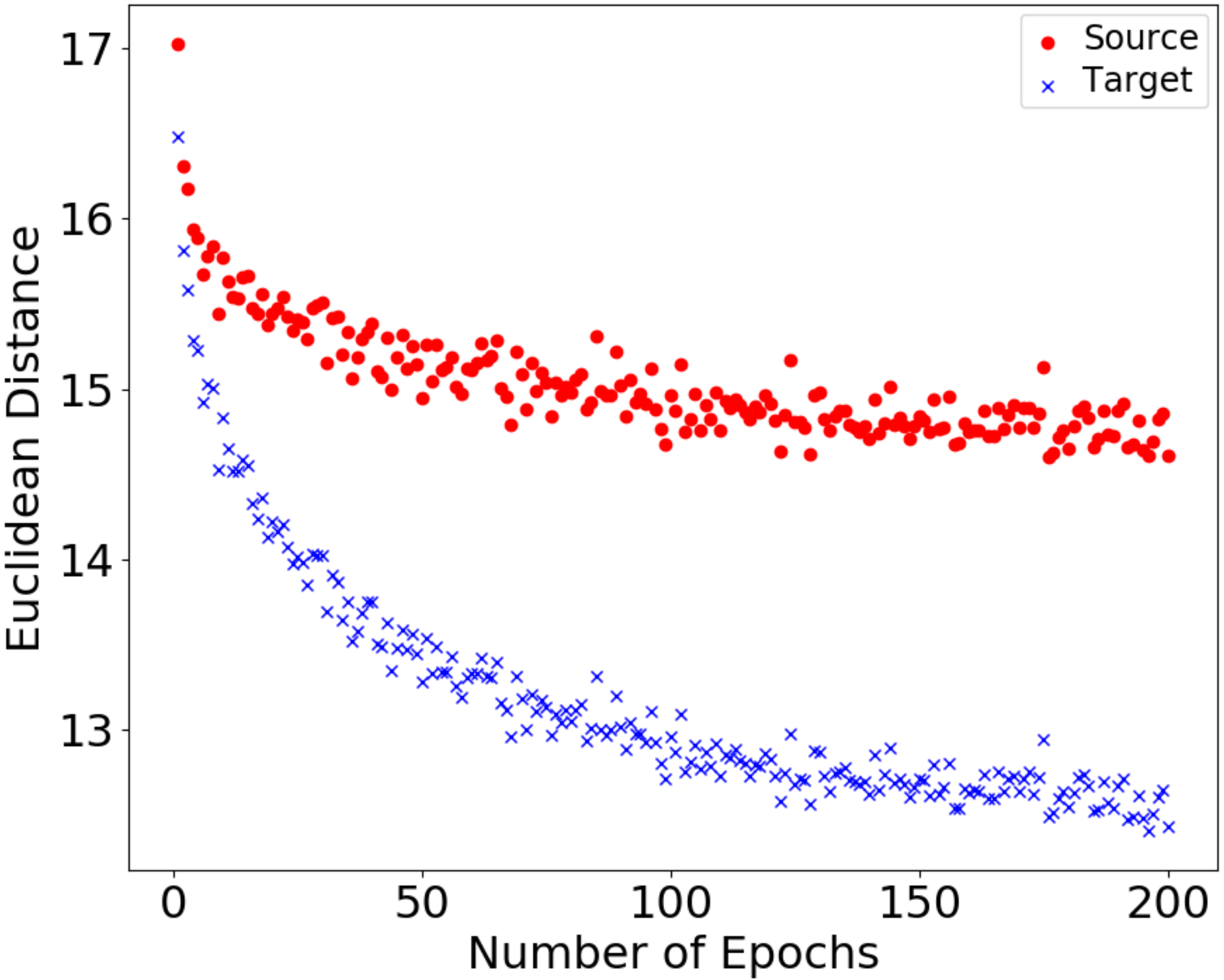}
			\label{fig:l2_dist_officehome:subfig1}
		\end{minipage}
	}%
	\subfloat[\footnotesize InsMean-to-Center (\textbf{Ar}$\rightarrow$\textbf{Rw})]{
		\begin{minipage}[t]{0.33\textwidth}
			\centering
			\includegraphics[height=1.6in]{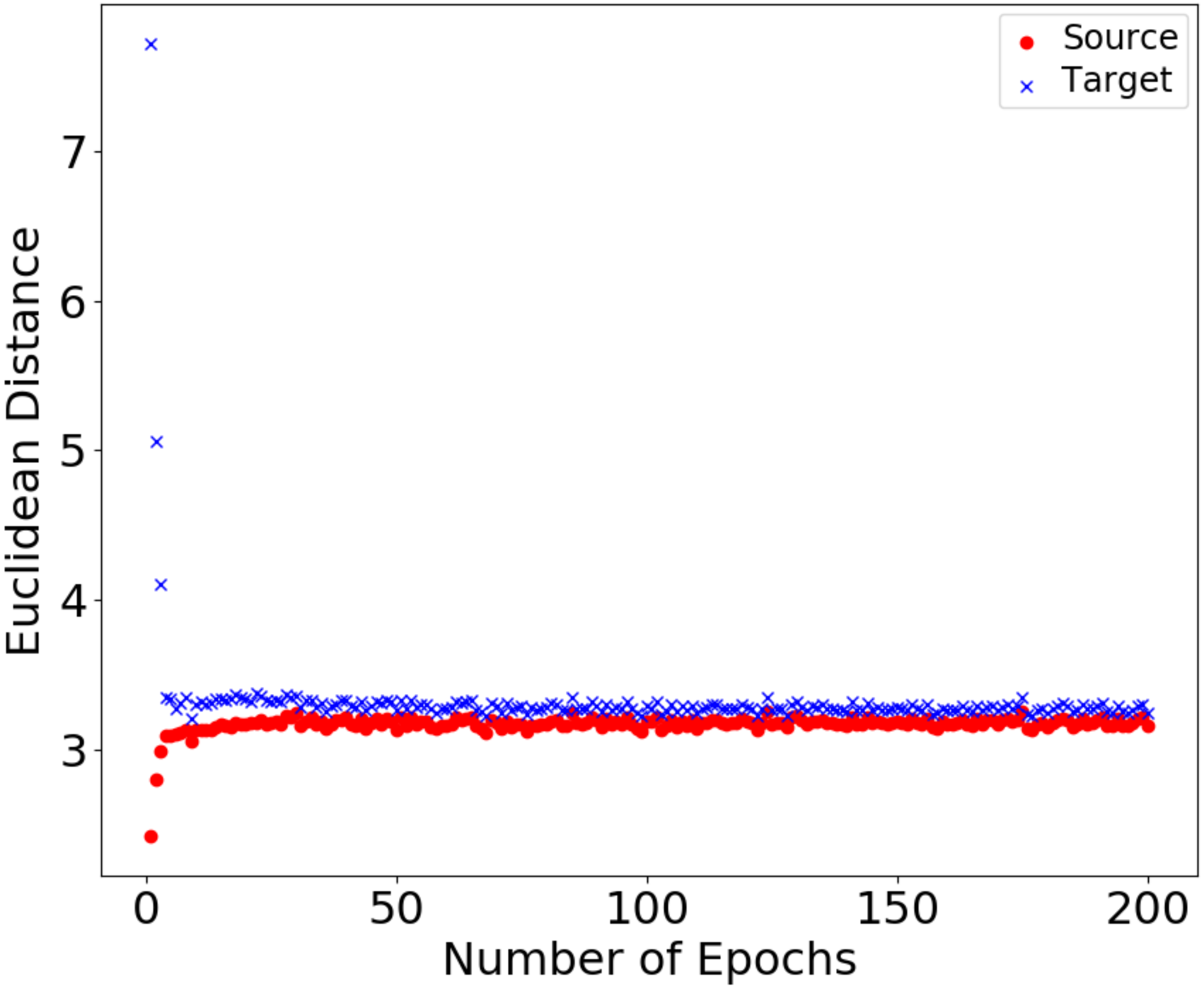}
			\label{fig:l2_dist_officehome:subfig2}
		\end{minipage}
	}%
	\subfloat[\footnotesize Instance-to-InsMean (\textbf{Ar}$\rightarrow$\textbf{Rw})]{
		\begin{minipage}[t]{0.33\textwidth}
			\centering
			\includegraphics[height=1.6in]{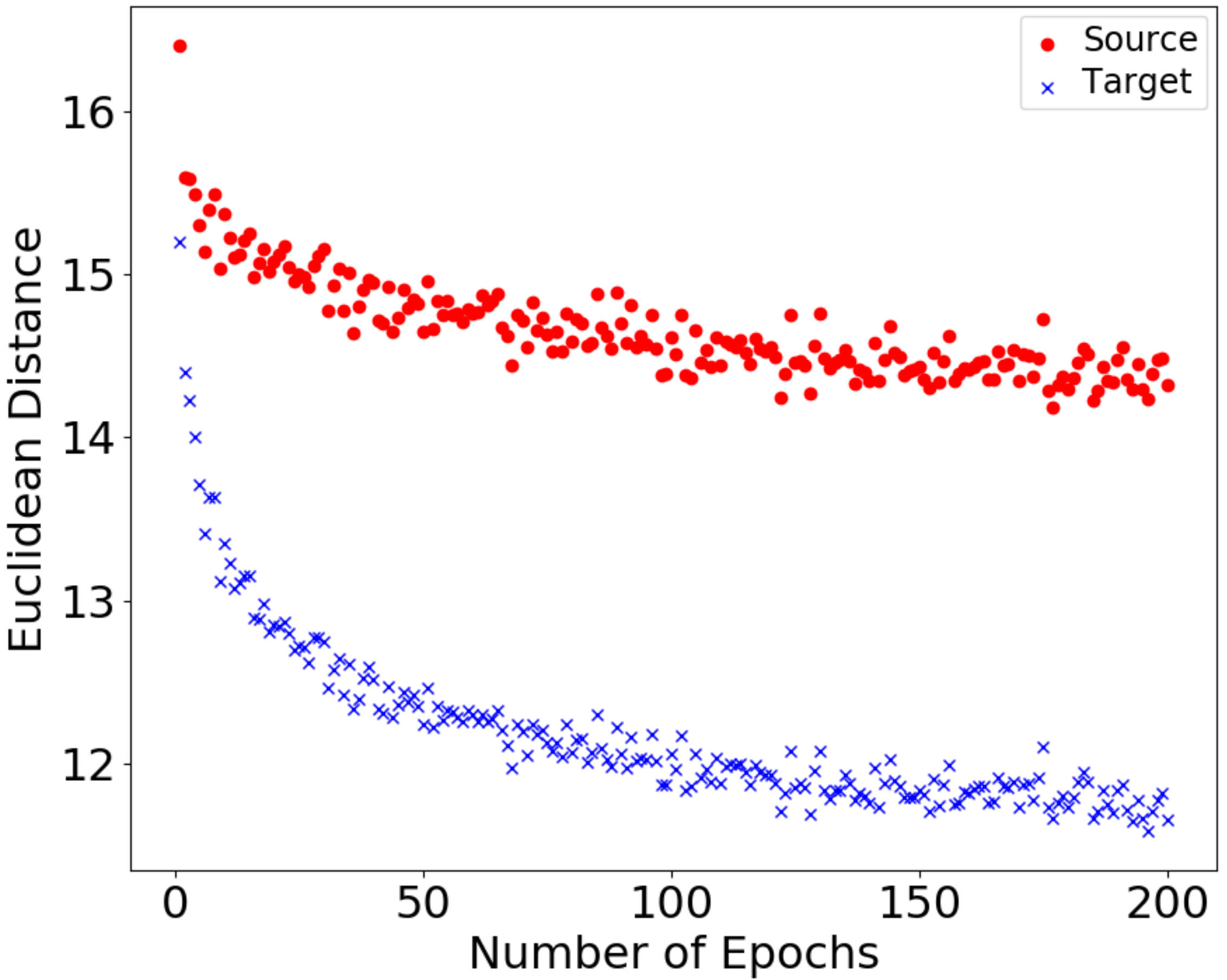}
			\label{fig:l2_dist_officehome:subfig3}
		\end{minipage}
	}%
	\\
	\subfloat[\footnotesize Instance-to-Centroid (\textbf{Ar}$\rightarrow$\textbf{Rw}) ]{
		\begin{minipage}[t]{0.33\textwidth}
			\centering
			\includegraphics[height=1.6in]{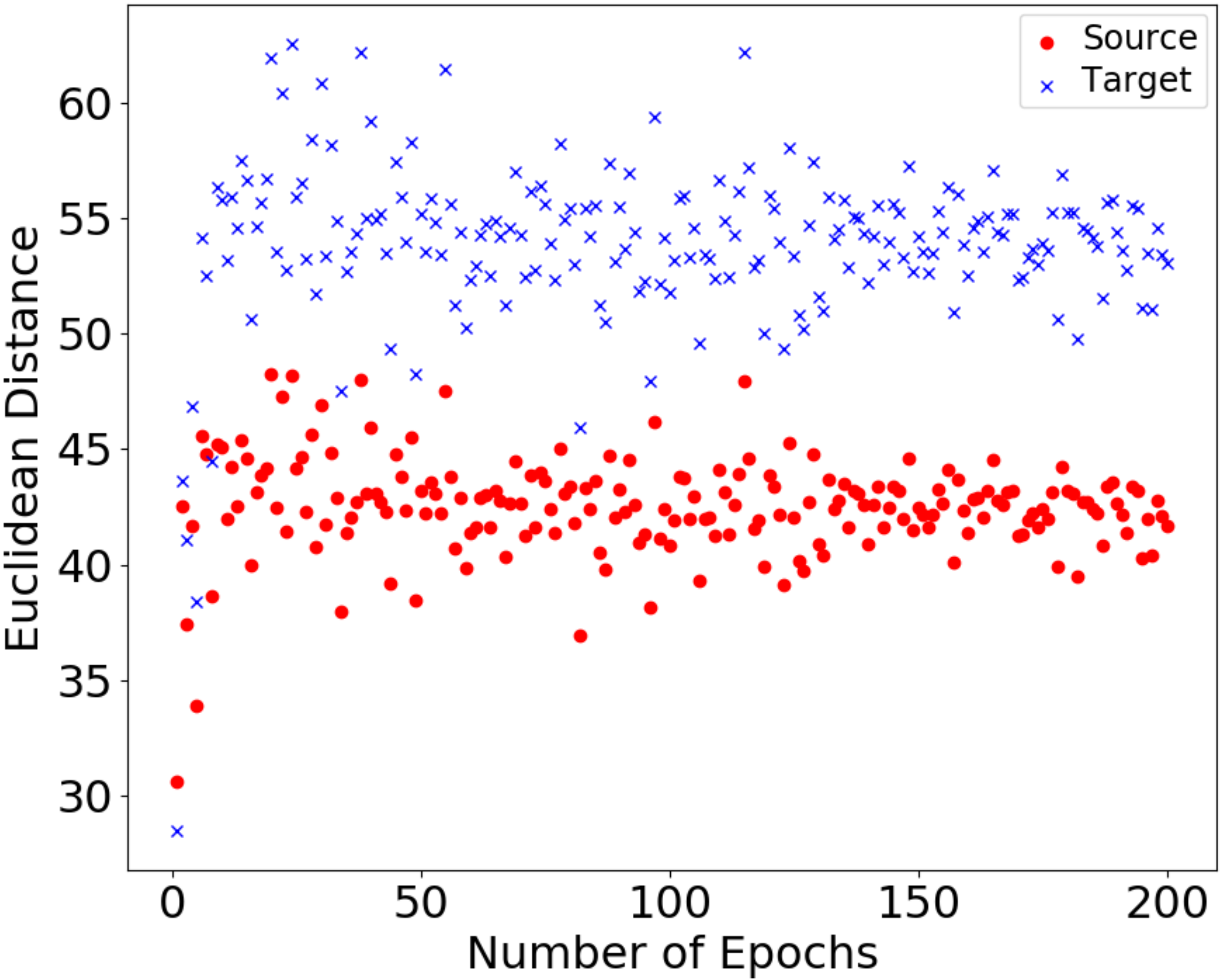}
			\label{fig:l2_dist_officehome:subfig4}
		\end{minipage}
	}%
	\subfloat[\footnotesize InsMean-to-Centroid (\textbf{Ar}$\rightarrow$\textbf{Rw})]{
		\begin{minipage}[t]{0.33\textwidth}
			\centering
			\includegraphics[height=1.6in]{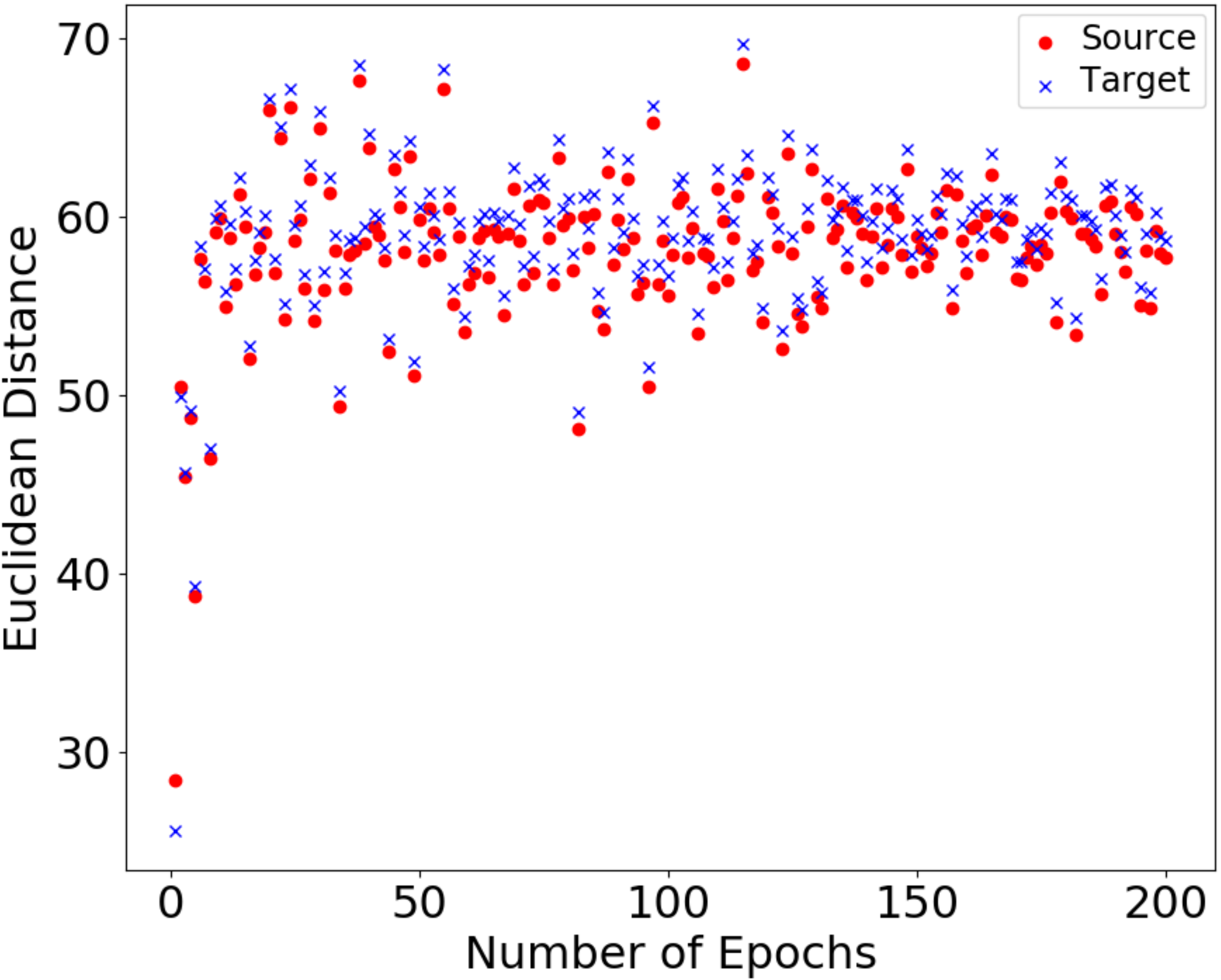}
			\label{fig:l2_dist_officehome:subfig5}
		\end{minipage}
	}%
	\subfloat[\footnotesize SrcInsMean-to-TgtInsMean (\textbf{Ar}$\rightarrow$\textbf{Rw})]{
		\begin{minipage}[t]{0.33\textwidth}
			\centering
			\includegraphics[height=1.6in]{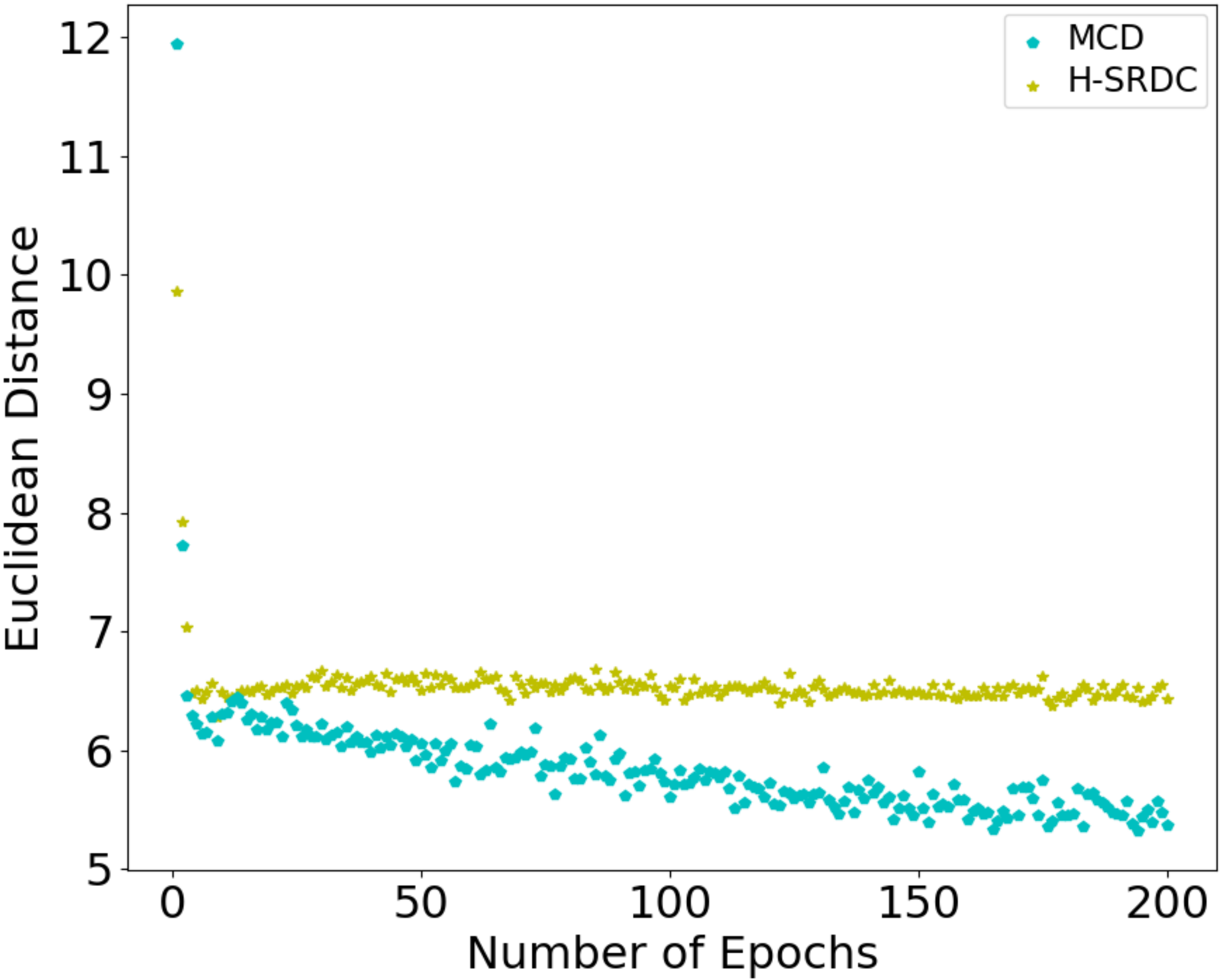}
			\label{fig:l2_dist_officehome:subfig6}
		\end{minipage}
	}%
	\\
	\subfloat[\footnotesize Instance-to-Center (\textbf{Rw}$\rightarrow$\textbf{Ar})]{
		\begin{minipage}[t]{0.33\textwidth}
			\centering
			\includegraphics[height=1.6in]{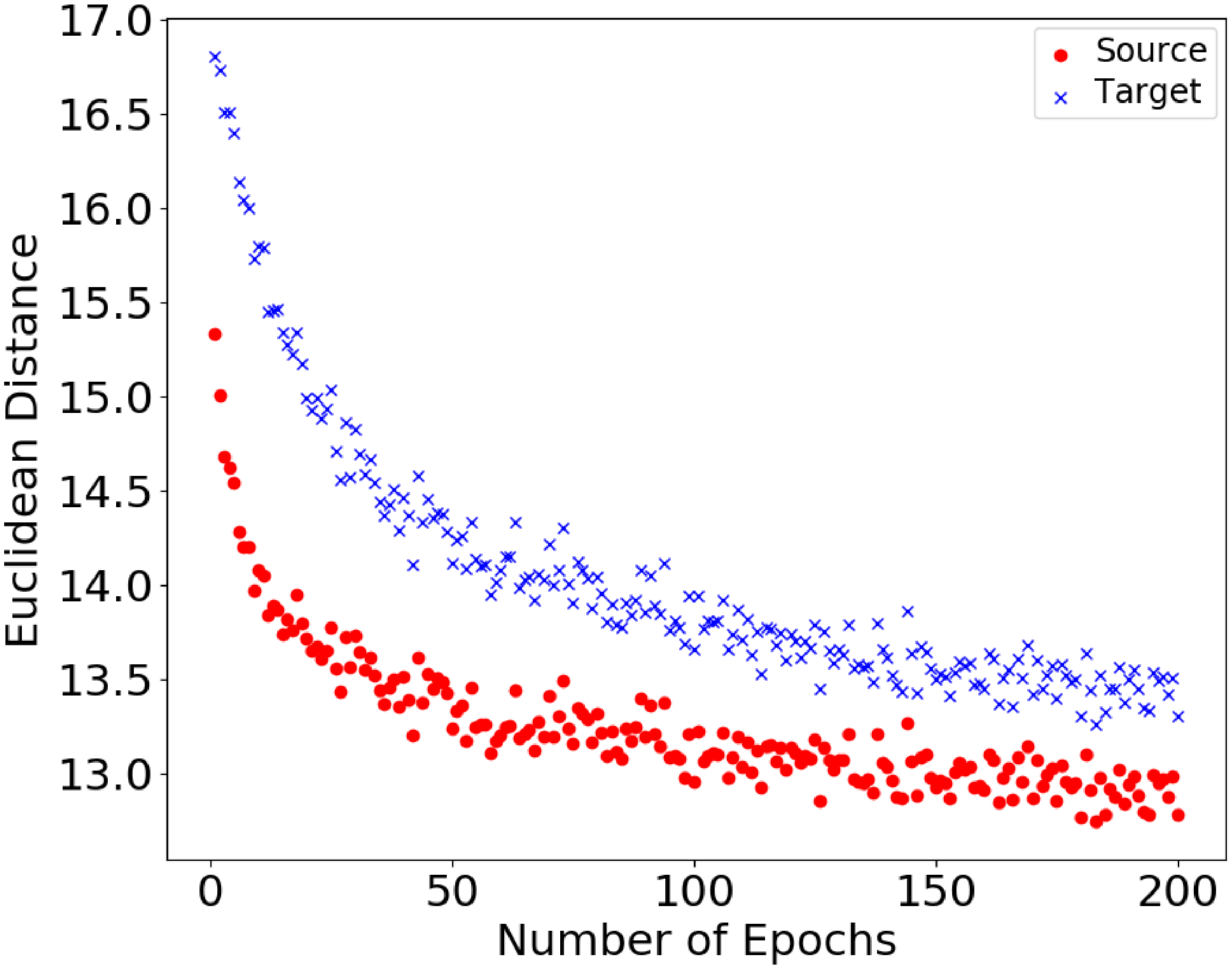}
			\label{fig:l2_dist_officehome:subfig7}
		\end{minipage}
	}%
	\subfloat[\footnotesize InsMean-to-Center (\textbf{Rw}$\rightarrow$\textbf{Ar})]{
		\begin{minipage}[t]{0.33\textwidth}
			\centering
			\includegraphics[height=1.6in]{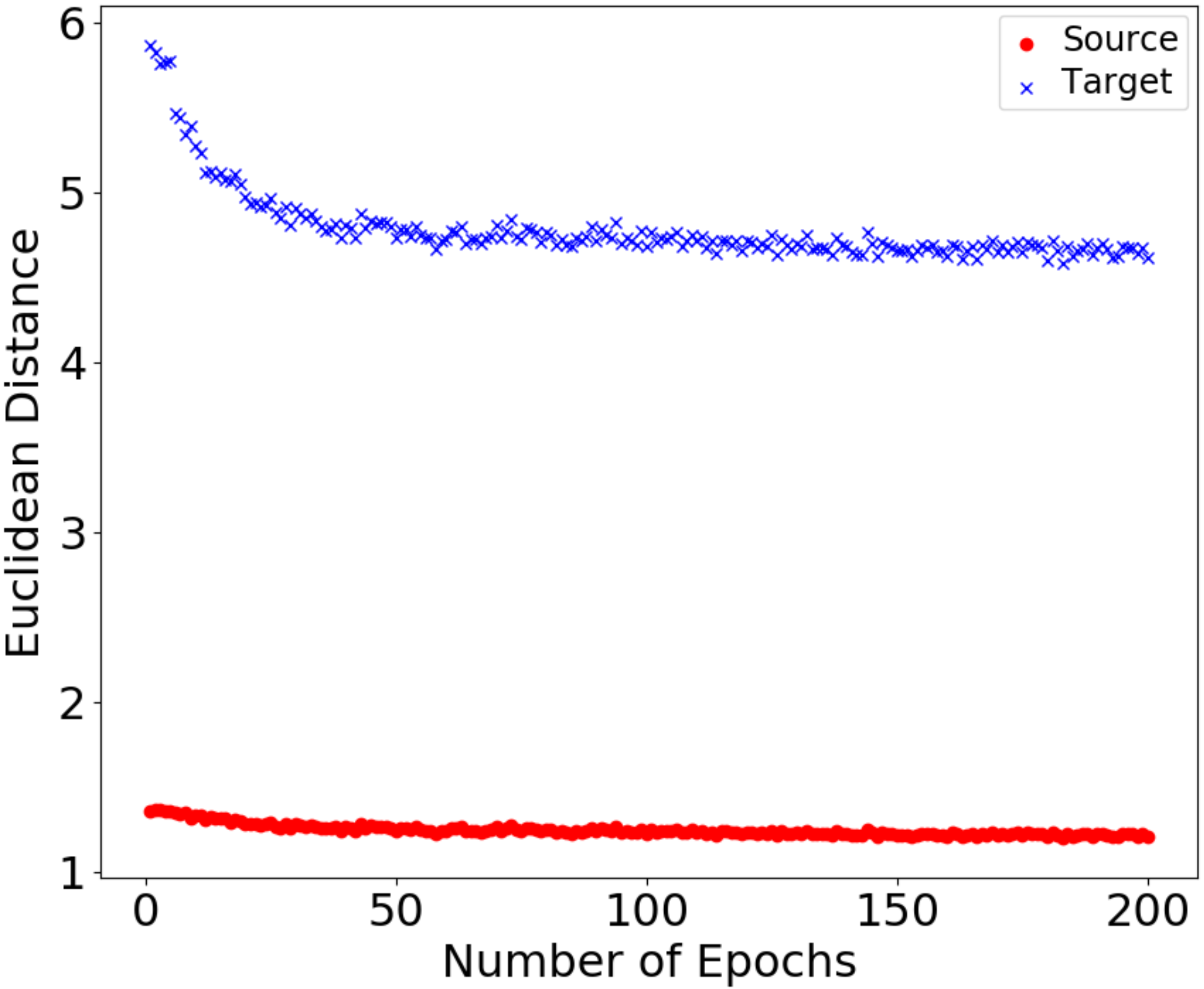}
			\label{fig:l2_dist_officehome:subfig8}
		\end{minipage}
	}%
	\subfloat[\footnotesize Instance-to-InsMean (\textbf{Rw}$\rightarrow$\textbf{Ar})]{
		\begin{minipage}[t]{0.33\textwidth}
			\centering
			\includegraphics[height=1.6in]{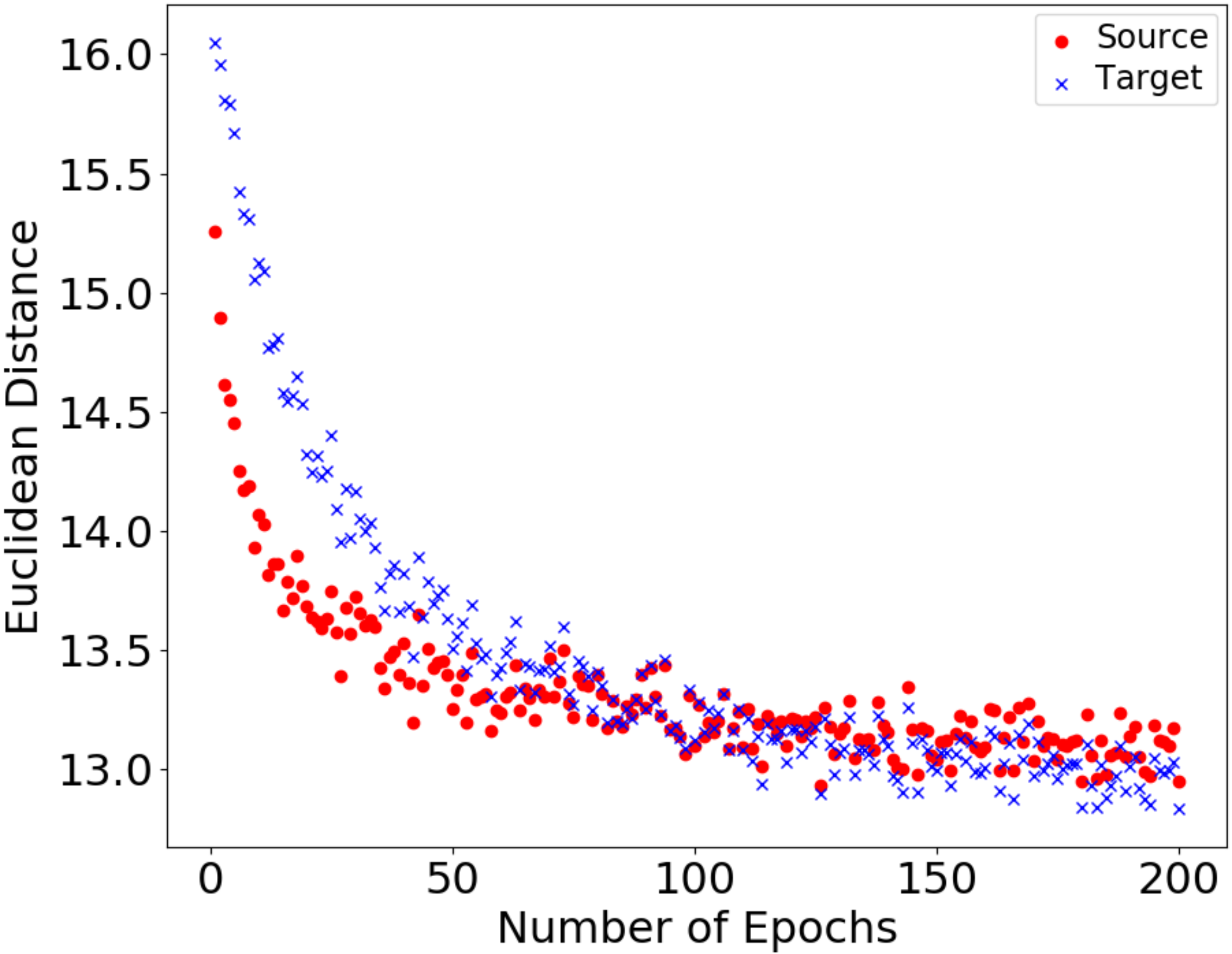}
			\label{fig:l2_dist_officehome:subfig9}
		\end{minipage}
	}%
	\\
	\subfloat[\footnotesize Instance-to-Centroid (\textbf{Rw}$\rightarrow$\textbf{Ar})]{
		\begin{minipage}[t]{0.33\textwidth}
			\centering
			\includegraphics[height=1.6in]{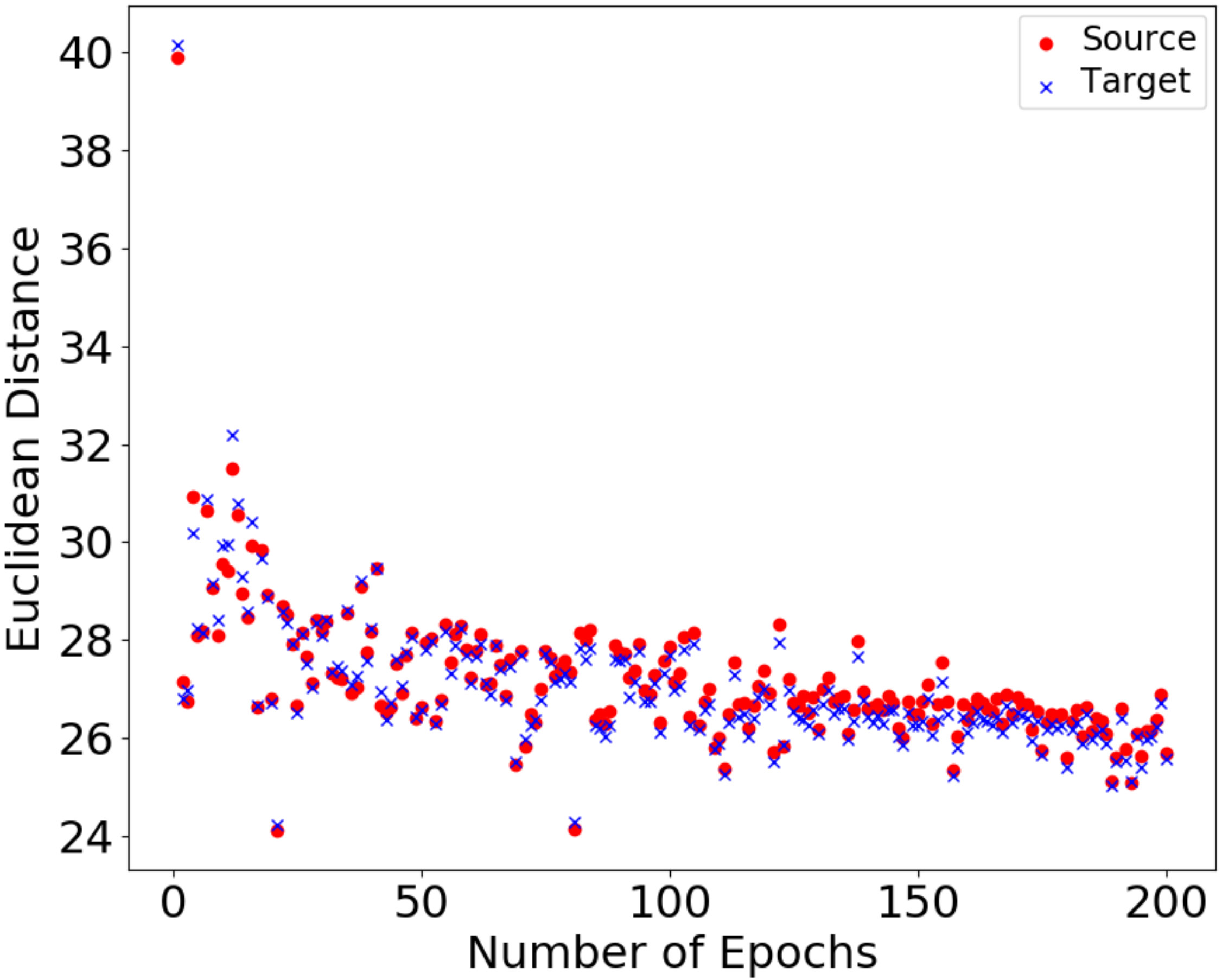}
			\label{fig:l2_dist_officehome:subfig10}
		\end{minipage}
	}%
	\subfloat[\footnotesize InsMean-to-Centroid (\textbf{Rw}$\rightarrow$\textbf{Ar})]{
		\begin{minipage}[t]{0.33\textwidth}
			\centering
			\includegraphics[height=1.6in]{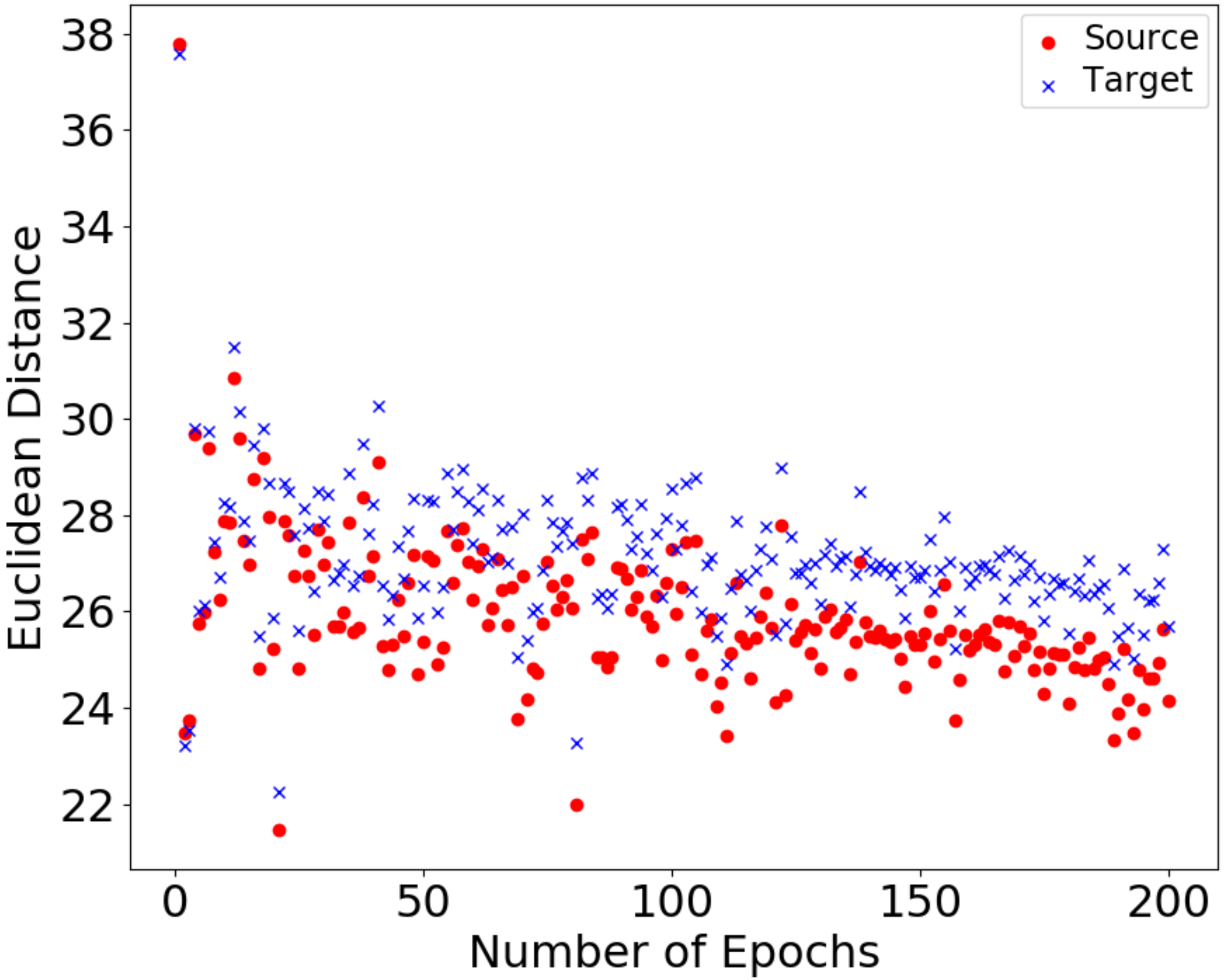}
			\label{fig:l2_dist_officehome:subfig11}
		\end{minipage}
	}%
	\subfloat[\footnotesize SrcInsMean-to-TgtInsMean (\textbf{Rw}$\rightarrow$\textbf{Ar})]{
		\begin{minipage}[t]{0.33\textwidth}
			\centering
			\includegraphics[height=1.6in]{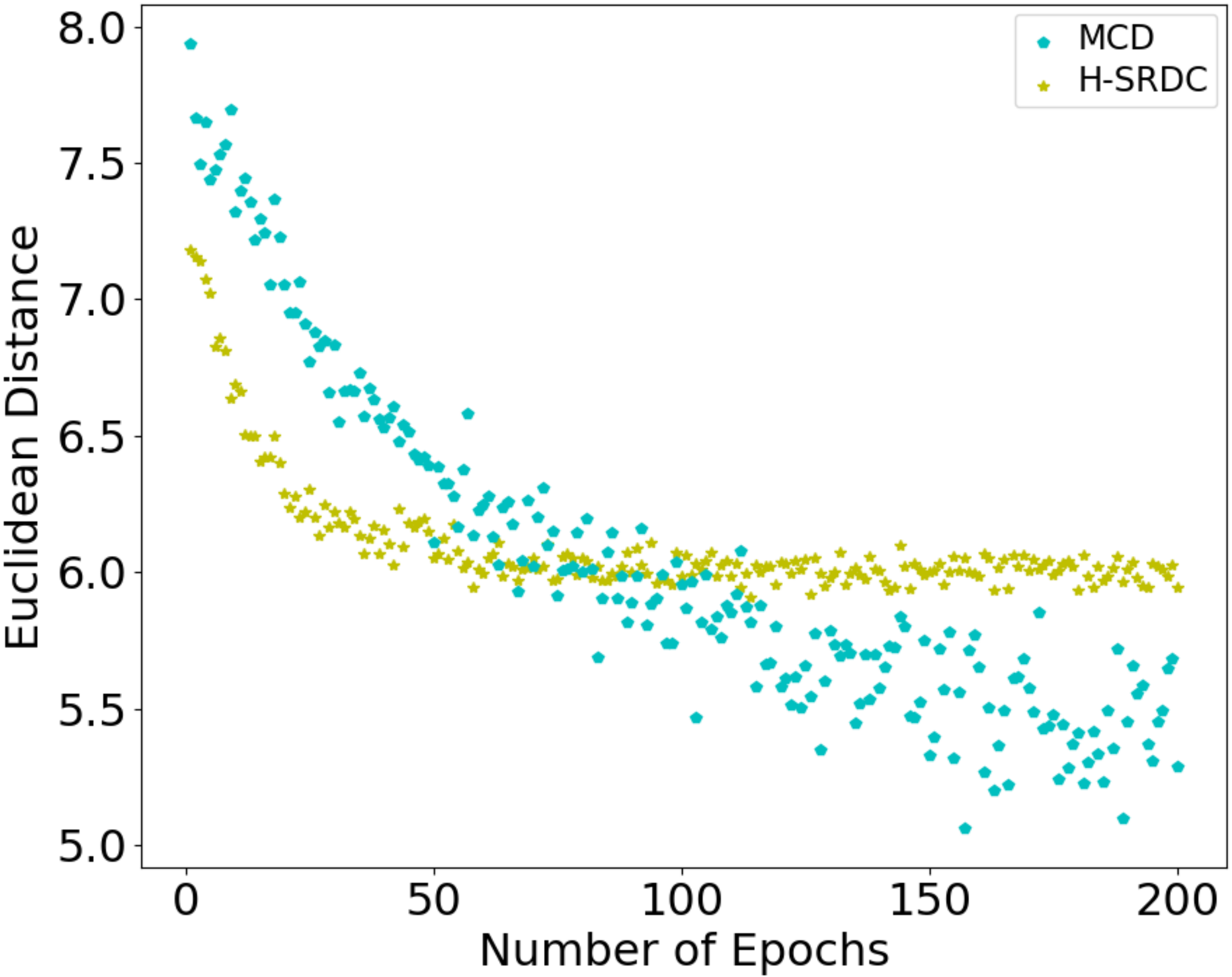}
			\label{fig:l2_dist_officehome:subfig12}
		\end{minipage}
	}%
	\caption{Learning diagnosis on the effect of the SRGenC objective (12) used in H-SRDC. Six types of distances for the source and target data are plotted against the training epochs. Comparisons between our H-SRDC and MCD \cite{mcd} are made in the sixth and last subfigures. The experiments are conducted on the adaptation tasks of \textbf{Ar}$\to$\textbf{Rw} and \textbf{Rw}$\to$\textbf{Ar} on the Office-Home benchmark \cite{officehome}. Refer to the main text for how these distances are defined and computed. 
	}
	\label{fig:l2_dist_officehome}
\end{figure*}

\begin{figure*}[!t]
	\centering
	\subfloat[\footnotesize Instance-to-Center (\textbf{Synthetic}$\rightarrow$\textbf{Real})]{
		\begin{minipage}[t]{0.31\textwidth}
			\centering
			\includegraphics[height=1.6in]{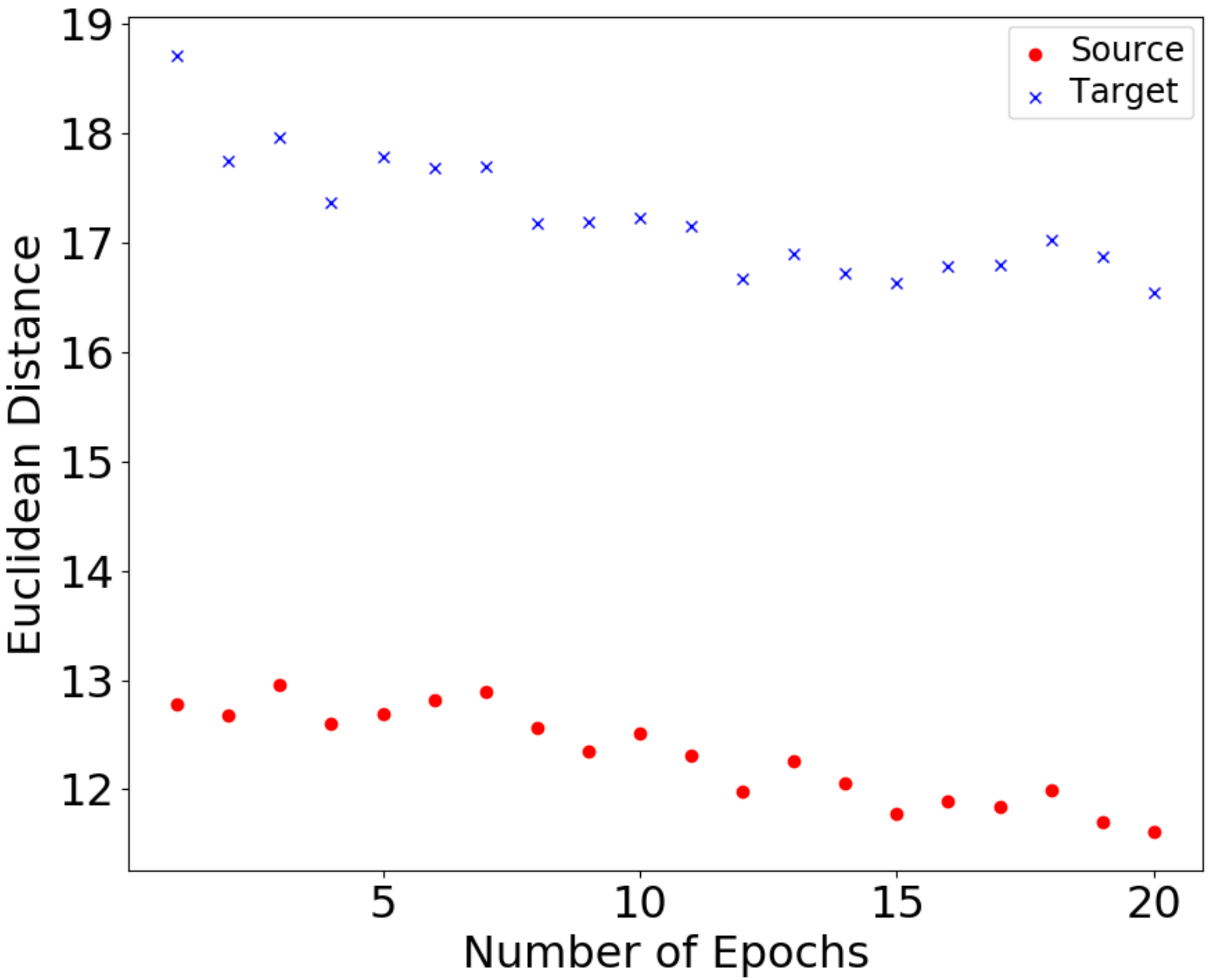}
			\label{fig:l2_dist_visda:subfig1}
		\end{minipage}
	}%
	\subfloat[\footnotesize InsMean-to-Center (\textbf{Synthetic}$\rightarrow$\textbf{Real})]{
		\begin{minipage}[t]{0.32\textwidth}
			\centering
			\includegraphics[height=1.6in]{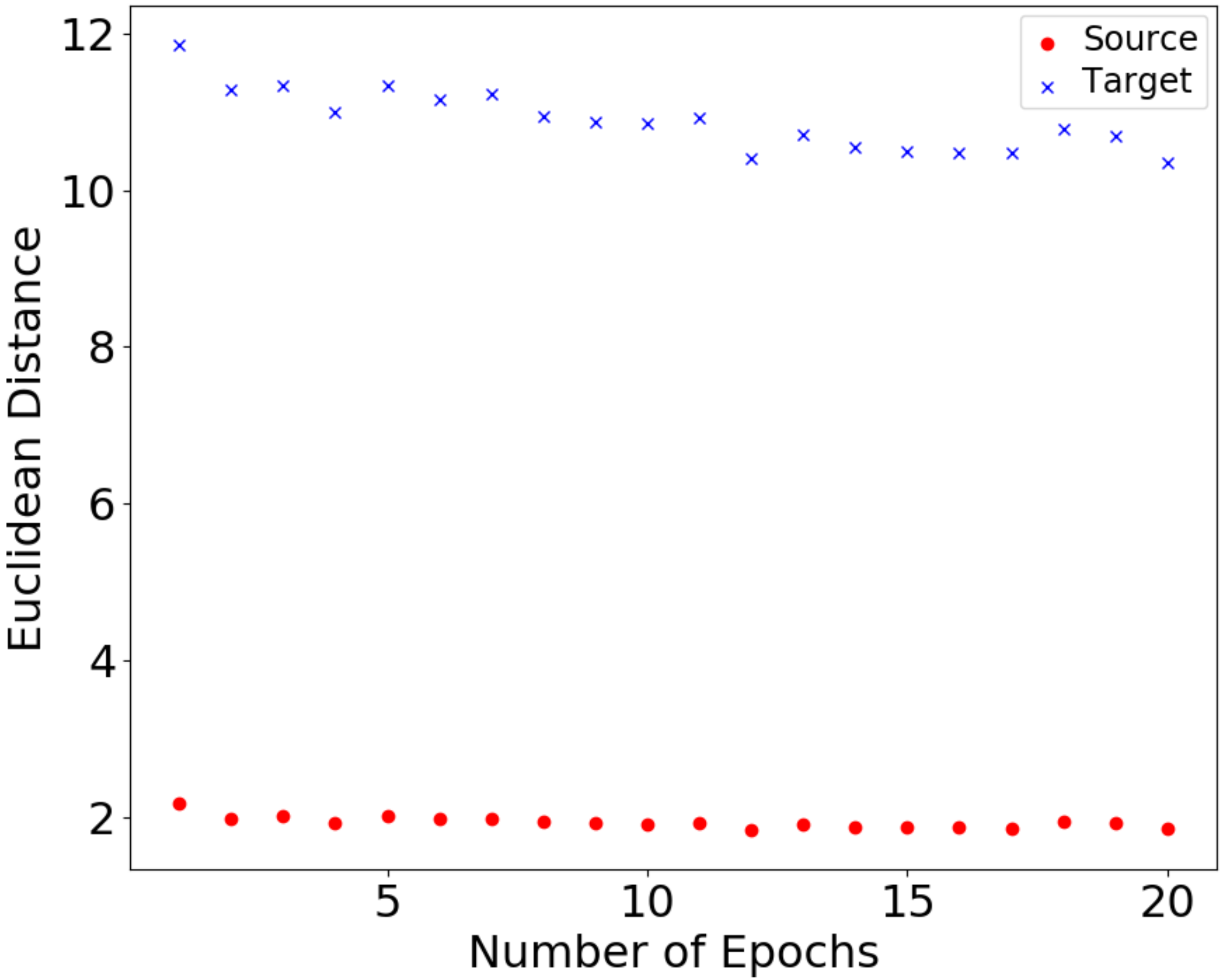}
			\label{fig:l2_dist_visda:subfig2}
		\end{minipage}
	}%
	\subfloat[\footnotesize Instance-to-InsMean (\textbf{Synthetic}$\rightarrow$\textbf{Real})]{
		\begin{minipage}[t]{0.34\textwidth}
			\centering
			\includegraphics[height=1.6in]{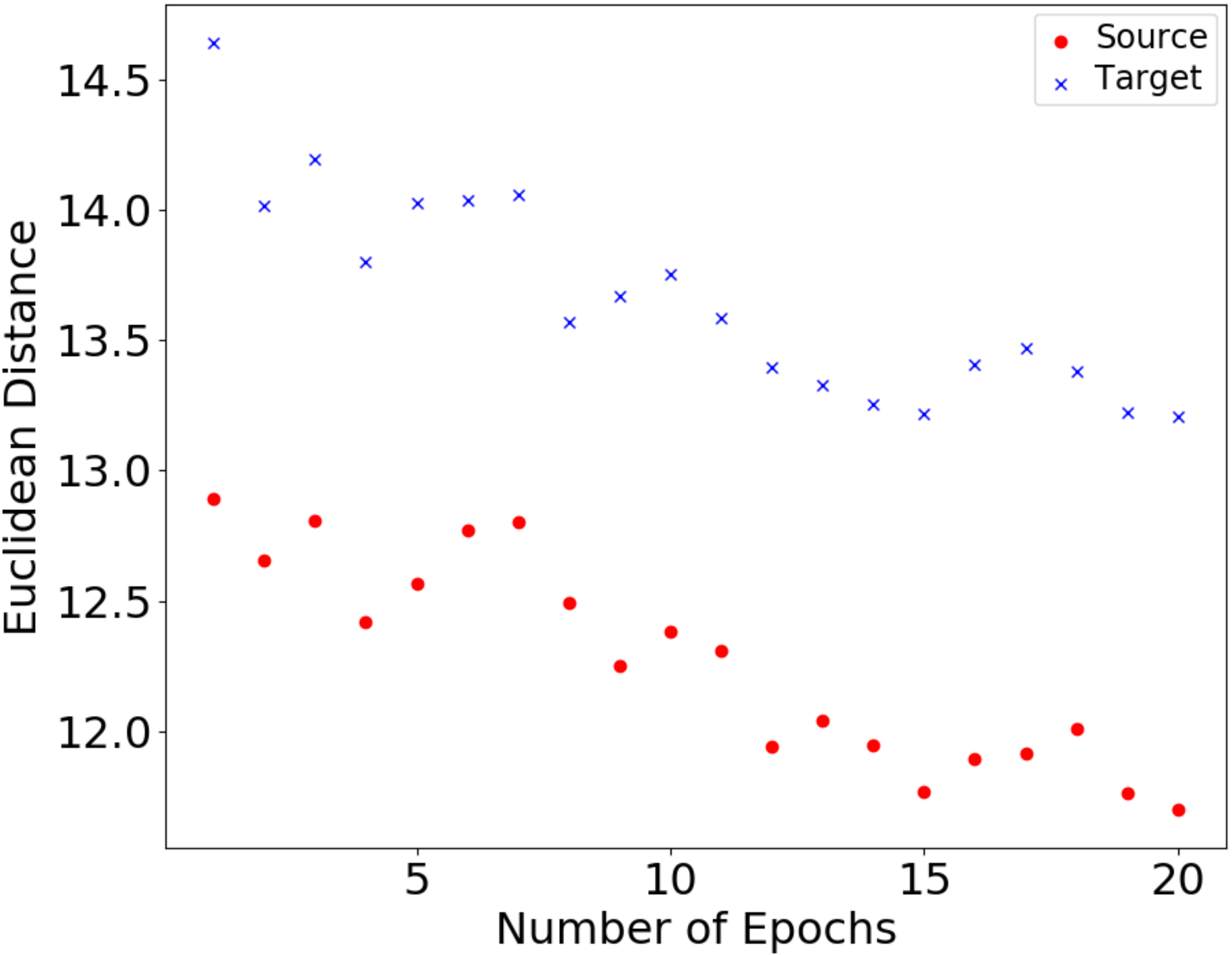}
			\label{fig:l2_dist_visda:subfig3}
		\end{minipage}
	}%
	\\
	\subfloat[\footnotesize Instance-to-Centroid (\textbf{Synthetic}$\rightarrow$\textbf{Real}) ]{
		\begin{minipage}[t]{0.31\textwidth}
			\centering
			\includegraphics[height=1.6in]{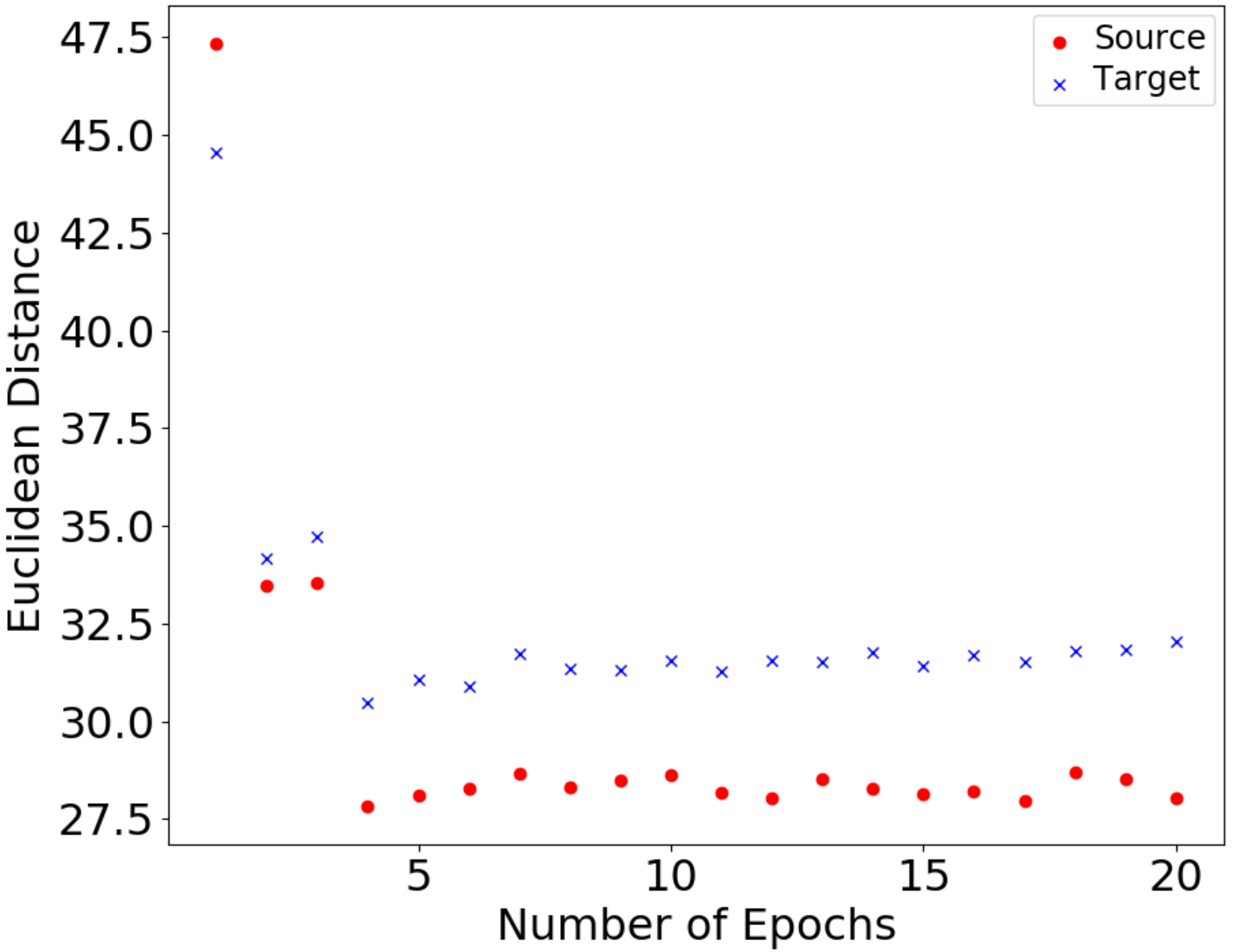}
			\label{fig:l2_dist_visda:subfig4}
		\end{minipage}
	}%
	\subfloat[\footnotesize InsMean-to-Centroid (\textbf{Synthetic}$\rightarrow$\textbf{Real})]{
		\begin{minipage}[t]{0.32\textwidth}
			\centering
			\includegraphics[height=1.6in]{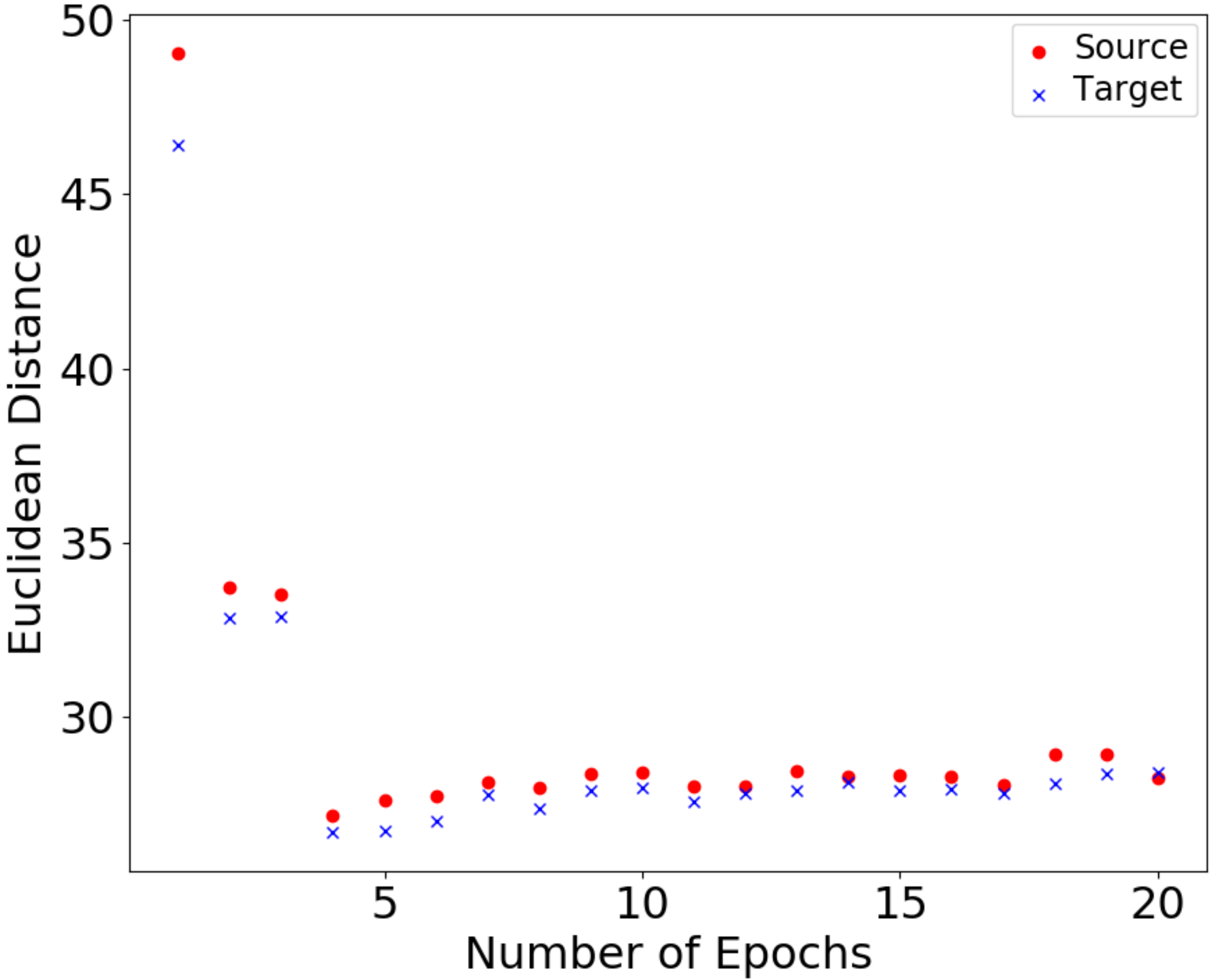}
			\label{fig:l2_dist_visda:subfig5}
		\end{minipage}
	}%
	\subfloat[\footnotesize SrcInsMean-to-TgtInsMean (\textbf{Synthetic}$\rightarrow$\textbf{Real})]{
		\begin{minipage}[t]{0.34\textwidth}
			\centering
			\includegraphics[height=1.6in]{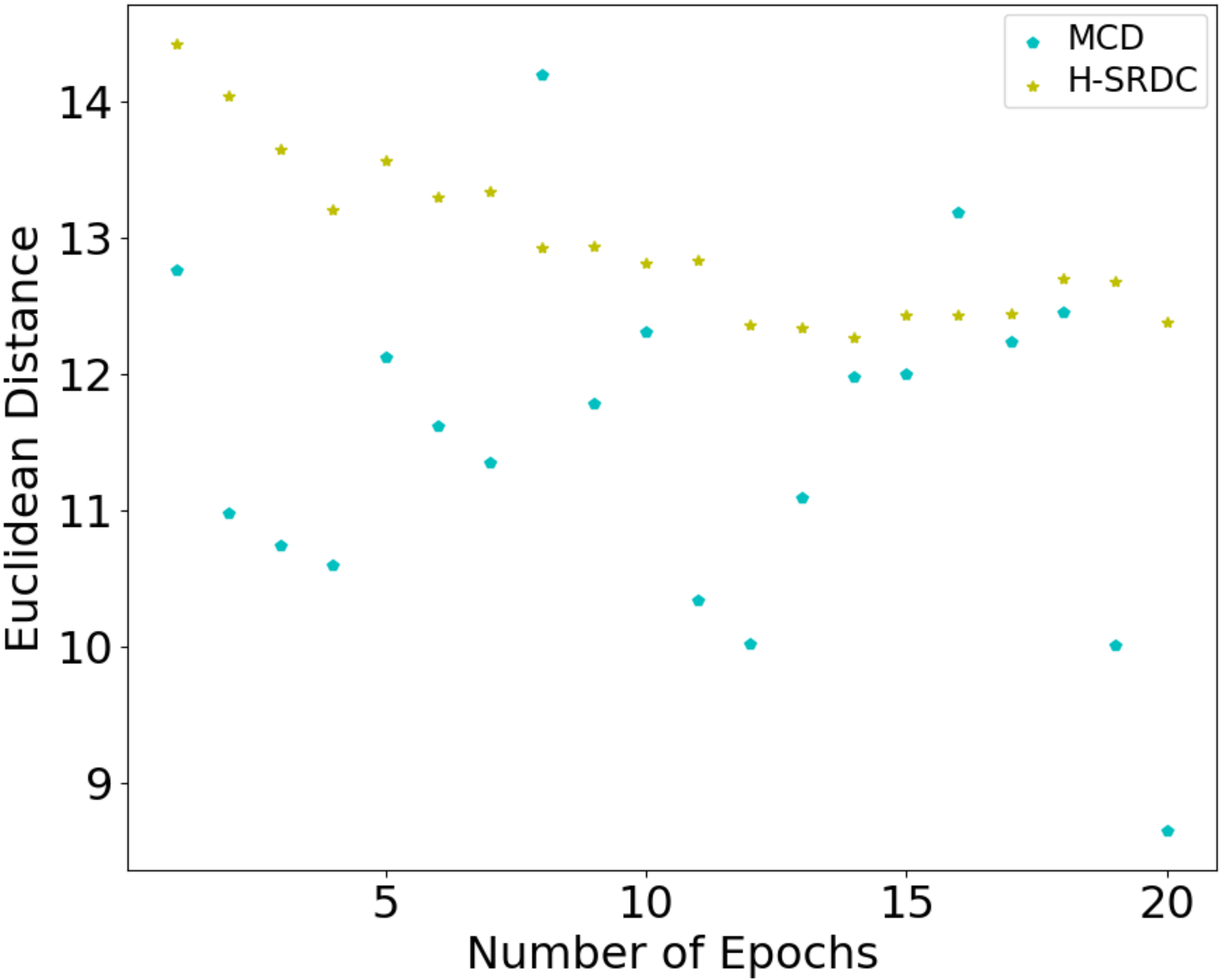}
			\label{fig:l2_dist_visda:subfig6}
		\end{minipage}
	}%
	\caption{Learning diagnosis on the effect of the SRGenC objective (12) used in H-SRDC. Six types of distances for the source and target data are plotted against the training epochs. Comparison between our H-SRDC and MCD \cite{mcd} is made in the last subfigure. The experiment is conducted on the adaptation task of \textbf{Synthetic}$\to$\textbf{Real} on the VisDA-2017 benchmark \cite{visda2017}. Refer to the main text for how these distances are defined and computed. 
	}
	\label{fig:l2_dist_visda}
\end{figure*}

Additional learning diagnoses on Office-Home \cite{officehome} and VisDA-2017 \cite{visda2017} are presented in Fig. \ref{fig:l2_dist_officehome} and Fig. \ref{fig:l2_dist_visda} respectively. Apart from the three types of distances described in Section 5.1 in the main text, we also show how the following three types of distances in the feature space evolve during training; they are the Euclidean distance between each instance and its class center ($\bm{\mu}_k$) computed on the combined, labeled source and pseudo-labeled target data, averaged over all the instances (dubbed as Instance-to-Center), the distance between the class mean of instances and their center, averaged over all the classes (dubbed as InsMean-to-Center), and the distance between each instance and its class mean, averaged over all the instances (dubbed as Instance-to-InsMean). 
Experiments are conducted on the adaptation tasks of \textbf{Ar}$\to$\textbf{Rw} and \textbf{Rw}$\to$\textbf{Ar} on the Office-Home benchmark, and \textbf{Synthetic}$\to$\textbf{Real} on the VisDA-2017 benchmark. We highlight the main observations below. 
{\bf (1)} For both the source and target domains, the distances of Instance-to-Centroid and InsMean-to-Centroid generally decrease or increase first, and then stabilize at certain levels during the training process, suggesting that our method does not enforce either source or target instances to collapse to the learned cluster centroids. 
{\bf (2)} The SrcInsMean-to-TgtInsMean distance of MCD decreases with the training, while that of our H-SRDC does not, 
indicating that our method is indeed modulating the feature space learning towards uncovering the intrinsic discrimination of target data, rather than aligning the features across domains. 
{\bf (3)} For both the source and target domains, Instance-to-Center, InsMean-to-Center, and Instance-to-InsMean first decrease or increase, and then stabilize at certain levels of distances with the training, suggesting that our method does not enforce either source or target instances to collapse to either the computed class centers or the class means of instances. 
{\bf (4)} In the corresponding two subfigures of Instance-to-Center and InsMean-to-Center, shifts between source and target are either similar or discrepant, since 
mathematically, the Instance-to-Centroid distance (averaged over all the instances) could be equal to, larger or smaller than the InsMean-to-Centroid distance (averaged over all the classes). 
{\bf (5)} Instance-to-Centroid and InsMean-to-Centroid are in a higher range of distances than Instance-to-Center and InsMean-to-Center respectively, indicating that the cluster centroid learner does not learn the computed class centers, but instead modulates the feature space learning towards uncovering the intrinsic discrimination of target data.

\subsection{Analysis of Convergence and Generalization}

\begin{figure*}[!t]
	\centering
	\subfloat[\textbf{Synthetic}$\to$\textbf{Real}]{
		\begin{minipage}[t]{0.49\linewidth}
			\centering
			\includegraphics[height=2.6in]{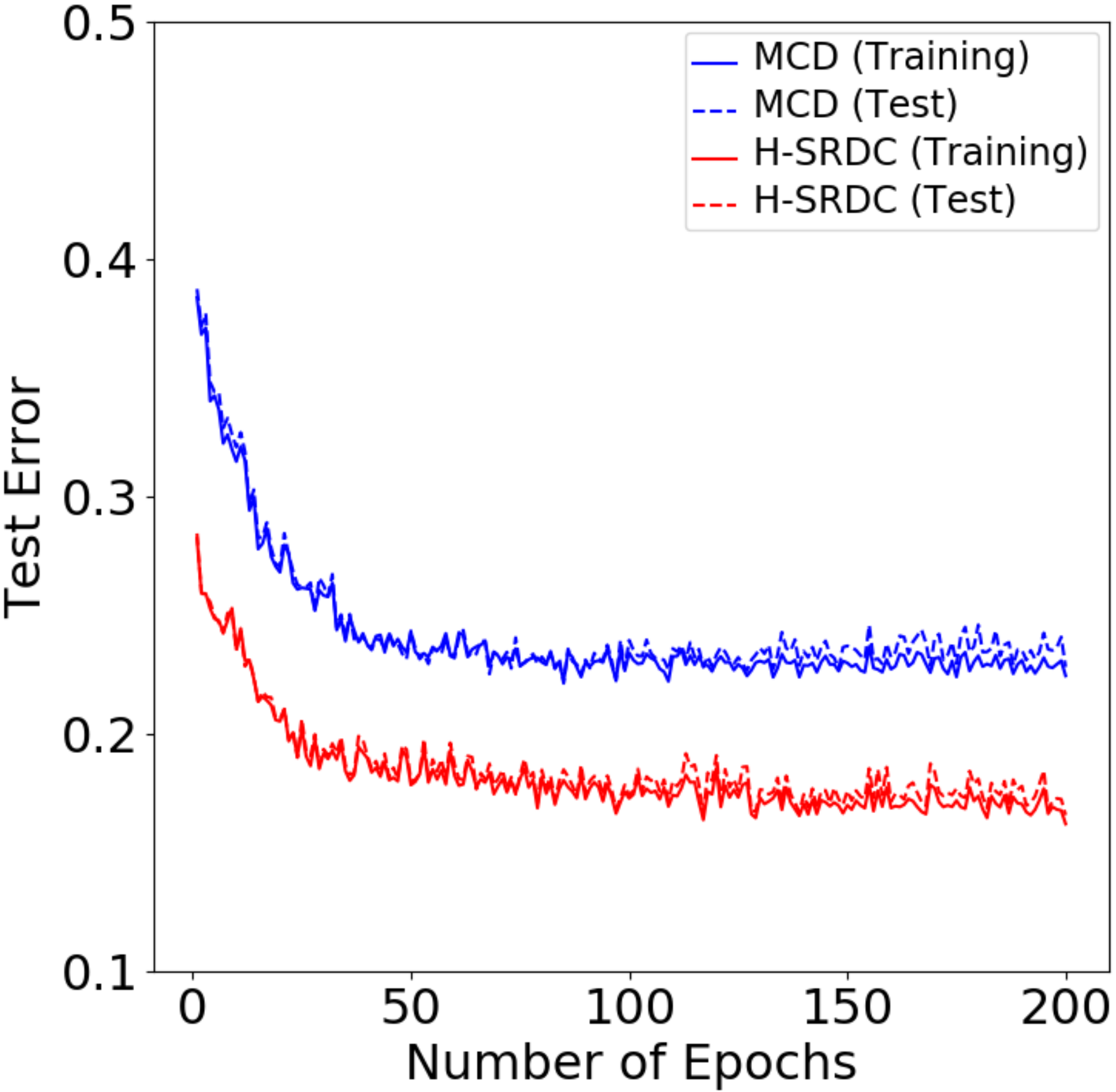}
			\label{fig:convergence_visda:subfig1}
		\end{minipage}
	}%
	\subfloat[\textbf{Synthetic}$\to$\textbf{Real}]{
		\begin{minipage}[t]{0.49\linewidth}
			\centering
			\includegraphics[height=2.6in]{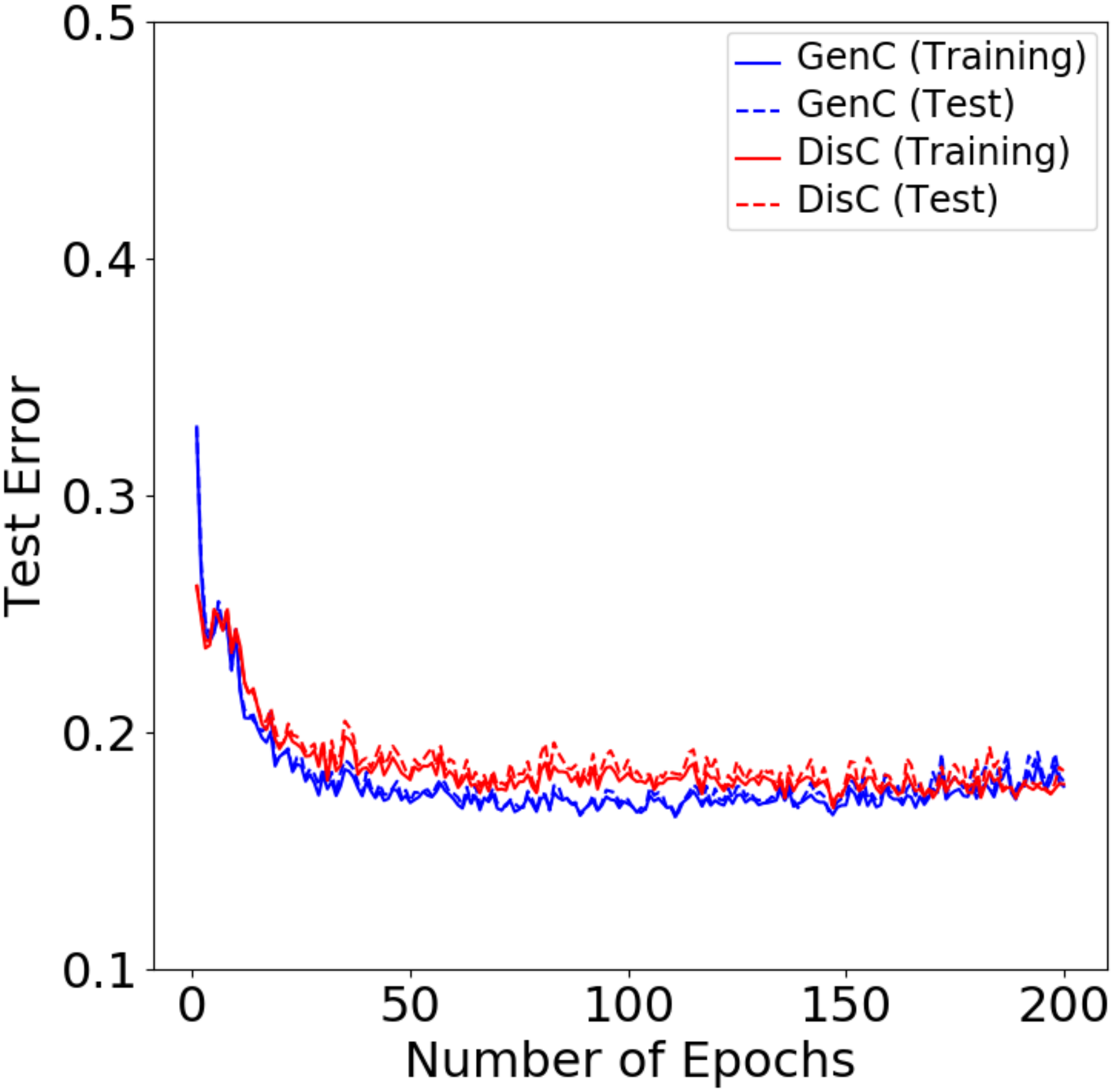}
			\label{fig:convergence_visda:subfig2}
		\end{minipage}
	}%
	\caption{Analysis of convergence and generalization. ``Training'' and ``Test'' refer to results on training and held-out test sets on the target domain, respectively. In (b), ``GenC'' and ``DisC'' refer to results of generative and discriminative clusterings during the training of H-SRDC, respectively. Experiment in the inductive UDA setting is conducted on the adaptation task of \textbf{Synthetic}$\to$\textbf{Real} on the VisDA-2017 benchmark \cite{visda2017} (10 tests per training epoch). 
	}
	\label{fig:convergence_visda}
\end{figure*}

In Fig. \ref{fig:convergence_visda}, we show the curves of convergence and generalization for the adaptation task of \textbf{Synthetic}$\to$\textbf{Real} on the VisDA-2017 benchmark. We observe that H-SRDC has much lower test errors than MCD, confirming the better generalization of our method.

In Fig. \ref{fig:convergence_visda}-(b), we show the convergence and generalization curves of H-SRDC in terms of both GenC and DisC results on the realistically significant task of \textbf{Synthetic}$\rightarrow$\textbf{Real} on VisDA-2017. It is observed that the DisC errors decrease and gradually approach the GenC ones with the training, suggesting the complementary effects of DisC and GenC in H-SRDC.

\section{Explanation on Algorithm Details}

In Algorithm 1 in the main text, steps 9-11 mean that the auxiliary distributions $\bm{Q}_{batch}^t$ and $\widetilde{\bm{Q}}_{batch}^t$ start to be updated from the second epoch. The reasons are twofold: {\bf (1)} at the first epoch, the class discrimination information from the source domain has not been encoded in the model, and consequently, the resulting target auxiliary distributions are noisy and unreliable; {\bf (2)} in step 1, we initialize $\{\{q_{i,k}^t=\widetilde{q}_{i,k}^t={\rm I}[k=\hat{y}_i^t]\}_{k=1}^K\}_{i=1}^{n_t}$ by k-means clustering and such discriminative information from the target domain should be exploited. Step 14 indeed includes updating the assignment of pseudo labels $\{\hat{y}_i^t\}_{i=1}^{n_t}$.

\section{More Insights on Layout-wise Consistency}


Curriculum domain adaptation \cite{curriculum_da} models the semantic layout by both the global label distributions over images and the local ones of landmark superpixels; it uses the layout similarity across domains as a priori knowledge. Accordingly, it first predicts both label distributions for the unlabeled target data by generators trained on the labeled source data, and then learns a segmentation model by pixel-wise classification of source images, together with global and local label-distribution regularization of target images. 
Our used layout regularization (\ie, matching distributions over weighted self-information maps across domains in an adversarial manner \cite{advent}), shares the same motivation with \cite{curriculum_da}, but differs in the aim to explicitly transfer source domain knowledge on semantic layout and enforce prediction consistency in local neighborhoods of target images. Our approach is thus in a distinctive perspective of implementing the assumed layout-wise similarity in a structured output space. 

\section{More Comparative Results in a Transductive Setting}
Comparative results on individual adaptation tasks of Office-31 \cite{office31}, ImageCLEF-DA \cite{imageclefda}, Office-Home \cite{officehome}, and VisDA-2017 \cite{visda2017} are shown in Tables \ref{table:results_office31}, \ref{table:results_imageclefda}, \ref{table:results_officehome}, and \ref{table:results_visda2017} respectively.

\begin{table*}[!t]
	\begin{center}
		\caption{Comparative results (\%) in the {\bf transductive} setting on the Office-31 benchmark \cite{office31}. All methods are based on the base model of ResNet-50. 
		}
		\label{table:results_office31}
		\begin{tabular}{|l|c|c|c|c|c|c|c|}
			\hline
			Method                & A $\rightarrow$ W & D $\rightarrow$ W & W $\rightarrow$ D & A $\rightarrow$ D & D $\rightarrow$ A & W $\rightarrow$ A & \em mean \\
			
			\hline
			\hline
			Source Only        & 77.8$\pm$0.2 & 96.9$\pm$0.1 & 99.3$\pm$0.1 & 82.1$\pm$0.2 & 64.5$\pm$0.2 & 66.1$\pm$0.2 & 81.1 \\
			
			
			DAN \cite{dan}          & 81.3$\pm$0.3 & 97.2$\pm$0.0 & 99.8$\pm$0.0 & 83.1$\pm$0.2 & 66.3$\pm$0.0 & 66.3$\pm$0.1 & 82.3 \\
			
			DANN \cite{dann}       & 81.7$\pm$0.2 & 98.0$\pm$0.2 & 99.8$\pm$0.0 & 83.9$\pm$0.7 & 66.4$\pm$0.2 & 66.0$\pm$0.3 & 82.6 \\
			
			
			
			
			VADA \cite{dirt_t}              & 86.5$\pm$0.5 & 98.2$\pm$0.4 & 99.7$\pm$0.2 & 86.7$\pm$0.4 & 70.1$\pm$0.4 & 70.5$\pm$0.4 & 85.4 \\
			
			
			ETD \cite{etd}          & 92.1 & \textbf{100.0} & \textbf{100.0} & 88.0 & 71.0 & 67.8 & 86.2 \\
			
			
			
			MCD \cite{mcd}                  & 88.6$\pm$0.2 & 98.5$\pm$0.1 & \textbf{100.0}$\pm$0.0 & 92.2$\pm$0.2 & 69.5$\pm$0.1 & 69.7$\pm$0.3 & 86.5 \\
			
			SAFN+ENT \cite{larger_norm}   & 90.1$\pm$0.8 & 98.6$\pm$0.2 & 99.8$\pm$0.0 & 90.7$\pm$0.5 & 73.0$\pm$0.2 & 70.2$\pm$0.3 & 87.1 \\
			
			
			
			rRevGrad+CAT \cite{cat}   & 94.4$\pm$0.1 & 98.0$\pm$0.2 & \textbf{100.0}$\pm$0.0 & 90.8$\pm$1.8 & 72.2$\pm$0.6 & 70.2$\pm$0.1 & 87.6 \\ 
			
			CDAN+E \cite{cdan}  & 94.1$\pm$0.1 & 98.6$\pm$0.1 & \textbf{100.0}$\pm$0.0 & 92.9$\pm$0.2 & 71.0$\pm$0.3 & 69.3$\pm$0.3 & 87.7 \\
			
			
			
			
			SymNets \cite{symnets}          & 90.8$\pm$0.1 & 98.8$\pm$0.3 & \textbf{100.0}$\pm$0.0 & 93.9$\pm$0.5 & 74.6$\pm$0.6 & 72.5$\pm$0.5 & 88.4 \\ 
			
			BSP+CDAN \cite{bsp}             & 93.3$\pm$0.2 & 98.2$\pm$0.2 & \textbf{100.0}$\pm$0.0 & 93.0$\pm$0.2 & 73.6$\pm$0.3 & 72.6$\pm$0.3 & 88.5 \\
			
			
			CDAN+BNM \cite{bnm}             & 92.8 & 98.8 & \textbf{100.0} & 92.9 & 73.5 & 73.8 & 88.6 \\
			
			
			CADA-P \cite{cada}              & \textbf{97.0}$\pm$0.2 & 99.3$\pm$0.1 & \textbf{100.0}$\pm$0.0 & 95.6$\pm$0.1 & 71.5$\pm$0.2 & 73.1$\pm$0.3 & 89.5 \\ 
			
			CAN \cite{can}                  & 94.5$\pm$0.3 & 99.1$\pm$0.2 & 99.8$\pm$0.2 & 95.0$\pm$0.3 & \textbf{78.0}$\pm$0.3 & 77.0$\pm$0.3 & 90.6 \\ 
			
			\hline
			SRDC \cite{srdc}        & 95.7$\pm$0.2 & 99.2$\pm$0.1 & \textbf{100.0}$\pm$0.0 & \textbf{95.8}$\pm$0.2 & 76.7$\pm$0.3 & 77.1$\pm$0.1 & 90.8 \\
			
			\bf \name{} 			& 96.2$\pm$0.1 & 99.2$\pm$0.1 & \textbf{100.0}$\pm$0.0 & 95.4$\pm$0.3 & 77.1$\pm$0.2 & \textbf{77.4}$\pm$0.2 & \textbf{90.9} \\			
			\hline
		\end{tabular}
	\end{center}
\end{table*}

\begin{table*}[!t]
	\begin{center}
		\caption{Comparative results (\%) in the {\bf transductive} setting on the ImageCLEF-DA benchmark \cite{imageclefda}. All methods are based on the base model of ResNet-50.  
		}
		\label{table:results_imageclefda}
		\begin{tabular}{|l|c|c|c|c|c|c|c|}
			\hline
			Methods                 & I $\rightarrow$ P & P $\rightarrow$ I & I $\rightarrow$ C & C $\rightarrow$ I & C $\rightarrow$ P & P $\rightarrow$ C & \em mean \\
			\hline
			\hline
			Source Only        & 74.8$\pm$0.3 & 83.9$\pm$0.1 & 91.5$\pm$0.3 & 78.0$\pm$0.2 & 65.5$\pm$0.3 & 91.2$\pm$0.3 & 80.7 \\
			
			DAN \cite{dan}                   & 74.5$\pm$0.4 & 82.2$\pm$0.2 & 92.8$\pm$0.2 & 86.3$\pm$0.4 & 69.2$\pm$0.4 & 89.8$\pm$0.4 & 82.5 \\
			
			
			DANN \cite{dann}                 & 75.0$\pm$0.6 & 86.0$\pm$0.3 & 96.2$\pm$0.4 & 87.0$\pm$0.5 & 74.3$\pm$0.5 & 91.5$\pm$0.6 & 85.0 \\
			
			
			
			rRevGrad+CAT \cite{cat}          & 77.2$\pm$0.2 & 91.0$\pm$0.3 & 95.5$\pm$0.3 & 91.3$\pm$0.3 & 75.3$\pm$0.6 & 93.6$\pm$0.5 & 87.3 \\
			
			
			CDAN+E \cite{cdan}               & 77.7$\pm$0.3 & 90.7$\pm$0.2 & 97.7$\pm$0.3 & 91.3$\pm$0.3 & 74.2$\pm$0.2 & 94.3$\pm$0.3 & 87.7 \\
			
			CAN \cite{can}                   & 77.2$\pm$0.6 & 90.3$\pm$0.5 & 96.0$\pm$0.2 & 90.9$\pm$0.3 & 78.0$\pm$0.6 & 95.6$\pm$0.6 & 88.0 \\
			
			CADA-P \cite{cada}               & 78.0 & 90.5 & 96.7 & 92.0 & 77.2 & 95.5 & 88.3 \\
			
			
			SAFN+ENT \cite{larger_norm}      & 79.3$\pm$0.1 & 93.3$\pm$0.4 & 96.3$\pm$0.4 & 91.7$\pm$0.0 & 77.6$\pm$0.1 & 95.3$\pm$0.1 & 88.9 \\
			
			ETD \cite{etd}                   & 81.0 & 91.7 & \textbf{97.9} & 93.3 & 79.5 & 95.0 & 89.7 \\
			
			SymNets \cite{symnets}           & 80.2$\pm$0.3 & 93.6$\pm$0.2 & 97.0$\pm$0.3 & 93.4$\pm$0.3 & 78.7$\pm$0.3 & 96.4$\pm$0.1 & 89.9 \\
			
			\hline
			SRDC \cite{srdc}    & 80.8$\pm$0.3 & 94.7$\pm$0.2 & 97.8$\pm$0.2 & 94.1$\pm$0.2 & 80.0$\pm$0.3 & 97.7$\pm$0.1 & 90.9 \\
			
			\bf \name{}			& \textbf{81.2}$\pm$0.3 & \textbf{95.0}$\pm$0.2 & 97.7$\pm$0.3 & \textbf{94.3}$\pm$0.1 & \textbf{80.5}$\pm$0.3 & \textbf{98.3}$\pm$0.2 & \textbf{91.2} \\
			
			\hline
		\end{tabular}
	\end{center}
\end{table*}

\begin{table*}[!t]
	\begin{center}
		\caption{Comparative results (\%) in the {\bf transductive} setting on the Office-Home benchmark \cite{officehome}. All methods are based on the base model of ResNet-50.  
		}
		\label{table:results_officehome}
		\resizebox{1.0\textwidth}{!}{
			\begin{tabular}{|l|c|c|c|c|c|c|c|c|c|c|c|c|c|}
				\hline
				Methods                         & Ar$\rightarrow$Cl & Ar$\rightarrow$Pr & Ar$\rightarrow$Rw & Cl$\rightarrow$Ar & Cl$\rightarrow$Pr & Cl$\rightarrow$Rw & Pr$\rightarrow$Ar & Pr$\rightarrow$Cl & Pr$\rightarrow$Rw & Rw$\rightarrow$Ar & Rw$\rightarrow$Cl & Rw$\rightarrow$Pr    & \em mean  \\
				\hline
				\hline
				Source Only    & 34.9      & 50.0     & 58.0      & 37.4      & 41.9      & 46.2     & 38.5     & 31.2     & 60.4     & 53.9     & 41.2     & 59.9 & 46.1 \\
				
				DAN \cite{dan}                  & 43.6     & 57.0     & 67.9      & 45.8      & 56.5      & 60.4     & 44.0     & 43.6     & 67.7     & 63.1     & 51.5     & 74.3  & 56.3 \\
				
				DANN \cite{dann}                & 45.6     & 59.3     & 70.1      & 47.0      & 58.5      & 60.9     & 46.1     & 43.7     & 68.5     & 63.2     & 51.8      & 76.8 & 57.6 \\
				
				
				
				
				CDAN+E \cite{cdan}              & 50.7     & 70.6     & 76.0     & 57.6       & 70.0      & 70.0     & 57.4     & 50.9     & 77.3      & 70.9      & 56.7     & 81.6 & 65.8 \\
				
				
				BSP+CDAN \cite{bsp}             & 52.0 & 68.6 & 76.1 & 58.0 & 70.3 & 70.2 & 58.6 & 50.2 & 77.6 & 72.2 & 59.3 & 81.9 & 66.3 \\
				
				SAFN \cite{larger_norm}         & 52.0 & 71.7 & 76.3 & 64.2 & 69.9 & 71.9 & 63.7 & 51.4 & 77.1 & 70.9 & 57.1 & 81.5 & 67.3 \\
				
				ETD \cite{etd}          & 51.3 & 71.9 & \textbf{85.7} & 57.6 & 69.2 & 73.7 & 57.8 & 51.2 & 79.3 & 70.2 & 57.5 & 82.1 & 67.3 \\
				
				
				SymNets \cite{symnets} & 47.7 & 72.9 & 78.5 & 64.2 & 71.3 & 74.2 & 64.2 & 48.8 & 79.5 & 74.5 & 52.6 & 82.7 & 67.6 \\
				
				
				CAN \cite{can}         & \textbf{58.5} & 75.3 & 75.1 & 61.7 & 74.5 & 70.1 & 61.3 & 54.6 & 75.9 & 72.4 & 58.3 & 82.4 & 68.3 \\
				
				
				CDAN+BNM \cite{bnm}    & 56.2 & 73.7 & 79.0 & 63.1 & 73.6 & 74.0 & 62.4 & 54.8 & 80.7 & 72.4 & 58.9 & 83.5 & 69.4 \\
				
				CADA-P \cite{cada}     & 56.9 & 76.4 & 80.7 & 61.3 & 75.2 & 75.2 & 63.2 & 54.5 & 80.7 & 73.9 & \textbf{61.5} & 84.1 & 70.2 \\
				
				\hline
				SRDC \cite{srdc}      & 52.3 & 76.3 & 81.0 & 69.5 & 76.2 & \textbf{78.0} & 68.7 & 53.8 & 81.7 & 76.3 & 57.1 & 85.0 & 71.3 \\
				
				\bf \name{}            & 58.4 & \textbf{77.5} & 81.3 & \textbf{69.7} & \textbf{76.5} & 77.2 & \textbf{68.9} & \textbf{56.9} & \textbf{82.0} & \textbf{76.4} & 61.0 & \textbf{85.2} & \textbf{72.6} \\
				\hline
			\end{tabular}
		}
	\end{center}
\end{table*}

\begin{table*}[!t]
	\begin{center}
		\caption{Comparative results (\%) in the {\bf transductive} setting on the VisDA-2017 benchmark \cite{visda2017}. All methods are based on the base model of ResNet-101.  
		}
		\label{table:results_visda2017}
		\begin{tabular}{|l|c|c|c|c|c|c|c|c|c|c|c|c|c|}
			\hline
			Methods                & plane & bcycl & bus & car & horse & knife & mcycl & person & plant & sktbrd & train & truck &\em mean \\
			\hline
			\hline
			
			Source Only      & 55.1 & 53.3 & 61.9 & 59.1 & 80.6 & 17.9 & 79.7 & 31.2 & 81.0 & 26.5 & 73.5 & 8.5 & 52.4 \\
			
			DANN \cite{dann}  & 81.9 & 77.7 & 82.8 & 44.3 & 81.2 & 29.5 & 65.1 & 28.6 & 51.9 & 54.6 & 82.8 & 7.8 & 57.4 \\
			
			DAN \cite{dan}         & 87.1 & 63.0 & 76.5 & 42.0 & 90.3 & 42.9 & 85.9 & 53.1 & 49.7 & 36.3 & 85.8 & 20.7 & 61.1 \\
			
			MCD \cite{mcd} & 87.0 & 60.9 & \textbf{83.7} & 64.0 & 88.9 & 79.6 & 84.7 & 76.9 & 88.6 & 40.3 & 83.0 & 25.8 & 71.9 \\
			
			GPDA \cite{gpda}  & 83.0 & 74.3 & 80.4 & 66.0 & 87.6 & 75.3 & 83.8 & 73.1 & 90.1 & 57.3 & 80.2 & 37.9 & 73.3 \\
			
			
			BSP+CDAN \cite{bsp} & 92.4 & 61.0 & 81.0 & 57.5 & 89.0 & 80.6 & 90.1 & 77.0 & 84.2 & 77.9 & 82.1 & 38.4 & 75.9 \\
			
			
			
			CAN \cite{can} & \textbf{97.0} & 87.2 & 82.5 & 74.3 & \textbf{97.8} & \textbf{96.2} & 90.8 & 80.7 & 96.6 & \textbf{96.3} & 87.5 & \textbf{59.9} & 87.2 \\
			
			\hline
			
			SRDC \cite{srdc} & 96.7 & 81.2 & 81.7 & \textbf{91.1} & 97.0 & 93.7 & \textbf{94.3} & 83.2 & \textbf{97.6} & 92.4 & \textbf{89.4} & 37.2 & 86.3 \\
			
			\bf \name{}	& 96.6 & \textbf{88.8} & \textbf{83.7} & 88.9 & 96.6 & 94.4 & 92.4 & \textbf{85.4} & 96.7 & 94.4 & 88.3 & 43.1 & \textbf{87.4} \\
			\hline
		\end{tabular}
	\end{center}
\end{table*}

%

%% file: main.bbl
\begin{thebibliography}{10}
\providecommand{\url}[1]{#1}
\csname url@samestyle\endcsname
\providecommand{\newblock}{\relax}
\providecommand{\bibinfo}[2]{#2}
\providecommand{\BIBentrySTDinterwordspacing}{\spaceskip=0pt\relax}
\providecommand{\BIBentryALTinterwordstretchfactor}{4}
\providecommand{\BIBentryALTinterwordspacing}{\spaceskip=\fontdimen2\font plus
\BIBentryALTinterwordstretchfactor\fontdimen3\font minus
  \fontdimen4\font\relax}
\providecommand{\BIBforeignlanguage}[2]{{%
\expandafter\ifx\csname l@#1\endcsname\relax
\typeout{** WARNING: IEEEtran.bst: No hyphenation pattern has been}%
\typeout{** loaded for the language `#1'. Using the pattern for}%
\typeout{** the default language instead.}%
\else
\language=\csname l@#1\endcsname
\fi
#2}}
\providecommand{\BIBdecl}{\relax}
\BIBdecl

\bibitem{transfer_learning_survey}
S.~J. {Pan} and Q.~{Yang}, ``A survey on transfer learning,'' \emph{IEEE Trans.
  Knowl. Data Eng.}, vol.~22, pp. 1345--1359, 2010.

\bibitem{da_theory2}
S.~Ben-David, J.~Blitzer, K.~Crammer, A.~Kulesza, F.~Pereira, and J.~W.
  Vaughan, ``A theory of learning from different domains,'' \emph{Machine
  Learning}, vol.~79, pp. 151--175, 2010.

\bibitem{survey_deep_vis_da}
M.~Wang and W.~Deng, ``Deep visual domain adaptation: A survey,''
  \emph{Neurocomputing}, vol. 312, pp. 135--153, 2018.

\bibitem{dan}
M.~{Long}, Y.~{Cao}, Z.~{Cao}, J.~{Wang}, and M.~I. {Jordan}, ``Transferable
  representation learning with deep adaptation networks,'' \emph{IEEE Trans.
  Pattern Anal. Mach. Intell.}, vol.~41, pp. 3071--3085, 2019.

\bibitem{dann}
Y.~Ganin, E.~Ustinova, H.~Ajakan, P.~Germain, H.~Larochelle, F.~Laviolette,
  M.~Marchand, and V.~Lempitsky, ``Domain-adversarial training of neural
  networks,'' \emph{J. Mach. Learn. Res.}, vol.~17, pp. 2096--2030, 2016.

\bibitem{mcd}
K.~{Saito}, K.~{Watanabe}, Y.~{Ushiku}, and T.~{Harada}, ``Maximum classifier
  discrepancy for unsupervised domain adaptation,'' in \emph{Proc. IEEE Conf.
  Comput. Vis. Pattern Recognit.}, 2018, pp. 3723--3732.

\bibitem{symnets}
Y.~{Zhang}, H.~{Tang}, K.~{Jia}, and M.~{Tan}, ``Domain-symmetric networks for
  adversarial domain adaptation,'' in \emph{Proc. IEEE Conf. Comput. Vis.
  Pattern Recognit.}, 2019, pp. 5026--5035.

\bibitem{pfan}
C.~{Chen}, W.~{Xie}, W.~{Huang}, Y.~{Rong}, X.~{Ding}, Y.~{Huang}, T.~{Xu}, and
  J.~{Huang}, ``Progressive feature alignment for unsupervised domain
  adaptation,'' in \emph{Proc. IEEE Conf. Comput. Vis. Pattern Recognit.},
  2019, pp. 627--636.

\bibitem{tpn}
Y.~{Pan}, T.~{Yao}, Y.~{Li}, Y.~{Wang}, C.~{Ngo}, and T.~{Mei}, ``Transferrable
  prototypical networks for unsupervised domain adaptation,'' in \emph{Proc.
  IEEE Conf. Comput. Vis. Pattern Recognit.}, 2019, pp. 2234--2242.

\bibitem{bnm}
S.~Cui, S.~Wang, J.~Zhuo, L.~Li, Q.~Huang, and Q.~Tian, ``Towards
  discriminability and diversity: Batch nuclear-norm maximization under label
  insufficient situations,'' in \emph{Proc. IEEE Conf. Comput. Vis. Pattern
  Recognit.}, 2020, pp. 3941--3950.

\bibitem{da_theory1}
S.~Ben-David, J.~Blitzer, K.~Crammer, and F.~Pereira, ``Analysis of
  representations for domain adaptation,'' in \emph{Proc. Neur. Info. Proc.
  Sys.}, 2007, pp. 137--144.

\bibitem{mansour09}
Y.~Mansour, M.~Mohri, and A.~Rostamizadeh, ``Domain adaptation: Learning bounds
  and algorithms,'' in \emph{The Conference on Learning Theory}, 2009.

\bibitem{da_theory3}
H.~Zhao, R.~T.~D. Combes, K.~Zhang, and G.~Gordon, ``On learning invariant
  representations for domain adaptation,'' in \emph{Proc. Int. Conf. Mach.
  Learn.}, vol.~97, 2019, pp. 7523--7532.

\bibitem{it_cluster_uda2}
Y.~Shi and F.~Sha, ``Information-theoretical learning of discriminative
  clusters for unsupervised domain adaptation,'' in \emph{Proc. Int. Conf.
  Mach. Learn.}, 2012, pp. 1275--1282.

\bibitem{InfoGAN}
X.~Chen, Y.~Duan, R.~Houthooft, J.~Schulman, I.~Sutskever, and P.~Abbeel,
  ``Infogan: Interpretable representation learning by information maximizing
  generative adversarial nets,'' in \emph{Proc. Neur. Info. Proc. Sys.}, 2016,
  pp. 2172--2180.

\bibitem{GMVAE}
L.~{Yang}, N.~{Cheung}, J.~{Li}, and J.~{Fang}, ``Deep clustering by gaussian
  mixture variational autoencoders with graph embedding,'' in \emph{Proc. IEEE
  Int. Conf. Comput. Vis.}, 2019, pp. 6439--6448.

\bibitem{DeepClusterRelativeEMICCV17}
K.~G. {Dizaji}, A.~{Herandi}, C.~{Deng}, W.~{Cai}, and H.~{Huang}, ``Deep
  clustering via joint convolutional autoencoder embedding and relative entropy
  minimization,'' in \emph{Proc. IEEE Int. Conf. Comput. Vis.}, 2017, pp.
  5747--5756.

\bibitem{sab}
J.~Lee, Y.~Lee, J.~Kim, A.~Kosiorek, S.~Choi, and Y.~W. Teh, ``Set transformer:
  A framework for attention-based permutation-invariant neural networks,'' in
  \emph{Proc. Int. Conf. Mach. Learn.}, 2019, pp. 3744--3753.

\bibitem{srdc}
H.~Tang, K.~Chen, and K.~Jia, ``Unsupervised domain adaptation via structurally
  regularized deep clustering,'' in \emph{Proc. IEEE Conf. Comput. Vis. Pattern
  Recognit.}, 2020, pp. 8725--8735.

\bibitem{Adapt_SegMap}
Y.~{Tsai}, W.~{Hung}, S.~{Schulter}, K.~{Sohn}, M.~{Yang}, and M.~{Chandraker},
  ``Learning to adapt structured output space for semantic segmentation,'' in
  \emph{Proc. IEEE Conf. Comput. Vis. Pattern Recognit.}, 2018, pp. 7472--7481.

\bibitem{advent}
T.~{Vu}, H.~{Jain}, M.~{Bucher}, M.~{Cord}, and P.~{Pérez}, ``Advent:
  Adversarial entropy minimization for domain adaptation in semantic
  segmentation,'' in \emph{Proc. IEEE Conf. Comput. Vis. Pattern Recognit.},
  2019, pp. 2512--2521.

\bibitem{office31}
K.~Saenko, B.~Kulis, M.~Fritz, and T.~Darrell, ``Adapting visual category
  models to new domains,'' in \emph{Proc. Eur. Conf. Comput. Vis.}, 2010, pp.
  213--226.

\bibitem{imageclefda}
``The imageclef-da dataset is available at
  \url{http://imageclef.org/2014/adaptation}.''

\bibitem{officehome}
H.~Venkateswara, J.~Eusebio, S.~Chakraborty, and S.~Panchanathan, ``Deep
  hashing network for unsupervised domain adaptation,'' in \emph{Proc. IEEE
  Conf. Comput. Vis. Pattern Recognit.}, 2017, pp. 5385--5394.

\bibitem{visda2017}
\BIBentryALTinterwordspacing
X.~{Peng}, B.~{Usman}, N.~{Kaushik}, D.~{Wang}, J.~{Hoffman}, and K.~{Saenko},
  ``Visda: A synthetic-to-real benchmark for visual domain adaptation,'' in
  \emph{Workshop of IEEE Conf. Comput. Vis. Pattern Recognit.}, 2018. [Online].
  Available: \url{http://ai.bu.edu/visda-2017/}
\BIBentrySTDinterwordspacing

\bibitem{svhn}
Y.~Netzer, T.~Wang, A.~Coates, A.~Bissacco, B.~Wu, and A.~Y. Ng, ``Reading
  digits in natural images with unsupervised feature learning,'' in
  \emph{Workshop of Proc. Neur. Info. Proc. Sys.}, 2011.

\bibitem{mnist}
Y.~LeCun, L.~Bottou, Y.~Bengio, and P.~Haffner, ``Gradient-based learning
  applied to document recognition,'' in \emph{Proceedings of the IEEE},
  vol.~86, 1998, p. 2278–2324.

\bibitem{usps}
J.~J. Hull, ``A database for handwritten text recognition research,''
  \emph{IEEE Trans. Pattern Anal. Mach. Intell.}, vol.~16, p. 550–554, 1994.

\bibitem{gta5}
S.~R. Richter, V.~Vineet, S.~Roth, and V.~Koltun, ``Playing for data: {G}round
  truth from computer games,'' in \emph{Proc. Eur. Conf. Comput. Vis.}, vol.
  9906, 2016, pp. 102--118.

\bibitem{synthia}
G.~{Ros}, L.~{Sellart}, J.~{Materzynska}, D.~{Vazquez}, and A.~M. {Lopez},
  ``The synthia dataset: A large collection of synthetic images for semantic
  segmentation of urban scenes,'' in \emph{Proc. IEEE Conf. Comput. Vis.
  Pattern Recognit.}, 2016, pp. 3234--3243.

\bibitem{cityscapes}
M.~{Cordts}, M.~{Omran}, S.~{Ramos}, T.~{Rehfeld}, M.~{Enzweiler},
  R.~{Benenson}, U.~{Franke}, S.~{Roth}, and B.~{Schiele}, ``The cityscapes
  dataset for semantic urban scene understanding,'' in \emph{Proc. IEEE Conf.
  Comput. Vis. Pattern Recognit.}, 2016, pp. 3213--3223.

\bibitem{rtn}
M.~Long, H.~Zhu, J.~Wang, and M.~I. Jordan, ``Unsupervised domain adaptation
  with residual transfer networks,'' in \emph{Proc. Neur. Info. Proc. Sys.},
  2016, pp. 136--144.

\bibitem{BeyondSW}
A.~Rozantsev, M.~Salzmann, and P.~Fua, ``Beyond sharing weights for deep domain
  adaptation,'' \emph{IEEE Trans. Pattern Anal. Mach. Intell.}, vol.~41, pp.
  801--814, 2019.

\bibitem{adda}
E.~{Tzeng}, J.~{Hoffman}, K.~{Saenko}, and T.~{Darrell}, ``Adversarial
  discriminative domain adaptation,'' in \emph{Proc. IEEE Conf. Comput. Vis.
  Pattern Recognit.}, 2017, pp. 2962--2971.

\bibitem{iCAN}
W.~{Zhang}, W.~{Ouyang}, W.~{Li}, and D.~{Xu}, ``Collaborative and adversarial
  network for unsupervised domain adaptation,'' in \emph{Proc. IEEE Conf.
  Comput. Vis. Pattern Recognit.}, 2018, pp. 3801--3809.

\bibitem{cdan}
M.~Long, Z.~Cao, J.~Wang, and M.~I. Jordan, ``Conditional adversarial domain
  adaptation,'' in \emph{Proc. Neur. Info. Proc. Sys.}, 2018, pp. 1647--1657.

\bibitem{mada}
Z.~Pei, Z.~Cao, M.~Long, and J.~Wang, ``Multi-adversarial domain adaptation,''
  in \emph{Association for the Advancement of Artificial Intelligence}, 2018,
  pp. 3934--3941.

\bibitem{cat}
Z.~{Deng}, Y.~{Luo}, and J.~{Zhu}, ``Cluster alignment with a teacher for
  unsupervised domain adaptation,'' in \emph{Proc. IEEE Int. Conf. Comput.
  Vis.}, 2019, pp. 9943--9952.

\bibitem{bsp}
X.~Chen, S.~Wang, M.~Long, and J.~Wang, ``Transferability vs. discriminability:
  Batch spectral penalization for adversarial domain adaptation,'' in
  \emph{Proc. Int. Conf. Mach. Learn.}, vol.~97, 2019, pp. 1081--1090.

\bibitem{ClusterAssumption}
O.~Chapelle and A.~Zien, ``Semi-supervised classification by low density
  separation,'' in \emph{Workshop of International Conference on Artificial
  Intelligence and Statistics}, 2005, pp. 57--64.

\bibitem{min_ent}
H.~Li, K.~Zhang, and T.~Jiang, ``Minimum entropy clustering and applications to
  gene expression analysis,'' in \emph{IEEE Computational Systems
  Bioinformatics Conference}, 2004, pp. 142--151.

\bibitem{em}
Y.~Grandvalet and Y.~Bengio, ``Semi-supervised learning by entropy
  minimization,'' in \emph{Proc. Neur. Info. Proc. Sys.}, 2004, pp. 529--536.

\bibitem{rca}
S.~{Cicek} and S.~{Soatto}, ``Unsupervised domain adaptation via regularized
  conditional alignment,'' in \emph{Proc. IEEE Int. Conf. Comput. Vis.}, 2019,
  pp. 1416--1425.

\bibitem{it_cluster_uda}
A.~Rastrow, F.~Jelinek, A.~Sethy, and B.~Ramabhadran, ``Unsupervised model
  adaptation using information-theoretic criterion,'' in \emph{Human Language
  Technologies: The Annual Conference of the North American Chapter of the
  Association for Computational Linguistics}, 2010, pp. 190--197.

\bibitem{dwt_mec}
S.~{Roy}, A.~{Siarohin}, E.~{Sangineto}, S.~R. {Bulò}, N.~{Sebe}, and
  E.~{Ricci}, ``Unsupervised domain adaptation using feature-whitening and
  consensus loss,'' in \emph{Proc. IEEE Conf. Comput. Vis. Pattern Recognit.},
  2019, pp. 9463--9472.

\bibitem{dirt_t}
R.~Shu, H.~Bui, H.~Narui, and S.~Ermon, ``A {DIRT}-t approach to unsupervised
  domain adaptation,'' in \emph{Proc. Int. Conf. on Learn. Rep.}, 2018.

\bibitem{larger_norm}
R.~{Xu}, G.~{Li}, J.~{Yang}, and L.~{Lin}, ``Larger norm more transferable: An
  adaptive feature norm approach for unsupervised domain adaptation,'' in
  \emph{Proc. IEEE Int. Conf. Comput. Vis.}, 2019, pp. 1426--1435.

\bibitem{can}
G.~{Kang}, L.~{Jiang}, Y.~{Yang}, and A.~G. {Hauptmann}, ``Contrastive
  adaptation network for unsupervised domain adaptation,'' in \emph{Proc. IEEE
  Conf. Comput. Vis. Pattern Recognit.}, 2019, pp. 4888--4897.

\bibitem{sntg}
Y.~Luo, J.~Zhu, M.~Li, Y.~Ren, and B.~Zhang, ``Smooth neighbors on teacher
  graphs for semi-supervised learning,'' in \emph{Proc. IEEE Conf. Comput. Vis.
  Pattern Recognit.}, 2018, pp. 8896--8905.

\bibitem{siban}
Y.~{Luo}, P.~{Liu}, T.~{Guan}, J.~{Yu}, and Y.~{Yang}, ``Significance-aware
  information bottleneck for domain adaptive semantic segmentation,'' in
  \emph{Proc. IEEE Int. Conf. Comput. Vis.}, 2019, pp. 6777--6786.

\bibitem{cycada}
J.~Hoffman, E.~Tzeng, T.~Park, J.-Y. Zhu, P.~Isola, K.~Saenko, A.~Efros, and
  T.~Darrell, ``{C}y{CADA}: Cycle-consistent adversarial domain adaptation,''
  in \emph{Proc. Int. Conf. Mach. Learn.}, vol.~80, 2018, pp. 1989--1998.

\bibitem{dise}
W.~{Chang}, H.~{Wang}, W.~{Peng}, and W.~{Chiu}, ``All about structure:
  Adapting structural information across domains for boosting semantic
  segmentation,'' in \emph{Proc. IEEE Conf. Comput. Vis. Pattern Recognit.},
  2019, pp. 1900--1909.

\bibitem{clan}
Y.~{Luo}, L.~{Zheng}, T.~{Guan}, J.~{Yu}, and Y.~{Yang}, ``Taking a closer look
  at domain shift: Category-level adversaries for semantics consistent domain
  adaptation,'' in \emph{Proc. IEEE Conf. Comput. Vis. Pattern Recognit.},
  2019, pp. 2502--2511.

\bibitem{curriculum_da}
Y.~{Zhang}, P.~{David}, H.~{Foroosh}, and B.~{Gong}, ``A curriculum domain
  adaptation approach to the semantic segmentation of urban scenes,''
  \emph{IEEE Trans. Pattern Anal. Mach. Intell.}, vol.~42, pp. 1823--1841,
  2020.

\bibitem{ccan}
J.~{Zhu}, T.~{Park}, P.~{Isola}, and A.~A. {Efros}, ``Unpaired image-to-image
  translation using cycle-consistent adversarial networks,'' in \emph{Proc.
  IEEE Int. Conf. Comput. Vis.}, 2017, pp. 2242--2251.

\bibitem{mdd}
Y.~Zhang, T.~Liu, M.~Long, and M.~Jordan, ``Bridging theory and algorithm for
  domain adaptation,'' in \emph{Proc. Int. Conf. Mach. Learn.}, vol.~97, 2019,
  pp. 7404--7413.

\bibitem{DeepClusterLink}
M.~Jabi, M.~Pedersoli, A.~Mitiche, and I.~Ben~Ayed, ``Deep clustering: On the
  link between discriminative models and k-means,'' \emph{arXiv:1810.04246},
  2018.

\bibitem{PeronaMIDisCluster}
A.~Krause, P.~Perona, and R.~G. Gomes, ``Discriminative clustering by
  regularized information maximization,'' in \emph{Proc. Neur. Info. Proc.
  Sys.}, 2010, pp. 775--783.

\bibitem{UnsupervisedEmbeddingICML16}
J.~Xie, R.~Girshick, and A.~Farhadi, ``Unsupervised deep embedding for
  clustering analysis,'' in \emph{Proc. Int. Conf. Mach. Learn.}, 2016, pp.
  478--487.

\bibitem{kmm}
J.~Huang, A.~Gretton, K.~Borgwardt, B.~Sch\"{o}lkopf, and A.~J. Smola,
  ``Correcting sample selection bias by unlabeled data,'' in \emph{Proc. Neur.
  Info. Proc. Sys.}, 2007, pp. 601--608.

\bibitem{density_estimate}
B.~Zadrozny, ``Learning and evaluating classifiers under sample selection
  bias,'' in \emph{Proc. Int. Conf. Mach. Learn.}, 2004, pp. 114--.

\bibitem{t_sne}
L.~van~der Maaten and G.~Hinton, ``Visualizing data using t-sne,'' \emph{Journ.
  of Mach. Learn. Res.}, vol.~9, p. 2579–2605, 2008.

\bibitem{dsbn}
W.~{Chang}, T.~{You}, S.~{Seo}, S.~{Kwak}, and B.~{Han}, ``Domain-specific
  batch normalization for unsupervised domain adaptation,'' in \emph{Proc. IEEE
  Conf. Comput. Vis. Pattern Recognit.}, 2019, pp. 7346--7354.

\bibitem{multi_head_attention}
A.~Vaswani, N.~Shazeer, N.~Parmar, J.~Uszkoreit, L.~Jones, A.~N. Gomez,
  {\L}.~Kaiser, and I.~Polosukhin, ``Attention is all you need,'' in
  \emph{Proc. Neur. Info. Proc. Sys.}, 2017, pp. 5998--6008.

\bibitem{mmd}
M.~Ghifary, W.~B. Kleijn, and M.~Zhang, ``Domain adaptive neural networks for
  object recognition,'' in \emph{Pacific Rim International Conference on
  Artificial Intelligence}, 2014, pp. 898--904.

\bibitem{cmd}
W.~Zellinger, T.~Grubinger, E.~Lughofer, T.~Natschl{\"a}ger, and
  S.~Saminger-Platz, ``Central moment discrepancy (cmd) for domain-invariant
  representation learning,'' in \emph{Proc. Int. Conf. on Learn. Rep.}, 2017.

\bibitem{imagenet}
J.~Deng, W.~Dong, R.~Socher, L.-J. Li, K.~Li, and L.~Fei-Fei, ``Imagenet: A
  large-scale hierarchical image database,'' in \emph{Proc. IEEE Conf. Comput.
  Vis. Pattern Recognit.}, 2009, pp. 248--255.

\bibitem{resnet}
K.~{He}, X.~{Zhang}, S.~{Ren}, and J.~{Sun}, ``Deep residual learning for image
  recognition,'' in \emph{Proc. IEEE Conf. Comput. Vis. Pattern Recognit.},
  2016, pp. 770--778.

\bibitem{adam}
D.~P. Kingma and J.~Ba, ``Adam: A method for stochastic optimization,'' in
  \emph{Proc. Int. Conf. on Learn. Rep.}, 2015.

\bibitem{lenet}
Y.~{Lecun}, L.~{Bottou}, Y.~{Bengio}, and P.~{Haffner}, ``Gradient-based
  learning applied to document recognition,'' \emph{Proceedings of the IEEE},
  vol.~86, pp. 2278--2324, 1998.

\bibitem{deeplab_v2}
L.~{Chen}, G.~{Papandreou}, I.~{Kokkinos}, K.~{Murphy}, and A.~L. {Yuille},
  ``Deeplab: Semantic image segmentation with deep convolutional nets, atrous
  convolution, and fully connected crfs,'' \emph{IEEE Trans. Pattern Anal.
  Mach. Intell.}, vol.~40, pp. 834--848, 2018.

\bibitem{drcn}
M.~Ghifary, W.~B. Kleijn, M.~Zhang, D.~Balduzzi, and W.~Li, ``Deep
  reconstruction-classification networks for unsupervised domain adaptation,''
  in \emph{Proc. Eur. Conf. Comput. Vis.}, 2016.

\bibitem{raan}
Q.~{Chen}, Y.~{Liu}, Z.~{Wang}, I.~{Wassell}, and K.~{Chetty}, ``Re-weighted
  adversarial adaptation network for unsupervised domain adaptation,'' in
  \emph{Proc. IEEE Conf. Comput. Vis. Pattern Recognit.}, 2018, pp. 7976--7985.

\bibitem{gpda}
M.~{Kim}, P.~{Sahu}, B.~{Gholami}, and V.~{Pavlovic}, ``Unsupervised visual
  domain adaptation: A deep max-margin gaussian process approach,'' in
  \emph{Proc. IEEE Conf. Comput. Vis. Pattern Recognit.}, 2019, pp. 4375--4385.

\bibitem{etd}
M.~Li, Y.-M. Zhai, Y.-W. Luo, P.-F. Ge, and C.-X. Ren, ``Enhanced transport
  distance for unsupervised domain adaptation,'' in \emph{Proc. IEEE Conf.
  Comput. Vis. Pattern Recognit.}, 2020, pp. 13\,936--13\,944.

\bibitem{cada}
V.~K. {Kurmi}, S.~{Kumar}, and V.~P. {Namboodiri}, ``Attending to
  discriminative certainty for domain adaptation,'' in \emph{Proc. IEEE Conf.
  Comput. Vis. Pattern Recognit.}, 2019, pp. 491--500.

\end{thebibliography}
